\documentclass[11pt]{article}

\usepackage[margin=1in]{geometry}
\usepackage{setspace}
\usepackage{times} % or newtxtext,newtxmath
\usepackage{microtype}

\usepackage{amsmath, amssymb, amsthm}
\usepackage{graphicx}
\usepackage{booktabs}
\usepackage[numbers,sort&compress]{natbib}
\usepackage{hyperref}
\usepackage{url}

\usepackage[table]{xcolor}
\usepackage{colortbl}
\usepackage{array}
\usepackage{multirow}
\usepackage{multicol}
\usepackage{makecell}
\usepackage{arydshln}
\usepackage{bm}
\usepackage{caption}
\usepackage{subfig} % \subfloat for sub-figures/sub-tables

\usepackage[bottom]{footmisc}
\newif\ifnotes
\ifnotes
    \newcommand{\hidek}[1]{\textcolor{gray}{#1}}
    \newcommand{\todo}[1]{\textcolor{red}{\textbf{[ToDo: #1]}}}
    \newcommand{\improve}[1]{\textcolor{cyan}{\textit{#1}}}
\else
    \newcommand{\hidek}[1]{}
    \newcommand{\todo}[1]{}
    \newcommand{\improve}[1]{#1}
\fi

\newcommand{\RoT}{\textsc{RoT }}
\newcommand{\rot}{\textsc{RoT}}
\newcommand*\tturl[1]{{
  \color{blue}
  \texttt{\expandafter\dottvar\detokenize{#1}\relax}
}}
\newcommand*\dottvar[1]{%
  \ifx\relax#1\else
    \expandafter\ifx\string-#1\allowbreak\fi
    \expandafter\ifx\string.#1\allowbreak\fi
    \ifx/#1\allowbreak\fi
    \expandafter\ifx\string:#1\allowbreak\fi
    #1%
    \expandafter\dottvar
  \fi
}

\title{\textbf{Rule of Thumb: Explaining Artificial Intelligence Systems \\ using Partial Information}}
\author{
	Kaivalya Rawal$^{\ast1}$,
	Daria Onitiu$^{2}$,
	Brent Mittelstadt$^{1,3}$,
	Sandra Wachter$^{1,2}$,
	Chris Russell$^{1}$ \and
	\small$^{1}$University of Oxford\and
	\small$^{2}$Hasso Plattner Institute\and
	\small$^{3}$Weizenbaum Institute\and
	\small$^\ast$\textit{address correspondence to: kaivalya.rawal@oii.ox.ac.uk}\and
}
\date{}

\begin{document}

\maketitle

\begin{abstract} \bfseries \boldmath

Explainable Artificial Intelligence (XAI) seeks to explain how an Artificial Intelligence (AI) system arrived at a particular decision.
We propose ``Rule of Thumb'' (\rot) explanations, a new approach to XAI based upon a novel formulation that identifies the most relevant features for predicting the behaviour of an AI system, for a particular datapoint.
We show how \RoT is well-suited to enable XAI in:
(a) zero-shot classification using large language models (LLMs), (b) auditing of opaque AI systems without model access, and (c) the use of AI in scientific discovery. 
Additionally, \RoT meets specific requirements from leading AI regulations, provides a familiar interface and visualisations for XAI practitioners, is model-agnostic, and is substantially faster than alternatives.

Code available at: \tturl{https://github.com/KaiRawal/Rule-of-Thumb-Explaining-Artificial-Intelligence-Systems-using-Partial-Information}

\end{abstract}

% \tableofcontents

\section{Introduction}

Artificial intelligence (AI) in the form of large language models has become ubiquitous. One of the key components that has made it disruptive and widespread is its ability to perform zero-shot tasks. Given an applicant's resume, it can say if they are a good fit for a job, or identify an animal in a photo. But does it work? As it stands, the only way to find out if a large language model (LLM) can perform a task, is to label a significant subset of data, and measure how consistent the LLM is with these labels. This data labeling process substantially decreases the disruptive nature of AI. While the AI itself may be zero-shot, the process of deciding if we should deploy it is not. Explainable AI (XAI) has long been promised as a tool to help with this kind of deployment decision. Better understanding of how, and in some cases, why AIs make decisions can increase confidence in deployment and help users identify and correct errors \cite{lime, grad_cam}.

\begin{figure}[!ht]
	\centering
	\captionsetup[subfloat]{justification=centering}
	\begin{tabular}{ccc}
		\subfloat[\RoT explanation for GPT-4o-mini classification label `Cat']{
			\includegraphics[trim={0 0 0 15},clip,width=0.30\textwidth]{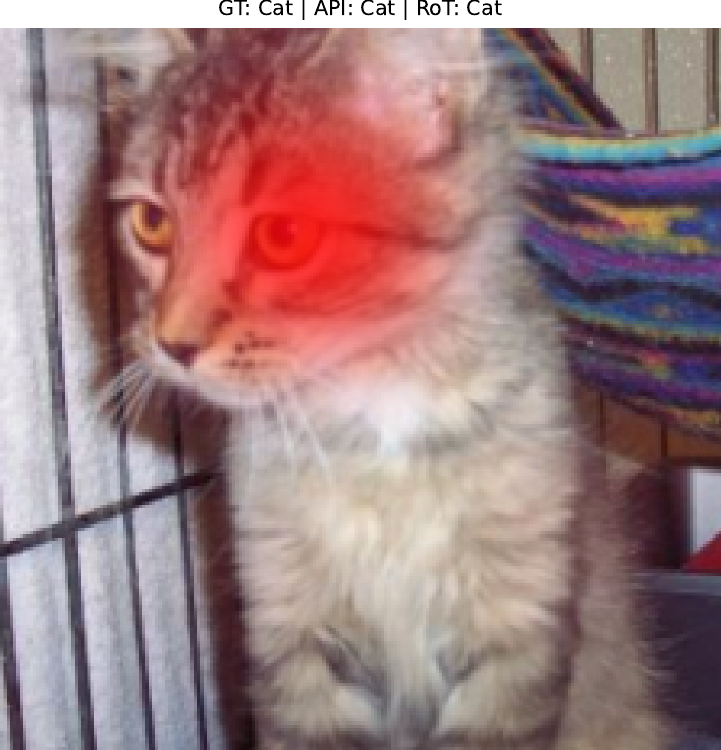}
		} & \subfloat[\RoT explanation for GPT-4o-mini classification label `Dog']{
			\includegraphics[trim={0 0 0 15},clip,width=0.30\textwidth]{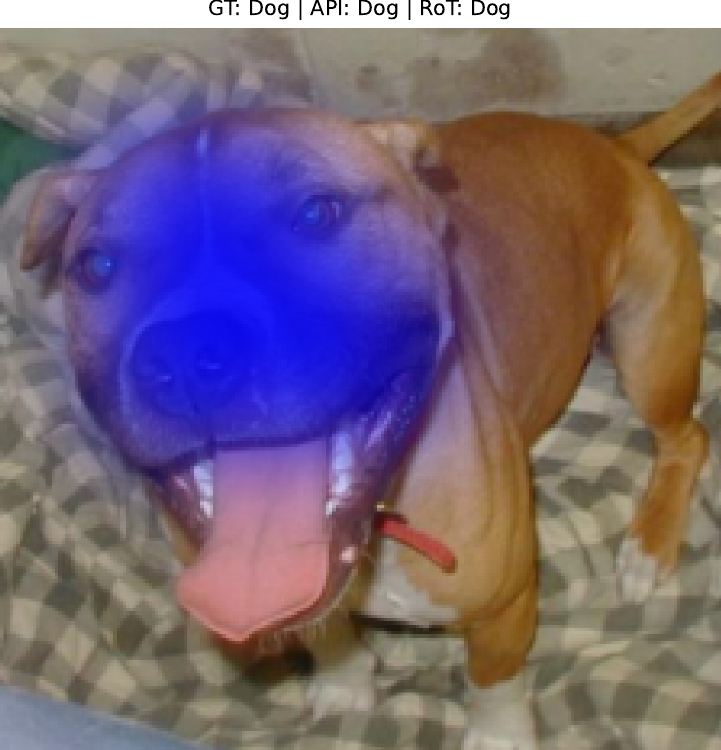}
		} & \subfloat[\RoT explanation for an image with both classes]{
			\includegraphics[trim={0 0 0 15},clip,width=0.30\textwidth]{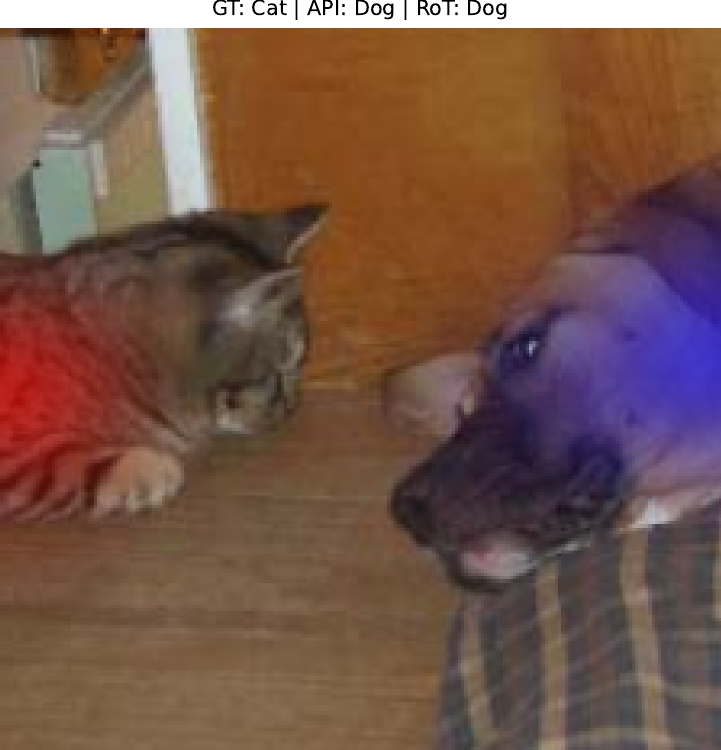}
		} \\
		\subfloat[\RoT explanation for GPT-4o-mini classification label `Cat']{
			\includegraphics[trim={0 0 0 15},clip,width=0.30\textwidth]{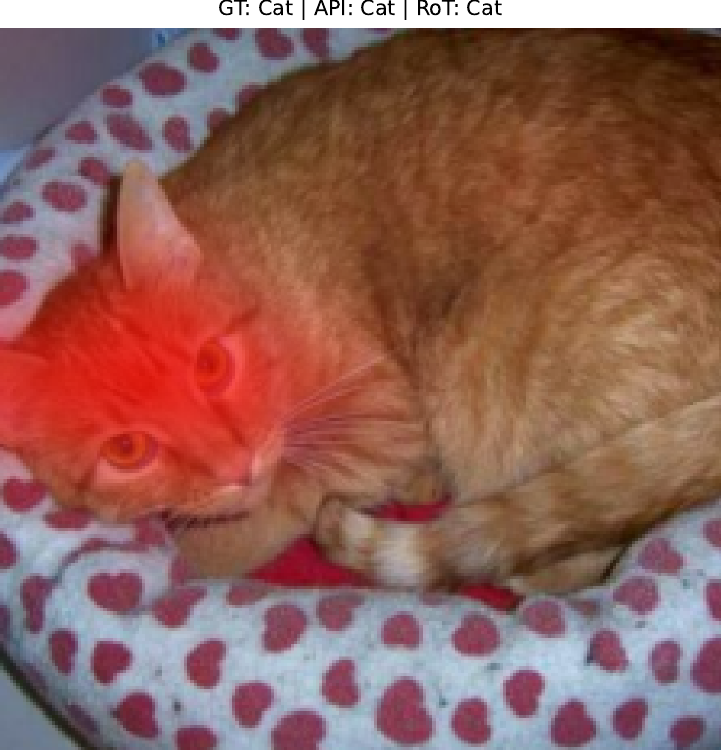}
		} & \subfloat[\RoT explanation for GPT-4o-mini classification label `Dog']{
			\includegraphics[trim={0 0 0 15},clip,width=0.30\textwidth]{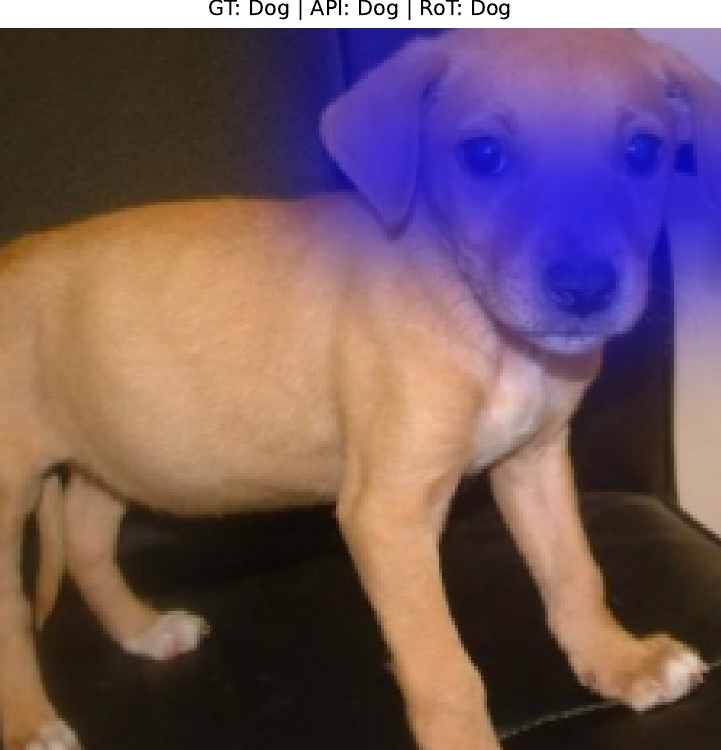}
		} & \subfloat[\RoT explanation for an image with neither class]{
			\includegraphics[trim={0 0 0 15},clip,width=0.30\textwidth]{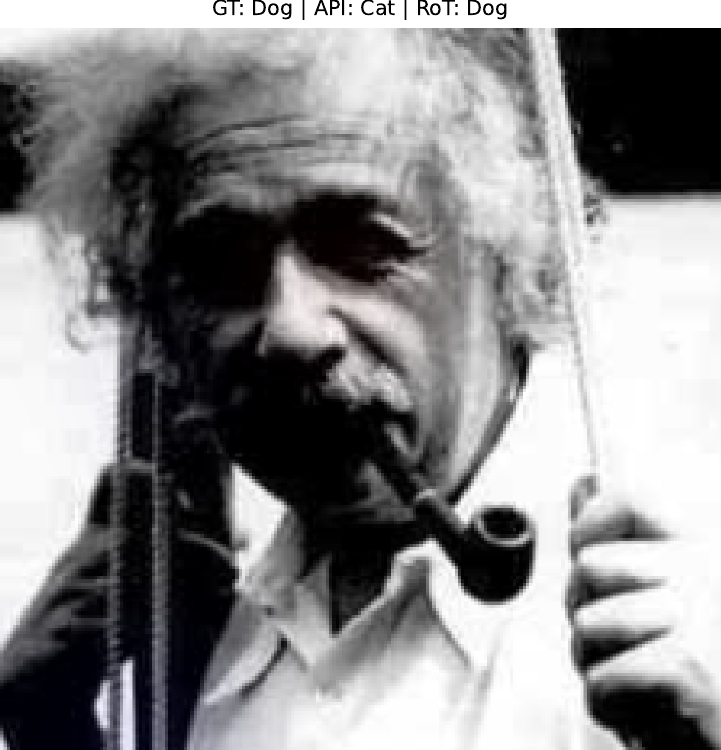}
		} \\
	\end{tabular}
	\caption{\textbf{Examples of local feature importances for zero-shot image classification performed via the OpenAI GPT-4o-mini API.} The model is prompted to perform zero-shot classification of pet images into labels `Cat' and `Dog' from a public dataset\cite{pet_images_kaggle}. \RoT is orchestrated to produce positive importances for input features indicative of `Cat', highlighted {\color{red} red}, and negative importances for features indicative of `Dog', highlighted {\color{blue} blue}. \RoT explanations can be computed without access to model weights, and correctly handle cases ((c) and (f)) where GPT-4o-mini disagrees with the provided labels. Here the explanation indicates that supporting regions for both classes can be found in (c), while no regions significantly support either class in (f).
    }
	\label{fig:explanation_example_pets}
\end{figure}

However, precisely when you would expect XAI to be making a resurgence, it is nowhere to be seen. In no small part, this is because of fundamental limitations of the methods. Existing Explainable AI methods try to either peer inside the algorithmic black-box and use access to the internals of models (grad-cam \cite{grad}, integrated gradients \cite{integratedgradients}, counterfactuals \cite{wachter2017counterfactual}, or mechanistic interpretability \cite{bereska2024tmlr-mechanistic}) which is inapplicable to many widely used closed source LLMs that only allow access via an application programming interface (API), or are {combinatorially enumerative}, and rely on passing a large amount of synthetic data to the model (SHAP \cite{SHAP:10.5555/3295222.3295230}, LIME \cite{lime}, Counterfactual explanations \cite{wachter2017counterfactual} or Recourse \cite{NEURIPS2020_8ee7730e}), which is either prohibitively expensive or simply infeasible.

Feature importance explanations, such as those from LIME or SHAP, are post-hoc explanations of individual predictions in the form of a signed vector of scores, denoting the importance of each input feature to the prediction\footnote{LIME and SHAP have together been cited over 100,000 times, and are the most common XAI used in practice.}. Prior techniques have been variants of sensitivity analysis, operationalising the intuition that important input features are those which, upon synthetic perturbation, cause the greatest changes in model outputs. We focus on post-hoc, local (per-datapoint) feature importance explanations because they are a popular and easy to use form of XAI, do not require retraining AI systems, and can be implemented separately from the predictions they seek to explain.

\section{Rule-of-Thumb: Explanations for Emerging XAI Applications}

We propose ``Rule-of-Thumb'' (\rot), a novel form of XAI that considers important inputs to be those that are most predictive of the outputs of an AI system. Rather than formulating XAI in terms of how altering an input value would alter the system output, we ask how knowledge of particular feature values should alter our predictions of the AI system's behaviour.

\begin{figure}[!b]
	\centering
	\includegraphics[page=1,trim=1cm 12.5cm 1cm 1cm,clip,width=1.0\linewidth]{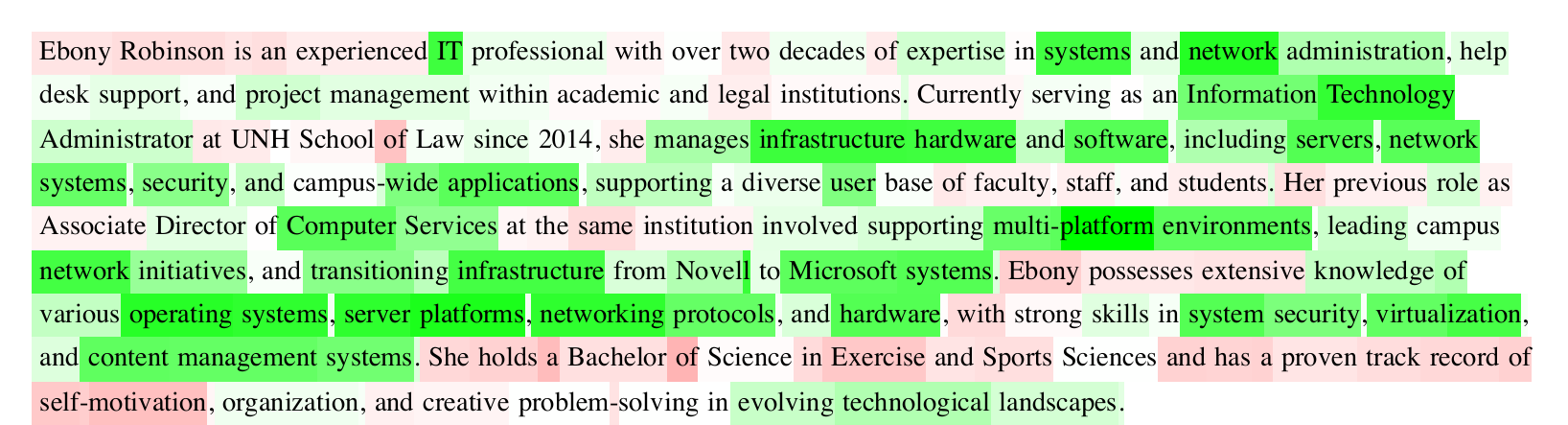}
    \rule{\linewidth}{0.5pt} % horizontal line
	\includegraphics[page=1,trim=1cm 12cm 1cm 0.5cm,clip,width=1.0\linewidth]{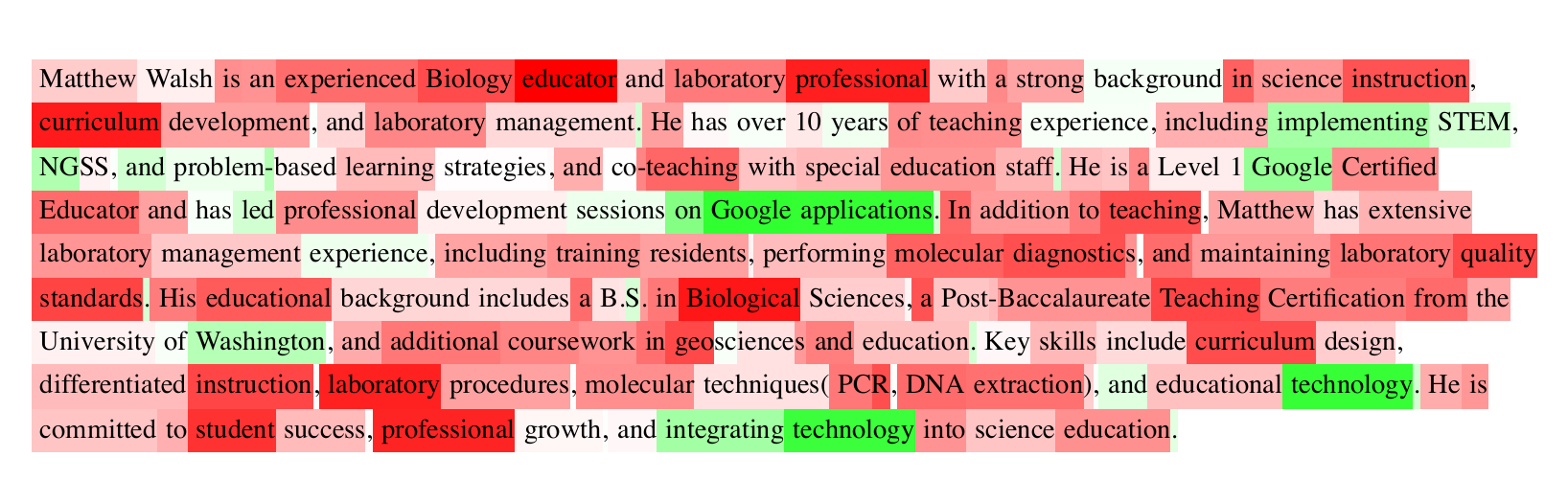}
    % \vspace{-0.5cm}
	\caption{\textbf{Resume summaries \cite{veldanda2023investigating} from an IT-worker (Ebony Robinson, top) and a biology teacher (Matthew Walsh, bottom), that  respectively pass and fail our AI filter}. The zero-shot resume classification system \cite{veldanda2023investigating} uses the GPT-4.1-nano API to find suitable candidates for an information technology (IT) role. \RoT explanations indicate the predictive value of individual text tokens, visualised here by using deeper colour for greater importances. Extra information passed to \RoT but found to be relatively unimportant, such as race, gender, and political preferences is not displayed. Tokens with positive contributions are {\color{green} green}, and with negative contributions are {\color{red} red}.}
	\label{fig:resume_examples}
\end{figure}

The outputs from \RoT are standard feature importance explanations, which can be analysed using existing visualisations such as pixel or token level saliency maps that highlight positive or negative contributions towards an output (Figures \ref{fig:explanation_example_pets} and \ref{fig:resume_examples}). These figures show the unique ability to use \RoT even when model access is available only via API, where other explainers are unviable\footnote{Sensitivity analysis via perturbation \cite{SHAP:10.5555/3295222.3295230,lime} requires prohibitively expensive API calls for many synthetically perturbed datapoints, whereas sensitivity analysis using gradient based methods \cite{grad,smoothgrad,integratedgradients,itg} requires access to weights}. For tabular data where model access is common, we are able to compare with explainers like SHAP and LIME using standard force-plots for a single datapoint (Figure \ref{fig:explanation_example}) or swarmplots for a collection of datapoints (Figure \ref{fig:swarmplots}).

In these latter comparisons, we note that applying sensitivity analysis implicitly requires several nontrivial assumptions: \emph{(i)} that it is possible to query the AI system with new datapoints; \emph{(ii)} the system responds similarly to synthetically perturbed data in the same way it does to real-world data, and \emph{(iii)} The outputs of a system are continuous, and typically vary when a limited number of features are changed \footnote{This is particularly problematic when modern AI systems answer yes/no questions e.g. "Is this a picture of a cat?"}. \RoT makes none of these assumptions, conferring specific advantages in the following three emerging XAI application areas:

\begin{figure}[!t]
	\centering
	\captionsetup[subfloat]{justification=centering}
	\subfloat[\textbf{Rule of Thumb (\rot)} explanation for a datapoint.\\Glucose and age ({\color{cyan}in blue}) contribute most to reducing\\the predicted probability]{
		\includegraphics[width=0.49\linewidth]{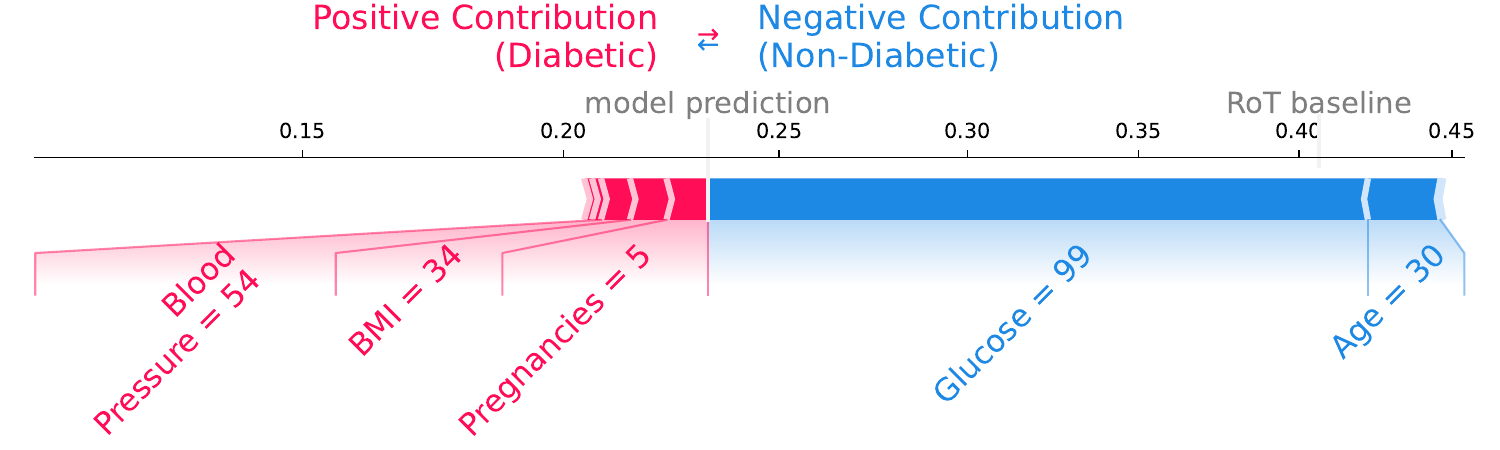}
	}
	% \hfill
	\subfloat[\textbf{SHAP} explanation for the same datapoint. Glucose\\({\color{cyan}in blue}) is still most important in reducing output\\probability, but here age ({\color{red} in red}) increases it]{
		\includegraphics[width=0.49\linewidth]{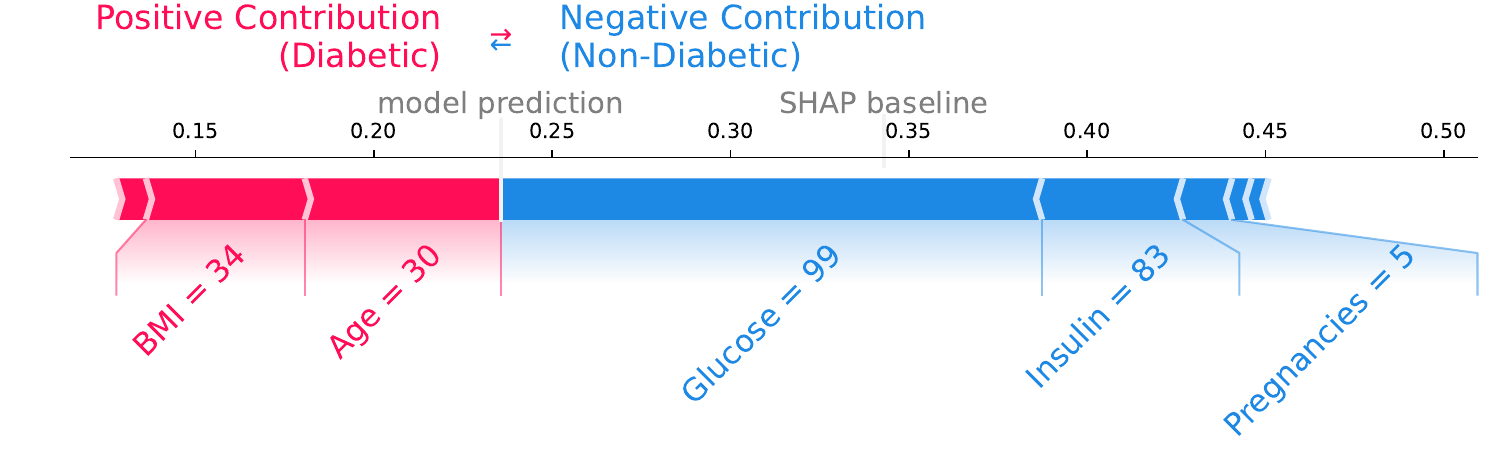}
	}
	\caption{\textbf{Examples of local feature importance explanations from a Random Forest diabetes prediction system.} The AI is trained on the Pima dataset \cite{pima} and for a datapoint with feature values: glucose level = 99, age = 30, BMI = 34, etc, predicts a diabetes probability of 0.235. \RoT and SHAP explain this differently, but glucose is the most important input feature for both. Standard force plots visualise how individual feature importances contribute to the prediction.}
	\label{fig:explanation_example}
\end{figure}

\paragraph{1. Explaining Zero-shot Classification with LLMs:} Classification with LLMs presents two practical challenges. First, weights and gradients are inaccessible for closed systems accessed via API, and second, even when the AI system is open-source, producing each explanation requires computing many additional LLM inferences, making explanations too cumbersome and slow to compute. We experiment with a movie sentiment prediction task and a judicial case outcome prediction task to reflect each of these settings. In both cases, we find \RoT explanations are aligned with expected explanations produced by human annotators. We also find \RoT feasible to compute quickly on consumer hardware, and as \RoT explanations do not require the training a model for each datapoint explained, our computational benefits increase with the number of data points explained. When explaining an open-weight LLM, asymptotically, 13 million new \RoT explanations can be computed in the time taken for a single additional SHAP explanation. \improve{We also conduct an additional experiment filtering resume texts, where we demonstrate the unique \RoT ability to explain predictions in terms of non-input features.}

\paragraph{2. Auditing Proprietary Models:} Auditing AI systems such as e-commerce recommendation engines can be challenging because access to the underlying AI is often restricted. A standard workaround is to first construct an additional mimic model to replicate the AI system, and then explain the mimic model in turn as a proxy for the restricted AI system itself. We go on to show that explanations from this commonly employed workaround can be contested by varying the mimic model used; whereas \RoT directly provides explanations without depending on mimic models.

\paragraph{3. Scientific Discovery using XAI:} 
We identify an emerging strategy to generate scientific hypotheses, which first collects data about a phenomenon of interest, then trains an AI system using the data, and finally explains its predictions in order to provide insights into the phenomenon itself. \improve{Using sensitivity analysis based explainers here is challenging because sometimes they generate incorrect hypotheses.} We analyse datasets about diabetes diagnoses, criminal justice, and lending; comparing hypotheses from \rot, SHAP, and LIME explanations. Through a carefully constructed scenario where a specific input feature is unpredictive of model behaviour, and another where a specific input feature is predictive, we demonstrate that only \RoT explanations correctly identify these features as unimportant and important respectively.

\subsection{Formulation}

\label{sec:form}

We wish to determine which features are most informative in estimating a given prediction from an AI system. Using additive functions that model the AI system outputs from individual features, we can both estimate the prediction and determine the relative importance of each feature value.

We make some simplifying assumptions to answer this efficiently. 
Given a model \( C(\cdot) \) defined over features $\mathbb J$, and dataset $X$, we want to find the optimal set of additive functions such that for any subset of features $J \subseteq \mathbb J$ and datapoint \( x \in X \) we  estimate $C(x)$ using their sum.

\begin{align}
    C(x) &\approx F \left( \sum_{j \in J} f_{\theta_{j}}(x_j) + G \right) \;\; \forall J \in \mathcal{P}(\mathbb{J})
    \label{eqn:rot_defn}
\end{align}

where $F$ is a sigmoid function (for classification) or an identity function (for regression), $f_{\theta_j}(x_j)$ is a learnt function representing the information provided by knowing that the $j^\text{th}$ feature takes value $x_j$, and $G$ a global bias term that indicates what prediction should be made without knowing anything about the datapoint $x$. We refer to $f_{\theta_j}(x_j)$ as the importance of a feature $j$ taking value $x_j$. This corresponds to how much a prediction of $C(x)$ should update, on average, after discovering this information. 

Given a random subset of features selected using uniform probability $p$, we can  approximate $C(x)$ as a the sum of importances. This idea of always having a rough estimate given any amount of limited information gives rise to the name  \emph{Rule of Thumb}. To find the optimal weights \( \theta_j\, \forall j \in \mathbb{J} \) for our model, we minimize the following objective \(\mathcal{L}\).

\begin{equation}
    \min_{\theta} \mathcal{L}
    =
    \min_\theta \;
    \textcolor{purple!95!black}{
        \underbrace{
            \textcolor{black}{
                \sum_{ {J} \subseteq \mathbb{J}}
            }
        }_{\substack{\text{all subsets}\\\text{of features}}}
    }
    \;
    \textcolor{purple!95!black}{
        \overbrace{
            \textcolor{black}{
                \sum_{x \in X}
            }
        }^{\text{all datapoints}}
    }
    \left(
        \textcolor{purple!95!black}{
            \underbrace{
                \textcolor{black}{
                    p^{ |J| } \; (1-p)^{(|\mathbb{J}|-|J|)}
                }
            }_{\substack{\text{probability of selecting}\\\text{feature subset }J}}
        }
        \; \cdot \;
        \textcolor{purple!95!black}{
            \overbrace{
                \textcolor{black}{
                    \ell\left[
                        C(x), \;\;
                        F \left(
                            \sum_{j \in J}f_{\theta_{j}}(x_j)
                            +
                            G
                        \right)
                    \right]
                }
            }^{\substack{\text{loss between }C(x)\text{ and the estimate}\\
                         \text{using only features in }J}}
        }
        \;
    \right)
    \label{eqn:loss_formulation}
\end{equation}

Here the term \(\ell\left[ C(x), \;\; F \left( \sum_{j \in J}f_{\theta_{j}}(x_j) + G \right) \right]\) is a standard training loss such as log loss or squared error that penalizes mispredictions of $C(x)$. This loss is not only marginalized over all datapoints, as is common in standard machine learning, but also all possible masks of the data. Although computing \(\mathcal{L}\) exactly is expensive and requires computing the standard loss over the entire dataset $2^{|\mathbb{J}|}$ times, it can be efficiently optimized using dropout\cite{JMLR:v15:srivastava14a} applied to the feature importances $f_{\theta_{j}}(x_j)$. Using dropout, the training time for fitting is almost the same as fitting the simple additive form of equation \eqref{eqn:rot_defn} that does not consider subsets of features. After fitting the model, the feature $j$ with the largest absolute value $|f_{\theta_j}(x_j)|$ is the one that should most change our confidence in the prediction, and can be considered the most important input feature.

\paragraph{Comparison with existing methods}
Gradient-based methods\cite{grad,smoothgrad,itg} are computationally efficient but require direct access to model weights, which are often restricted. With LLMs ,it is common to provide access only via API access. Gradient-based approaches cannot be applied to any model that only offers API access.

Sensitivity-based approaches LIME and SHAP optimise a per-datapoint objective. While \RoT masks information and fits one model for the entire dataset, these approaches peturb datapoints and fit one linear model per datapoint. This creates substantial differences in applicability -- not only is sensitivity analysis badly suited for analyzing opaque AI systems that only return yes/no decisions (such as those provided by an LLM API), but it also requires many times (typically $500$ for SHAP to $5000$ for LIME) more inferences or API calls to explain each prediction. Finally, by definition, sensitivity analysis can explain AI systems only in terms of their inputs. \RoT can utilise non-input features by fitting on extended datasets if required. This enables explaining model predictions using external models such as MobileNet\cite{mobilenetv3} or BERT\cite{devlin-etal-2019-bert} to generate more informative explanations. See Appendix \ref{appendix:rel} for additional details situating \RoT with respect to prior XAI approaches.

\section{Experiments}

Explanations obtained through sensitivity-based XAI techniques often serve as an aid for users and developers in understanding and debugging model behaviour. However, they fall short in key emerging application areas: zero shot classification with LLMs, auditing black-box AI, and scientific discovery using AI. \improve{ \RoT confers key advantages here: it can be used even if models are slow, expensive, accessible only via API, or inaccessible for additional inferences entirely. It can generate model explanations using input and non-input features, allowing explanations to supplement inputs with other data to explain model behaviour. Finally, \RoT makes no causal assumptions and is invulnerable to adversarial attacks that make other explainers ill-suited to proposing scientific hypotheses or auditing black-box systems.}

Compared to SHAP, \RoT reveals how informative it is to know on average, that a feature takes a certain value, while SHAP reveals the average change that occurs upon altering a feature to a certain value from its mean. This is valuable because it allows us to visualise \RoT explanations using the same familiar plots as SHAP explanations (eg. the force-plots in Figure \ref{fig:explanation_example}). This relationship makes SHAP a natural basis for experimental comparisons with \RoT. 

\subsection{Model Evaluation and Zero-Shot Classification with LLMs}
\label{sec:llms}

LLMs are often used for zero-shot classification when there is no labelled dataset to train a classifier model. For each datapoint, an LLM is prompted to output a classification. Since there are no target labels to compare predictions with, typical model performance metrics such as accuracy, precision or recall cannot be readily computed. XAI offers a complementary path to performance evaluation, allowing us to inspect and verify that the ``correct'' features are predictive of the model's behaviour.

A key challenge is that new predictions can be difficult to obtain for arbitrarily modified model inputs. When using commercial models, this manifests as increased costs (by default, SHAP requires 500 additional API calls to explain a single API prediction, while LIME requires 5000). Even with an open source model, this causes explanations to be prohibitively slow to compute. By contrast \RoT does not require additional API calls for explanation. To improve \rot's predictive power on text, we generate per token embeddings and treat these as features, performing dropout at the token level.

\begin{figure}[!b]
	\centering
    \captionsetup[subfloat]{justification=centering}
	\subfloat[A datapoint with text segments highlighted\\as per \RoT importances]{
		\includegraphics[width=0.49\linewidth]{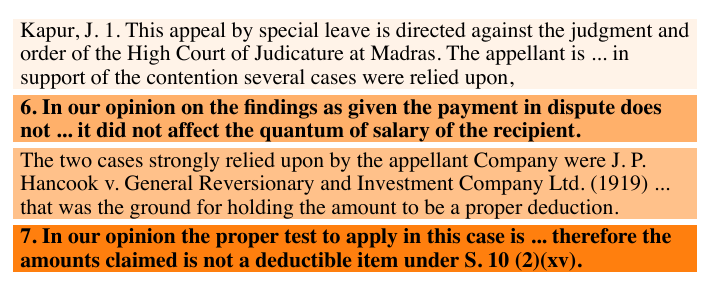}
	}
	% \hfill
	\subfloat[The same datapoint with segments highlighted\\as per SHAP importances]{
		\includegraphics[width=0.49\linewidth]{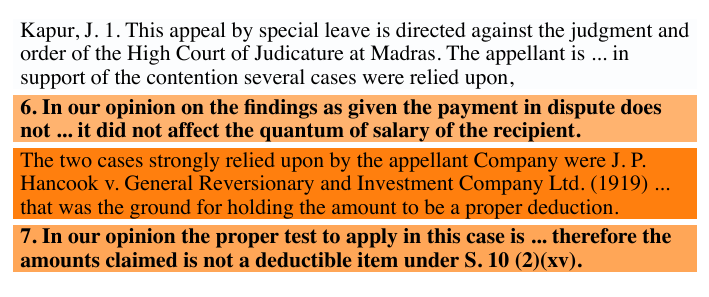}
	}
	\caption{\textbf{Token level explanations for zero-shot text classification with LLMs.} A judicial appeal is rejected by an AI system (fine-tuned RoBERTa LLM)\cite{PredEx} making zero-shot predictions. This prediction is explained by \RoT and SHAP, with more important text segments highlighted with a deeper colour. Human annotation of the segments is indicated in bold, and aligns better with \RoT than with SHAP. We present abridged texts here, with unabridged results from \rot, SHAP, LIME, and Integrated Gradients presented in Figures 
    \ref{fig:good_case_example_377_RoT}, \ref{fig:good_case_example_377_SHAP}, \ref{fig:good_case_example_377_IG}, \ref{fig:good_case_example_377_LIME500}, and \ref{fig:good_case_example_377_LIME5000}.}
	\label{fig:good_case_example_377}
\end{figure}

\begin{table}[!t]
	\centering
	\caption{ \textbf{Measuring alignment between human annotations and token explanations.} For legal case outcomes predicted by an LLM\cite{PredEx}, binary annotations indicate which text segments from the judgement are important. Token importance explanations from an explainer can be considered to ``predict'' these annotations, which we evaluate using the area under the receiver-operating-characteristic curve, weighted by the length per evaluated text segment. \RoT and SHAP explanations are the most aligned with human annotations.}
	\begin{tabular}{l!{\vrule width 0.8pt}cccccc}
	\toprule
	 & \multirow{2}{*}{ \makecell[c]{ \textbf{\rot} } } & \textbf{SHAP} (default & \textbf{Integrated} & \textbf{LIME} (default & \multirow{2}{*}{ \makecell[c]{ \textbf{Random} } } \\
	 & & 500 samples) & \textbf{Gradients} &  5000 samples) &  \\
	\midrule
	\textbf{Avg. wAUROC} & \textit{\textbf{0.77}} & 0.74 & 0.65 & 0.62 & 0.47 \\
	\bottomrule
	\end{tabular}
\label{tab:case_aucs}
\end{table}

We replicate three uses of LLMs for zero-shot classification: the use of commercial LLM APIs to filter resumes \cite{veldanda2023investigating}, the use of API based LLMs to classify movie reviews \cite{movies, eraser}, and the use of a fine-tuned language model to predict the outcomes of judicial appeals from a court summary\cite{PredEx}. Examples of explanations from the resume filtering task can be seen in Figure \ref{fig:resume_examples}. A word-cloud constructed from token-level importances across the corpus is visualised in Figure \ref{fig:word_clouds}. Since this experiment was conducted entirely via APIs, no other explainers can be used to provide baseline comparisons. We run into the same problem with our movie review sentiment classification experiment, but we are able to compare \RoT generated explanations against ``ground-truth'' token importances from human annotators. This provides us with a quantitative measure of the agreement between humans explaining movie review sentiments and \RoT explaining an APIs classification of the same sentiments. For Judicial case outcome prediction, we take advantage of access to the model weights to compare \RoT explanations with other explainers, and visualise the agreement competing explanations have with human annotations in Figure \ref{fig:good_case_example_377}. Overall, we find \RoT and SHAP explanations to be better aligned with human annotators, and \RoT explanations fastest to compute.

\subsubsection{Explaining Judicial Case Predictions from a Fine-Tuned RoBERTa Model}
To test \RoT explanations on LLMs, we study an AI system\cite{PredEx} to predict appeal outcomes in Indian courts. Figure \ref{fig:good_case_example_377} presents an abridged example of explanations from one prediction. The data includes annotation labels for individual segments in each case. These indicate which portions of the text were found important to the judgement by human annotators\cite{PredEx}, allowing us to measure explanation quality\footnote{Quality here refers only to alignment with the judgement of law students, not necessarily to better explanations} by computing the area under the reciever operating characteristic (AUROC) between \RoT explanation importances and known ground-truth importances per text segment. As a fine-tuned RoBERTa classifier is used here, we are able to use other explainers, and Table \ref{tab:case_aucs} presents AUROC scores from different explainers. A detailed analysis of program runtimes is provided in Figure \ref{fig:llm_speedups}, but it is worth noting that \RoT is the only explainer able to produce results in reasonable time on consumer hardware. Details about data preparation, explanation generation, and weighted average AUROC evaluation can be found in Appendix \ref{M_indian_cases}, and individual case explanations can be perused at: \tturl{https://kairawal.github.io/Rule-of-Thumb-Explaining-Artificial-Intelligence-Systems-using-Partial-Information/JudicialCaseOutcomePrediction/Code/results.html}.

\subsubsection{Explaining Movie Review Classifications from an LLM API}
We conduct a similar experiment with zero-shot movie review sentiment classification, using an AI system built upon OpenAI's GPT-4.1-nano LLM, accessible only via API. When compared with ground truth importance annotations\cite{movies, eraser} for review sentiments, \RoT achieves a weighted AUROC of 0.72, while a random baseline only achieves 0.50. Gradient based explainers (Integrated Gradients) cannot be compared, as access to GPT-4.1-nano weights is restricted, and perturbation based explainers (LIME, SHAP) are infeasible to apply even at this modest scale. This showcases a unique \RoT ability, where we can explain discrete decisions without either access to its weights, ors without making additional API calls, simply by looking at predictions already made by the AI. Additional details are in Appendix \ref{M_movie_reviews}.

\begin{figure}[!t]
	\centering
	\includegraphics[width=0.98\linewidth]{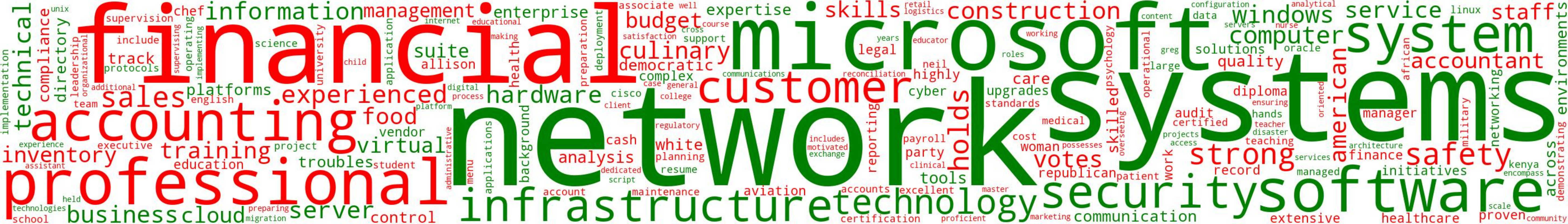}
	\caption{\textbf{Word-cloud summary of token importances for zero-shot resume classification.} A zero-shot resume filter system uses the GPT-4.1-nano LLM API to find suitable candidates for an information technology (IT) role. \RoT explanations indicate the importance of individual text tokens, which we visualise using an importance-weighted word cloud. Alongside IT workers which pass the filter, our corpus contains resumes of bankers, accountants, teachers, and others which our filter rejects. Tokens which increase the likelihood of making a positive prediction are in {\color{teal} green}, and those that decrease the likelihood are in {\color{red} red}.}
	\label{fig:word_clouds}
\end{figure}

\subsubsection{Explaining Resume Filtering from Proprietary LLMs}
In addition to classifying movie reviews, we replicated a zero-shot resume classification system to filter candidates for information-technology (IT) jobs using the GPT-4.1-nano API \cite{veldanda2023investigating}. \RoT was then used to generate explanations identifying which parts of the resume text were important to selecting candidates as suitable for an IT position. We augment our text-based features with three attributes -- race, gender, and political orientation, utilising \rot's unique ability to produce explanations using non-input features. We verify findings from the original study indicating that, on this dataset, GPT ascribes negligible importance to race, gender, and political orientation\footnote{\improve{This was evaluated as per\cite{veldanda2023investigating} by artificially varying names and political orientation on otherwise static resumes}} when making hiring decisions. We evaluate the system with a corpus of resumes containing both IT workers and other occupations. The word-cloud in Figure \ref{fig:word_clouds} presents the text identified as predictive of selection and rejection across the resume corpus, and Figure \ref{fig:resume_examples} shows examples of an IT worker's and a teacher's resume, with tokens highlighted according to the \RoT importances. Additional data and implementation details can be found in Appendix \ref{M_resumes}.

\subsection{Auditing Proprietary Black-Box AI Systems}
\label{sec:auditng}
We replicate an audit of Amazon's e-commerce recommendation system conducted by The Markup \cite{markup}. Lacking access to Amazon's AI system, we adopt the standard practice\cite{surrogate:10.1145/3366424.3383110, arize_surrogate_models, fiddler_surrogate_models, surrogate_azure} of using a dataset of scraped listings from Amazon to train a mimic AI that predicts product rankings. 
We then explain mimic predictions using SHAP and LIME, and find these to vary significantly depending on the choice of mimic. This ambiguity is missing from competing \RoT explanations, which explain the AI recommendation system directly without using any mimic AI. Our analysis using \RoT is also able to surface insights missed by mimic based explanations.

\begin{figure}[!t]
	\centering
    \captionsetup[subfloat]{justification=centering}
	\subfloat[Explanations from The Markup's\\Random Forest mimic model. Using\\SHAP, they report ``brand is amazon''\\and ``product reviews'' to be important]{
		\includegraphics[width=0.32\linewidth]{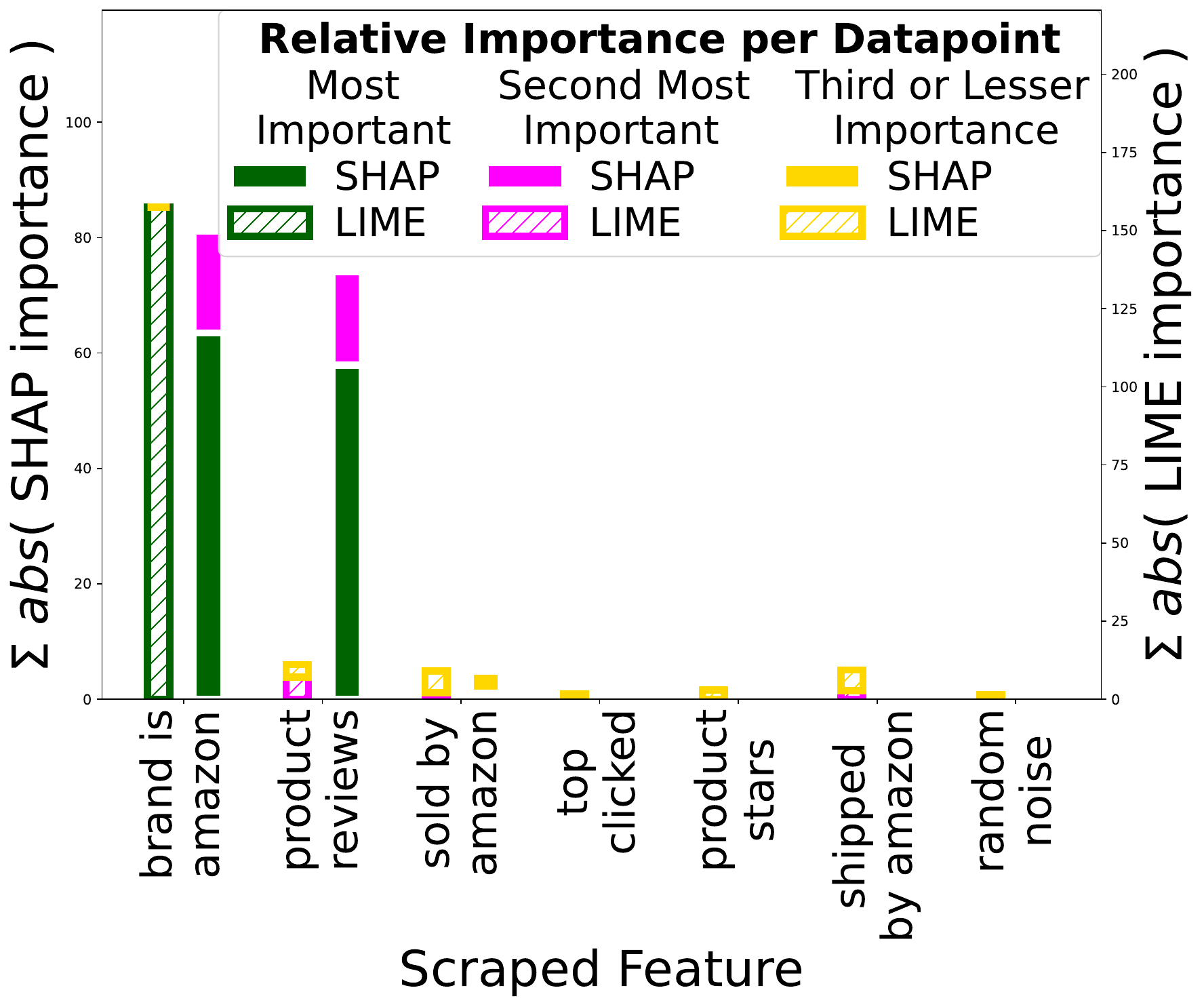}
	}
	\hfill
    % \captionsetup[]{justification=centering}
	\subfloat[Explanations from the scikit-learn\\default Random Forest mimic model\\differ, but continue to indicate the\\importance of ``brand is amazon'']{
		\includegraphics[width=0.32\linewidth]{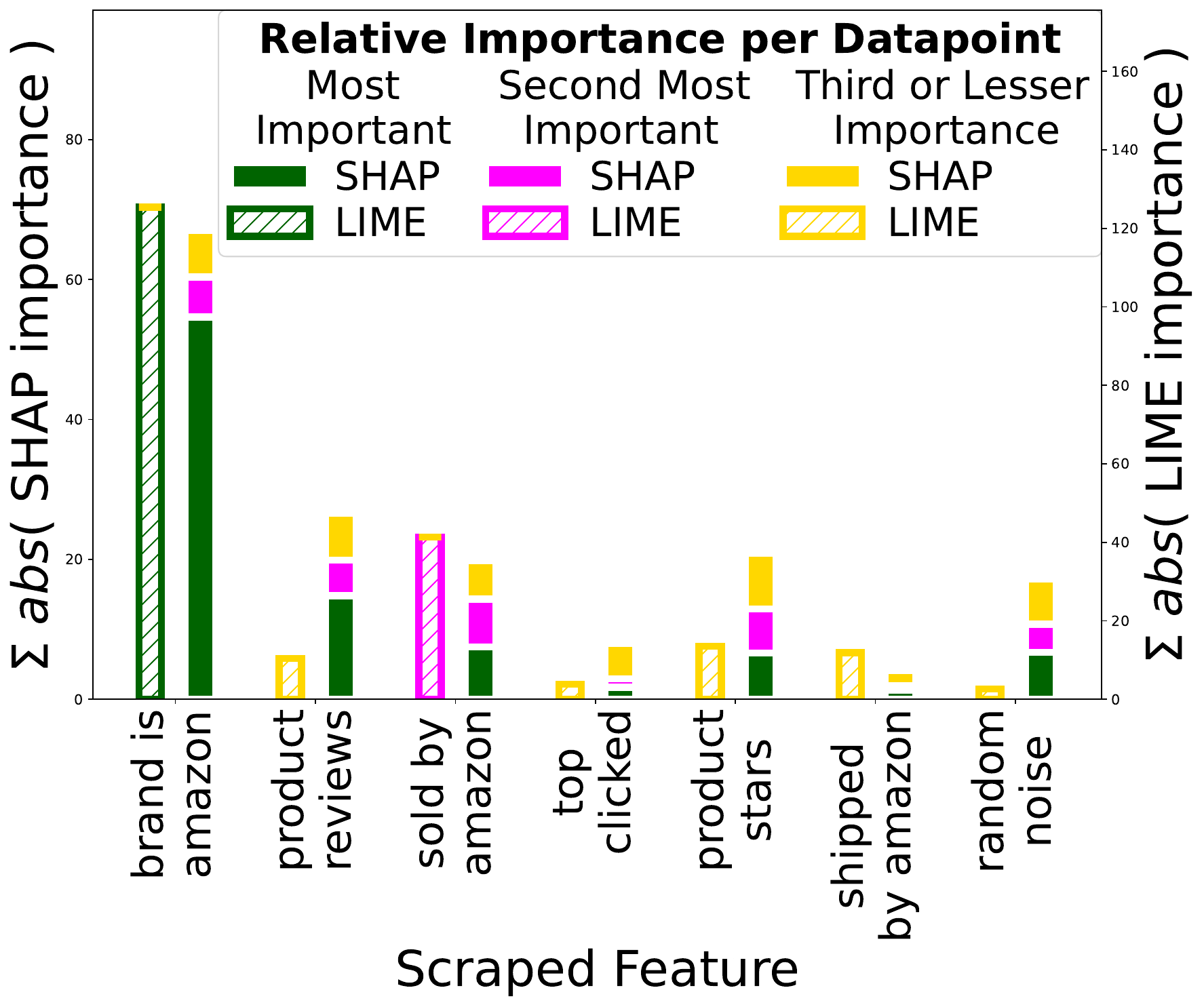}
	}
	\hfill
    % \captionsetup[]{justification=centering}
	\subfloat[\RoT importances obtained without\\mimic models: ``brand is amazon'' and\\``sold by amazon'' are important,\\``product reviews'' is unimportant]{
		\includegraphics[width=0.32\linewidth]{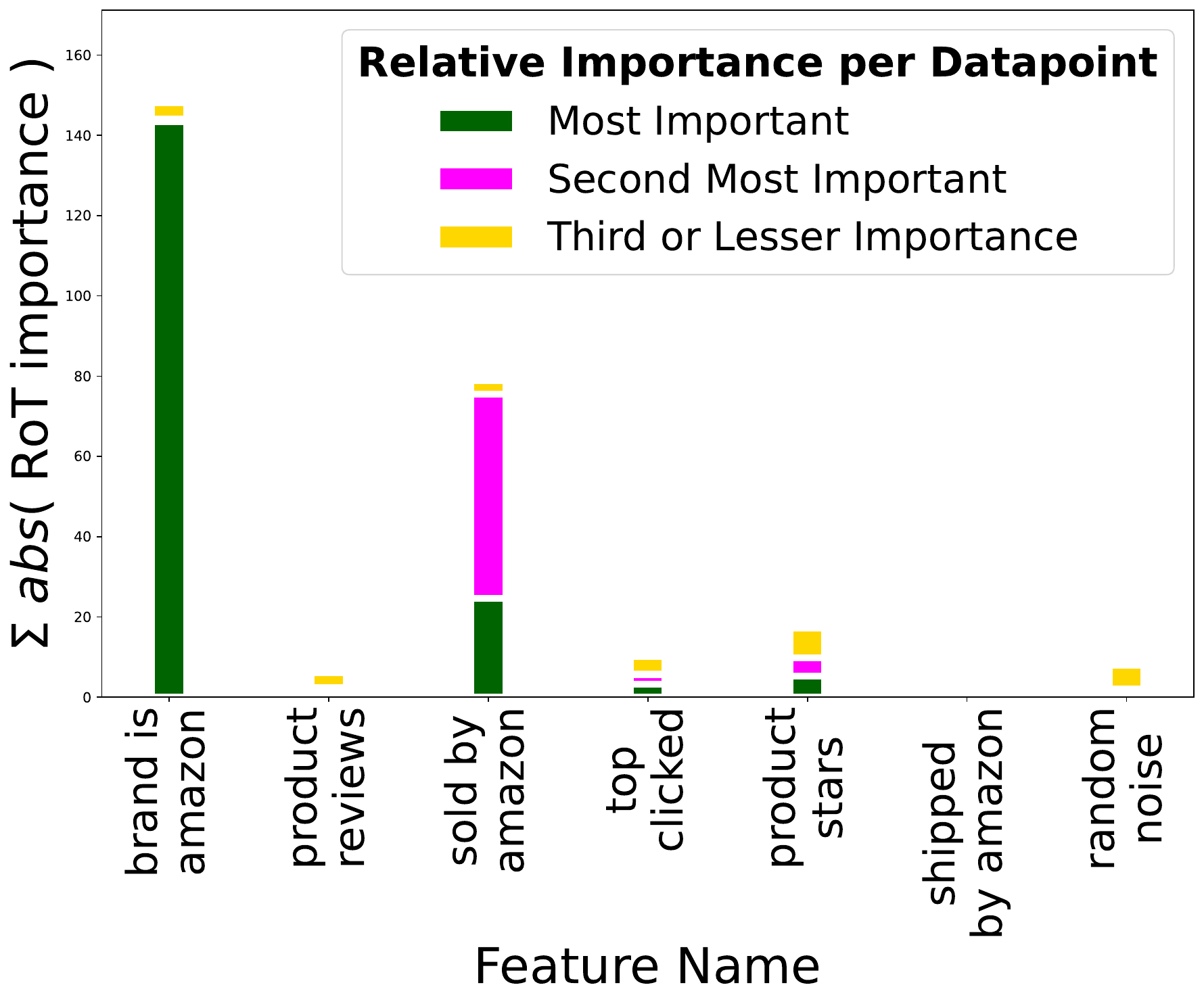}
	}
	\caption{\textbf{Auditing Amazon's proprietary recommendation system through mimic models (Table \ref{table:markup_mimic_accuracies}),  and directly with \rot}. We visualize explanations via their absolute feature importances, summed over all datapoints. We also indicate the proportion of the importance sum where the feature was the first ({\color{teal} in green}) or second ({\color{magenta} in pink}) most important for each datapoint. With SHAP, Markup's mimic model shows brand and product reviews as important features \textbf{(a)}. But the default Random Forest mimic model shows different feature importances (\textbf{b}), demonstrating that sensitivity based explainers produce different explanations for different mimics. This undermines the veracity of the audit. By contrast, \RoT (\textbf{c}) does not need any mimic models or access to the proprietary model. Extended results are in Figures \ref{fig:markup_bars_app}, \ref{fig:markup_bars_app_lime}, \ref{fig:markup_violins_app} and \ref{fig:markup_violins_app_lime}.}
	\label{fig:markup_bars}
\end{figure}

\begin{table}[!b]
	\centering
	\caption{ \textbf{Model accuracies for different mimic models.} Amazon's product recommendation system is not publicly accessible, making audits difficult. Different models are trained to mimic Amazon predictions, enabling audits by providing query-able interfaces for explainers. We train Random Forest models with hyperparameters from The Markup's audit or from scikit-learn defaults, and various Logistic Regression models. They all have similar accuracies, forming a Rashomon set of models that make similar predictions.}
	\begin{tabular}{l!{\vrule width 0.8pt}ccccc}
	\toprule
	 & \textbf{RF (Markup)} & \textbf{RF (Default)} & \textbf{LogReg} & \textbf{LogReg (L1)} & \textbf{LogReg (L2)} \\
	\midrule
	\textbf{Test Accuracy} & 0.69 & 0.69 & 0.71 & \textbf{\textit{0.72}} & 0.69 \\
	\bottomrule
	\end{tabular}
\label{table:markup_mimic_accuracies}
\end{table}

When explaining a proprietary AI system, restrictions can prevent access to model weights as well as the ability to query outputs for synthetically perturbed inputs. This usually disqualifies sensitivity based explainers, but a common workaround is to collect input-output pairs from real-world deployments of the AI system, and use this dataset to build a ``mimic'' AI system. With unrestricted access to the mimic, predictions can be obtained on new synthetically-perturbed inputs, allowing typical post-hoc explanations such as those from SHAP or LIME to be computed\cite{surrogate:10.1145/3366424.3383110, arize_surrogate_models, fiddler_surrogate_models, surrogate_azure}. This is distinct from XAI that consists of an interpretable ``glass-box'' surrogate model, which approximates and AI system holistically as a global explanation (different from local feature importance explanations) itself. Building a mimic AI system and then using it to compute post-hoc feature importance explanations is a distinct explanation category that involves two potential sources of error: the overall approximation of the AI system by the mimic AI system, and the local approximation of the mimic using sensitivity analysis, like SHAP.

Mimics can however suffer from the ``Rashomon effect'' where they can make predictions identical to the original AI system, but through different internal mechanisms \cite{rashomon:10206657, rudin2024amazingthingscomehaving, renard2024}, yielding different local sensitivity and consequently different explanations \cite{pmlr-v119-marx20a, xai_rashomon:10.1007/978-3-031-43418-1_28}. In our audit experiment, the SHAP or LIME explanations from TheMarkup's mimic AI (Figure \ref{fig:markup_bars} panel a) reveal the most sensitive input feature to be whether the product brand is Amazon (``brand is amazon''), followed by ``product reviews''. However, switching to a different, equally performant mimic AI (Figure \ref{fig:markup_bars} panel b) alters these feature importances. We find many equally performant (see Table \ref{table:markup_mimic_accuracies}) possible mimics in the Rashomon set of roughly equally performant AI systems, which seem to have different internal decision mechanisms (see Table \ref{table:markup_mimic_importances}).

This `different local behaviour but equivalent performance' casts doubt on the mimic approach. An audit using a particular mimic can be undermined by adversarially selecting a different mimic. Further, AI audit results can be contested because insights from any explanation can be attributed to the choice of mimic rather than the underlying AI system. By contrast, \RoT explanations (Figure \ref{fig:markup_bars} panel c) can be computed without access to the original AI system or the ability to obtain outputs for new inputs. This makes \RoT valuable for enforcing regulations which do not grant access to third-party AI systems or source code. For example under Article 74(12) of the EU AI Act\cite{EUAIA}, regulators are granted access to observe AI outputs on fixed input data, but not the ability to obtain new predictions on arbitrary inputs.

Like other explainers, \RoT also uncovers ``brand is amazon'' to be the most important feature. However, critically, it also shows ``sold by amazon'' to be an important feature, something that remained hidden in preceding mimic based analysis. This is a new insight signalling potentially self-preferencing behaviour where the Amazon platform does not just promote its own brands, but also third party product sold directly by Amazon over the same product sold by third-party merchants. This finding is harder to undermine because the explanation is independent of idiosyncrasies arising from the choice of mimic. As such, \RoT explanations can be useful in identifying \improve{potentially} self-preferencing practices in regulatory contexts such as the EU Digital Markets Act, where establishing causal links is not strictly necessary\cite{DMA}.

\subsection{Scientific Discovery using XAI}
\label{sec:science}

We consider a diabetes prediction scenario where XAI needs to award low feature importance scores to unimportant features, and recidivism and loan approval scenarios where XAI needs to award high importance scores to important features in the presence of misleading alternatives. In both situations, \RoT is good at hypothesis generation because it correctly identifies the unimportant and important input features. These explanations are being sought not to explain the behaviour of a specific AI system, but to understand the underlying data it was trained on. XAI has been advocated as a means of scientific discovery \cite{xai4sci:9007737, xai4chem:Esterhuizen2022, D0NJ02592E, xai4bio:poshap, Garbulowski2021} in diverse fields like medicine\cite{EJIYI2023100166}, engineering\cite{HU2024118022}, chemistry\cite{Fu2024}, and many more \cite{CAMPEDELLI2022101898, NIU2023166662, Cremades2024, Sun2023, WANG2024133707, Cao2022, Cao2024, GOU2024130651}

Upon analyzing 1151 research articles at the intersection of XAI and science, we found 4.9\% of them used explanations of AI systems as a means of proposing novel scientific hypotheses. Of these, 70\% used SHAP. We detail the methodology behind these statistics in Appendix \ref{M_lit_review}. XAI use in scientific discovery suffers the same challenges as AI auditing -- like mimics, AI systems trained to replicate scientific phenomena are not guaranteed to faithfully duplicate underlying causal mechanisms. This can be problematic when using SHAP, because sensitivity analysis generates explanations by making predictions on synthetic, out-of-distribution datapoints, which may include physically implausible values. The behaviour of the AI system on non-realistic datapoints could be anything, and consequently the feature importances could be arbitrary \cite{kumar2020problemsshapleyvaluebasedexplanationsfeature,janzing2019featurerelevancequantificationexplainable,taufiq2023manifoldrestrictedinterventionalshapley}.By contrast, \RoT importances are inspired by feature ``predictiveness'' rather than sensitivity, overcoming these limitations without implicitly introducing harder problems such as learning causal models \cite{Neuberg_2003} or density estimation \cite{vapnik1998statistical}.

\RoT also satisfies leading AI regulations better, and its advantage over explainability methods such as SHAP and LIME can even inform (future) best practices in drug discovery. Specifically, the European Medicines Agency (EMA) and US Food and Drug Administration (FDA) recently published joint guidance noting a set of principles whereby interpretability and explainability are enumerated as best model design and development practices\cite{EMA_FDA_2026_Good_AI_Practice}. The principles do not mandate specific XAI methods, but the guidance makes clear that transparency, reliability and accountability \improve{are key to acceptable deployment}.

\begin{figure}[!b]
	\centering
    \captionsetup[subfloat]{justification=centering}
	\subfloat[\RoT finds age to\\be unimportant]{
		\includegraphics[width=0.32\linewidth]{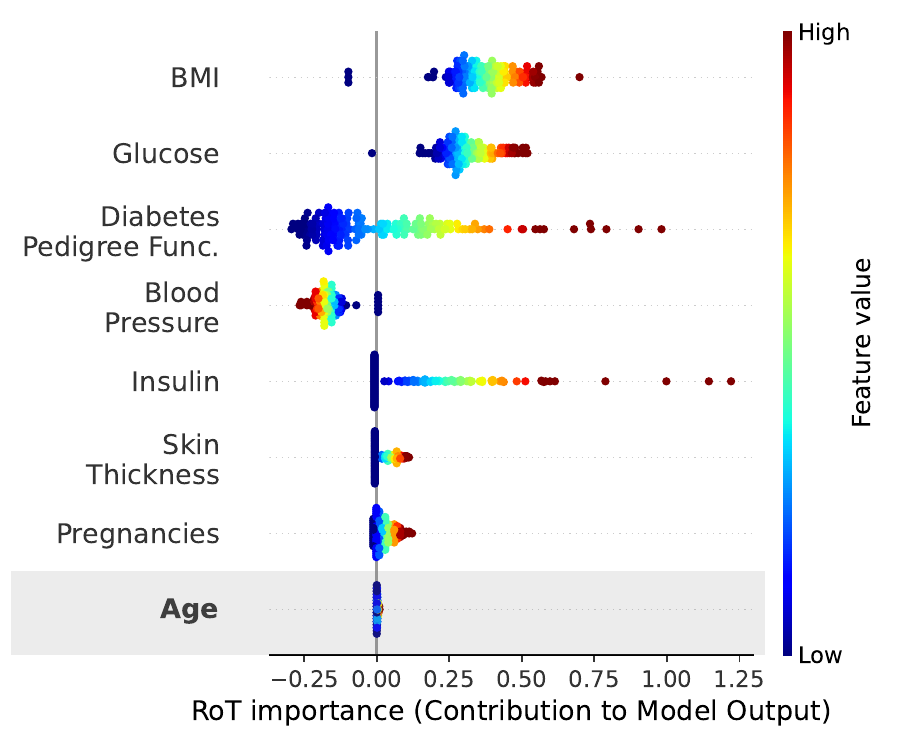}
	}
	\hfill
	\subfloat[SHAP finds age to be\\somewhat important]{
		\includegraphics[width=0.32\linewidth]{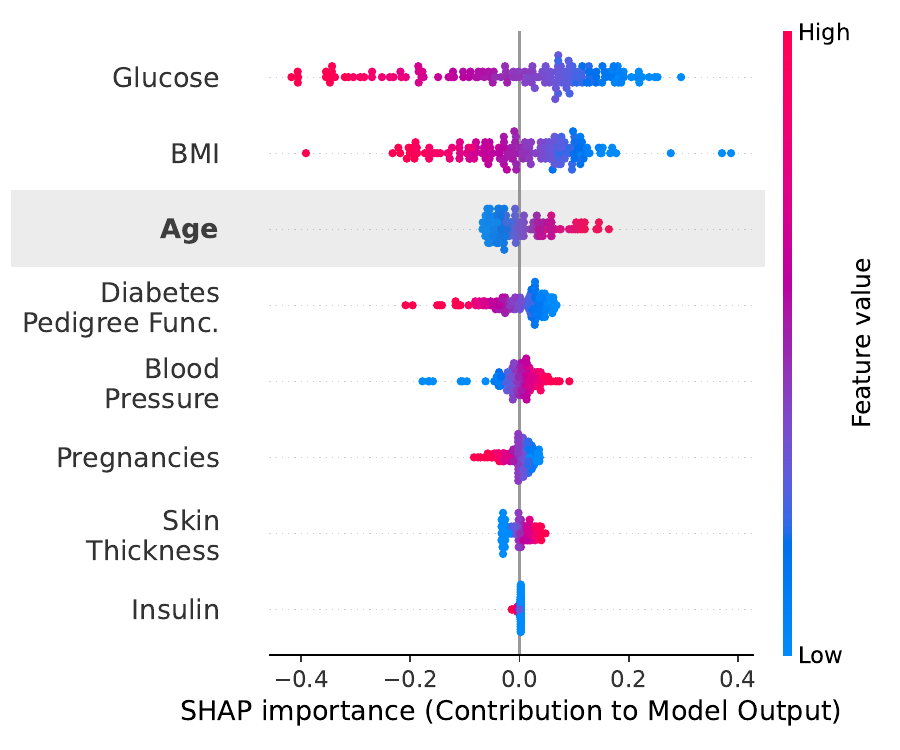}
	}
	\hfill
	\subfloat[LIME finds age to be\\somewhat important]{
		\includegraphics[width=0.32\linewidth]{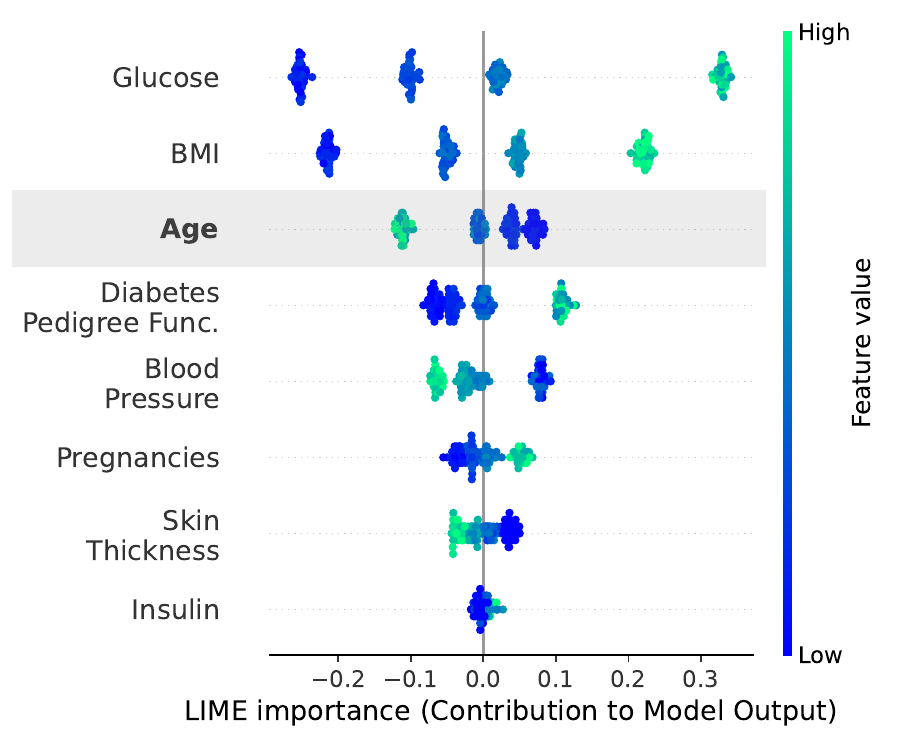}
	}
	\caption{ \textbf{\RoT explanations do not erroneously find unimportant features to be important.} An AI system for diabetes prediction is trained on ``Pima Indians data''\cite{pima} with a synthetic transformation such that age is uncorrelated with diabetes. For this transformed data, XAI should ideally consider age to be unimportant. Swarmplots display feature importances across the dataset. While \RoT correctly ascribes negligible importance to age, sensitivity based explainers SHAP and LIME find it to be somewhat important. When using XAI for scientific discovery, \RoT thus provides clear benefits over SHAP and LIME.}
	\label{fig:swarmplots}
\end{figure}

\subsubsection{Correctly Identifying Uninformative Features}
If an explainer indicates that a specific feature impacts a phenomenon of interest, but that feature is extraneous to the phenomenon, then the explainer is poorly suited for scientific discovery. But sensitivity based explainers \improve{like SHAP and LIME} consider any input feature important if applying any perturbations to it leads to changes in the AI-modeled output phenomenon\cite{kumar2020problemsshapleyvaluebasedexplanationsfeature} , even if the changes are not meaningful. Uninformative features can inadvertently seem important this way.

We consider the same diabetes prediction task as in Figure \ref{fig:explanation_example}, using a dataset specially collected to study the disease \cite{pima}. Using a predictive AI system in such a manner is one way to study diabetes \cite{pima_xai}. However, for our experiment, we manually transform the dataset using FairPCA\cite{fair_pca} to ensure that the ``age'' input feature is statistically uncorrelated with the ``diabetes'' target feature being predicted. Since age has been deliberately rendered uninformative, an XAI technique useful for scientific discovery should correctly identify that it does not affect diabetes.

Figure \ref{fig:swarmplots} displays feature importances from \rot, SHAP, and LIME on transformed diabetes data. \RoT is the only explainer to consider age unimportant. SHAP and LIME both lead to spurious hypotheses that indicate the importance of age in diabetes prediction. However, if our interest is in learning about the laws and relations that govern the data and phenomenon of diabetes, \RoT is the better method because it correctly identifies age as unpredictive. Model details and our use of FairPCA\cite{fair_pca} to ensure statistical independence between age and diabetes are detailed in Appendix \ref{M_fair_pca}.

\begin{figure}[!b]
	\centering
    \captionsetup[subfloat]{justification=centering}
	\subfloat[Distribution of feature importances from SHAP and \RoT explanations across ``sensitive'', ``foil'' or other features for an adversarial model]{
		\includegraphics[width=0.47\linewidth]{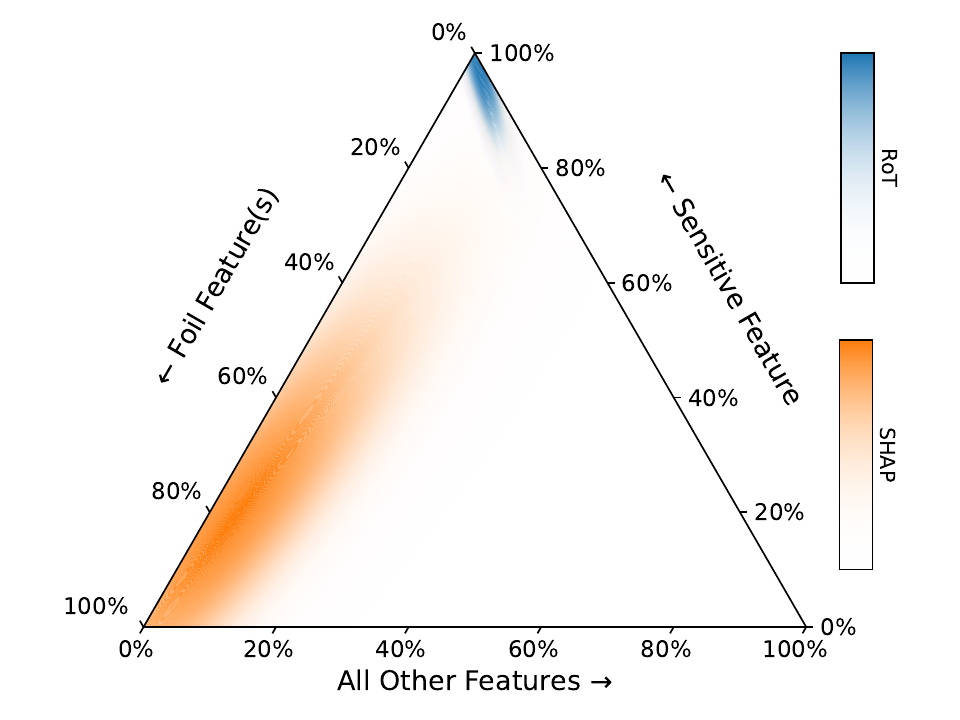}
	}
	\hfill
	\subfloat[Distribution of feature importances from LIME and \RoT explanations across ``sensitive'', ``foil'' or other features for an adversarial model]{
		\includegraphics[width=0.47\linewidth]{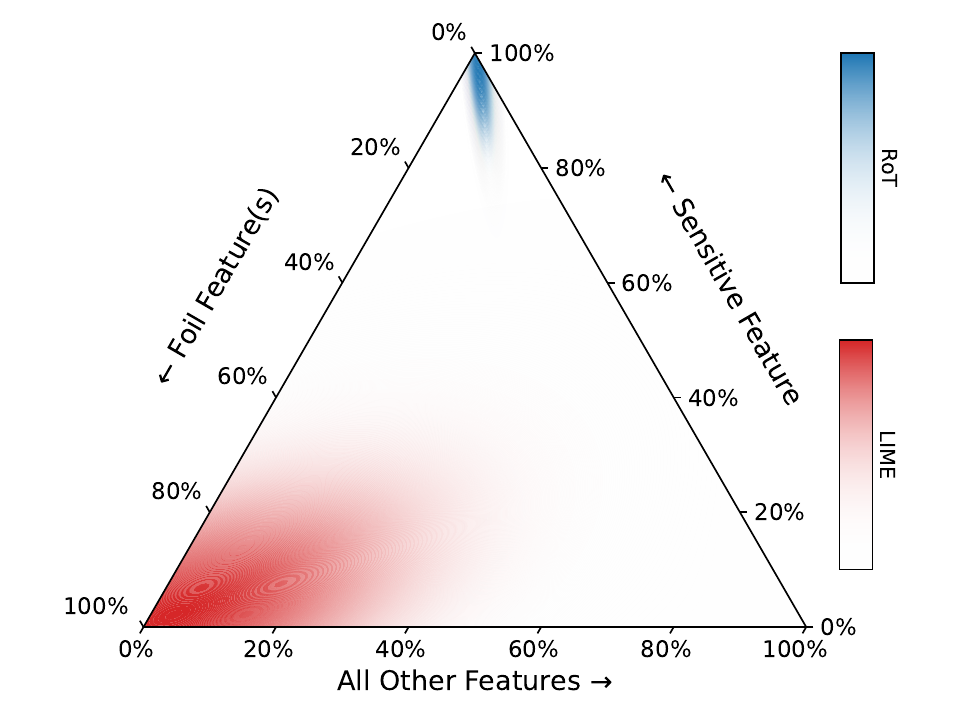}
	}
	\caption{ \textbf{\RoT explanations are immune to manipulation from certain adversarial attacks.} The adversarial attack misleads explainers into ascribing importance to spurious ``foil'' features when the underlying AI system actually makes predictions using a ``sensitive'' demographic feature. We plot the distribution of feature importances across the ``sensitive'', ``foil'', and other features, and find that both SHAP and LIME are misled by the adversary, but \RoT correctly ascribes importance only to the ``sensitive'' feature.}
	\label{fig:advattack}
\end{figure}

\subsubsection{Identifying Important Features in the presence of Misleading Features}
We have seen that explainers can be vulnerable to arbitrary out-of-distribution behaviour from an AI system. Adversarial attacks have been proposed to exploit this vulnerability by modifying AI systems to deliberately poison out-of-distribution outputs \cite{adv_shap:10.1145/3375627.3375830}. This effectively presents different behaviour to SHAP or LIME than to real-world inputs, masking the actual importances of ``sensitive'' features and manipulating explainers into selecting misleading ``foil'' features as highly important instead. In black-box AI audits, adversial attacks cause ``fairwashing'' by allowing malicious actors to deliberately hide AI systems with discriminatory behaviour behind benign-looking explanations \cite{adv_shap:10.1145/3375627.3375830}. For scientific discovery, the same thing can cause candidate hypotheses generated using XAI to be misleading.

To verify that \RoT does not generate such spurious explanations, we replicate adversarial AI systems for credit-scoring and criminal justice, which deliberately hide the importance of an important ``sensitive'' feature by introducing one or more misleading ``foil'' features. We find that this is ineffective at misleading \rot, which is able to correctly identify the sensitive features (race for recidivism prediction, and gender for loan approvals) as important and the foil features (simulated, extraneous features) as unimportant. In Figure \ref{fig:advattack} we show the distribution of SHAP, LIME, and \RoT importances over three feature categories: sensitive, adversarial, and other, from AI systems subjected to an adversarial attack \cite{adv_shap:10.1145/3375627.3375830}. Details about the datasets used, model behaviours introduced, ``sensitive'' input features masked and adversarially misleading ``foil'' features introduced are available in Appendix \ref{M_adv_attack}.

We report detailed statistics about the proportion of explanations that identified the sensitive feature as most important (success), or the foil feature (adversarial failure) or other features (non-adversarial failure) in Table \ref{tab:advattack}.
\improve{Across 10 experiments, for LIME or SHAP the rate of recovering the sensitive feature as most important does not exceed 5\% except once, where adversarial failures (foil feature is most important) still outnumber success (sensitive feature is most important). \RoT has a success rate always greater than 89\%, and in 6 of the 10 experiments \RoT shows a perfect 100\% recovery rate, always correctly identifying the sensitive feature to be the most important.}

\subsection{Computational Efficiency and Environmental Impacts}

Of all the experiments listed so far, generating \rot, SHAP, LIME, and Integrated Gradients explanations for Judicial Case Outcome Prediction with a finetuned LLM was the largest in terms of dataset size and computations involved for prediction. For this experiment, generating the \emph{first} \RoT explanation took 1160 seconds, and each additional explanation took less than 0.1 milliseconds. By contrast, the \emph{average} time taken for each SHAP explanation was 13 million times longer at 1320 seconds. This was without GPU acceleration, to simulate consumer hardware conditions. In practice, it is impractical to compute SHAP or LIME explanations for LLM predictions, except if only a handful of datapoints need explaining. Gradient methods are quicker, but require access to model weights when most commercial LLMs can only be accessed via API. \RoT is the first explainer that can actually be deployed in practice, bringing feature importance XAI to LLMs.

\begin{figure}[!th]
	\centering
    \captionsetup[subfloat]{justification=centering}
    \hfill
	\subfloat[Explanation Runtimes: Judicial Case\\Outcome Prediction with LLMs]{
		\includegraphics[width=0.35\linewidth]{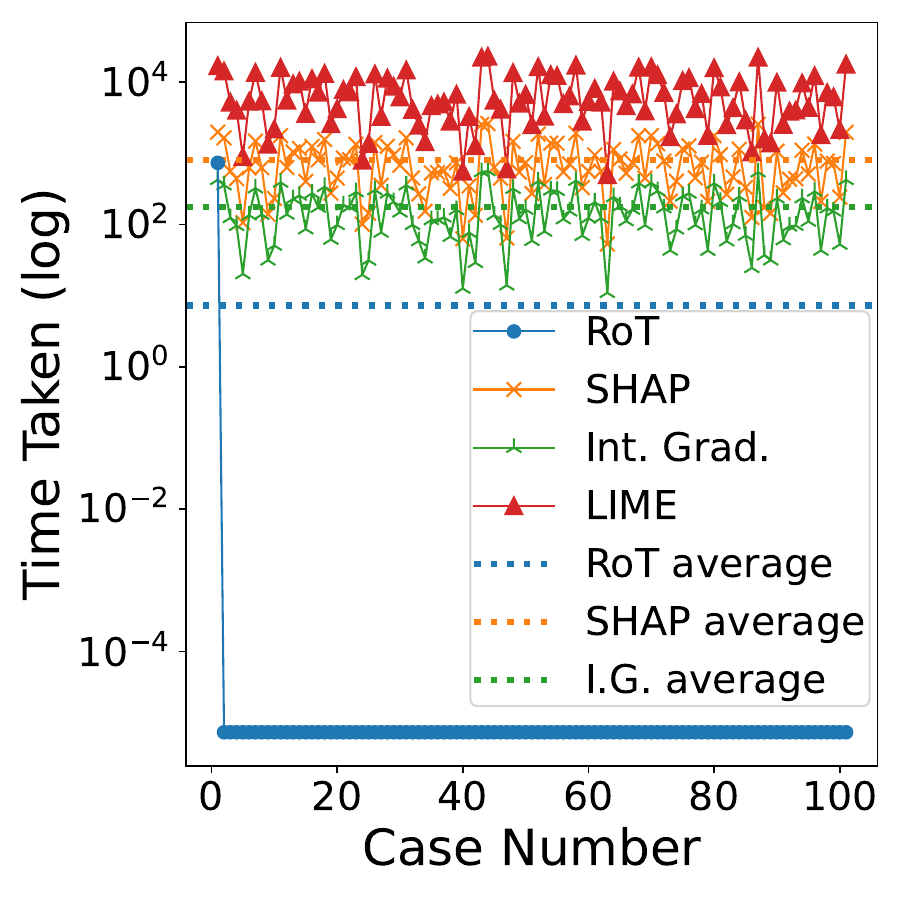}
	}
	\hfill
	\subfloat[Cumulative Explanation Runtimes\\(Quickest First): Case Prediction]{
		\includegraphics[width=0.35\linewidth]{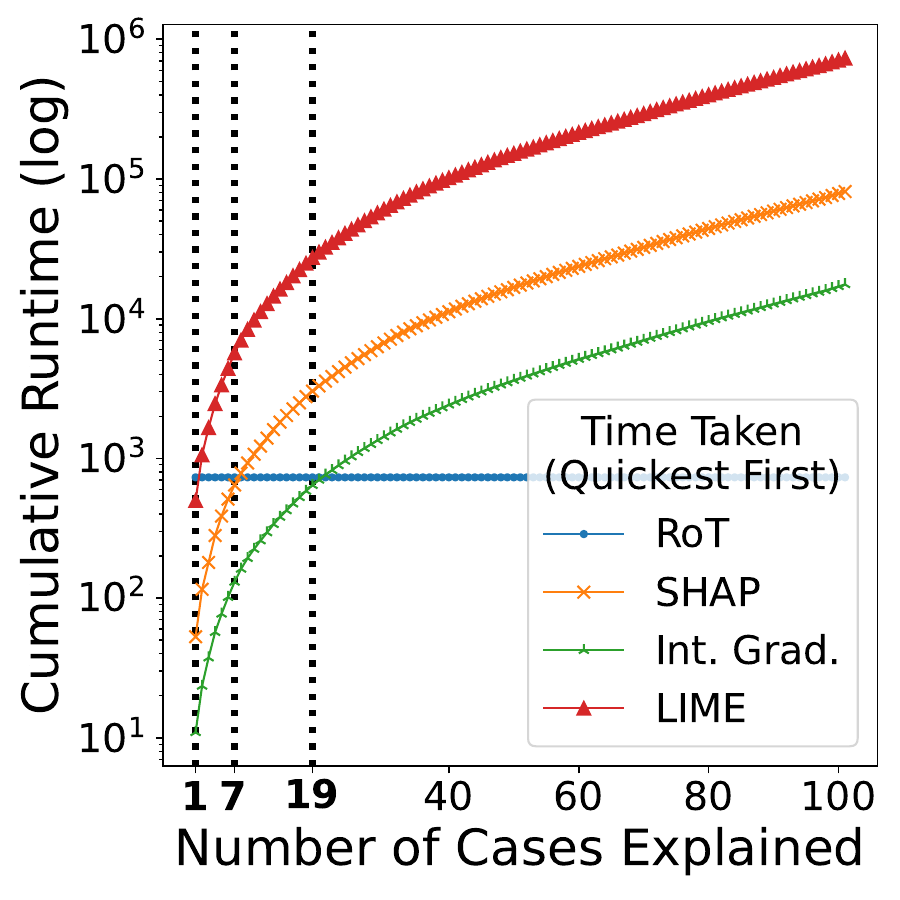}
	}
    \hspace*{0.10\linewidth}
	\caption{ \textbf{Comparing runtimes for \RoT and other explainers from the Judicial Case Outcome Prediction Experiment.} While the first \RoT explanation takes time, additional explanations are virtually instantaneous, unlike SHAP, LIME, or Integrated Gradients. The time-complexity of SHAP thus appears linear in number of cases, while \RoT appears to be constant, and \emph{13 million RoT explanations} can be computed \emph{for each additional SHAP explanation}, after the first. Even the 7 fastest SHAP explanations take as long as all 101 \RoT explanations.
    }
	\label{fig:llm_speedups}
\end{figure}

\RoT was also found to be orders of magnitude faster than SHAP and LIME in our benchmark XAI comparison, detailed in Appendix \ref{M_openxai}, which we document in Tables \ref{tab:speedups_shap} and \ref{tab:speedups_lime}. This increase in computational efficency, evidenced by faster program running times, is a direct result of \rot's unique formulation that does not require obtaining additional predictions for each explanation. Local feature importance explainers often function by collecting data about an AI system's behaviour and then fitting an interpretable model. Both SHAP and LIME need to fit a new interpretable model for each prediction they explain. This requires data collection by making arbitrarily time-consuming calls to the AI system for each additional explanation. By contrast, \RoT uses dropout to fit a single interpretable ensemble on the entire unperturbed dataset, and each explanation is generated using per-feature importance scores from this model. Since this inference is quicker than continuous training for each explanation, this makes \RoT extremely fast once fit. When explaining predictions from large AI systems such as computationally intensive LLMs, the efficiency gains from \RoT would translate directly to reduced environmental impacts.

\section{Regulations Concerning Emerging XAI Applications}

Compared to other XAI methods, \RoT has a range of strengths that make it an ideal tool for making existing AI systems auditable and accountable. Its narrow focus on directly predicting the behaviour of an existing system on real data, gives it robustness against adversarial attacks, and removes the need for mimic models. This  gives it a clear advantage in meeting the EU AI Act’s requirements for accuracy, robustness, and adversarial testing of high-risk AI systems under Article 15 and for robustness and resilience under Article 15 (4) – (5) \cite{EUAIA}. This in turn, supports transparency and instructions of use requirements under Article 13 (1) and (3) (b) (ii). \rot's computational efficiency compared to approaches requiring construction of a mimic model give it clear advantages in meeting environmental sustainability goals \cite{hacker2024sustainable}. Most importantly, it allows auditing proprietary models, overcoming inherent transparency limitations found in many AI regulatory frameworks \cite{wachter2024limitations}.

\RoT enables participatory governance of both narrow and general-purpose AI systems, facilitating ``independent external testing'' and adversarial testing of general-purpose systems with systemic risk, as required under Recital 114; Article 55 (1) (a) of the AI Act \cite{GPAI}. In this regard, \RoT may prove particularly valuable for enforcement of the AI Act. While Article 74(13) allows regulators to access testing datasets and model outputs, it does not always allow full model access or the ability to study model responses on new data, as would be required for typical sensitivity based explainers.

RoT can also be a powerful tool to identify and challenge anti-competitive practices of major technology companies that are increasingly embedding AI systems across their platforms. \RoT can help identify self-preferencing practices, such as search result manipulation, that may be illegal under horizontal competition laws and the EU Digital Markets Act (DMA) that require establishing a link between model behaviour and anti-competitive practices \cite{ECJ2024Google, GeneralCourt2021GoogleShopping}.
We elaborate further on the specific methods and limitations of \RoT in Appendix \ref{M_reg}.

In addition to auditing, explainability and interpretability are also important safeguards for AI use in drug development\cite{1_FDA2025AIRegulatoryDecisionMaking} and related scientific discovery\cite{2_FDA2025UsingAIMLDrugDevelopment, 34_EMA2024ReflectionAI}, with each fulfilling distinct expectations depending on context. For instance, the EMA notes that for the review and monitoring of black box models, explainability metrics, such as SHAP and LIME should be used\cite{34_EMA2024ReflectionAI}. The EMA--FDA joint guidance provides a broad view on the use of AI in evidence generation and monitoring across all phases of a medicine including early research. %\cite{5_EMA2026CommonAIPrinciplesNews}. 
While this recent document does not recommend a specific explainability metric, it aims to promote a baseline on the use of transparent models that are reliable, robust, generalisable and utilises data that is `fit-for-purpose' \cite{EMA_FDA_2026_Good_AI_Practice}. Here we see the added value of \rot, clarifying why SHAP and LIME would not be sufficient for ensuring accountability, transparency and reliability, and providing the more appropriate baseline for future policy guidance and reform.

\section{Conclusion}
This paper presents a new approach to explainable AI based on the predictiveness of individual feature values, rather than their sensitivity to alteration. This \emph{Rule of Thumb} approach allows us to approximate any decision making process by summing the importance response over any subset of feature values. The most useful feature values for a prediction are naturally those with largest importance, and their predictive power overrides the contribution of others. As such, \RoT can be understood as a post-hoc and per-datapoint explainer with more in common with ablation studies than sensitivity analysis.

\RoT addresses several long-standing concerns in the community about the validity and practical utility of XAI, and importantly does so without making additional assumptions. It is flexible and useful in a range of emerging XAI applications areas where other forms of XAI simply can not be used -- evaluating LLMs and explaining their behaviour in zero-shot classification, auditing proprietary AI systems where we can only observe their predictions and not alter input features, and using XAI to understand real-world phenomena and propose novel scientific hypotheses. As a local, post-hoc explanation method, \RoT can be used as a drop in replacement for other explainers in existing XAI codebases, and visualised the same way, providing a fast and computationally efficient (and therefore environmentally friendly) method of explaining AI.

\clearpage
\newpage

\section*{Acknowledgments}

The work of all the authors has been supported through research funding provided by the Wellcome Trust (grant nr 223765/Z/21/Z), Sloan Foundation (grant nr G-2021-16779), Department of Health and Social Care, EPSRC (grant nr EP/Y019393/1), and Luminate Group. Their funding supports the Trustworthiness Auditing for AI project and Governance of Emerging Technologies research programme at the Oxford Internet Institute, University of Oxford. Kai Rawal is funded by an EPSRC doctoral scholarship. The work of Sandra Wachter, Daria Onitiu, Chris Russell, and Kai Rawal has also been supported by the Alexander von Humboldt Foundation in the framework of the Alexander von Humboldt Professorship (Humboldt Professor of Technology and Regulation) endowed by the Federal Ministry of Education and Research via the Hasso Plattner Institute.

\paragraph*{Data and materials availability:} All code used in our experiments can be found at \tturl{https://github.com/KaiRawal/Rule-of-Thumb-Explaining-Artificial-Intelligence-Systems-using-Partial-Information}. The scripts rely on permissively licensed publicly available datasets, and instructions for downloading these, where applicable, are included with the code.

\bibliographystyle{unsrtnat}
\bibliography{references}

\clearpage
\newpage

\appendix

\begin{center}
\section*{Supplementary Information}
\end{center}

\renewcommand{\thefigure}{S\arabic{figure}}
\renewcommand{\thetable}{S\arabic{table}}
\renewcommand{\theequation}{S\arabic{equation}}
\renewcommand{\thepage}{S\arabic{page}}
\setcounter{figure}{0}
\setcounter{table}{0}
\setcounter{equation}{0}
\setcounter{page}{1} % not 0 as \newpage already started a supplementary page

% \section{Materials and Methods}

\section{Motivation and Related Work}
\label{appendix:rel}

\subsection{Explainability in Machine Learning, Feature Attribution Methods, and SHAP}

Machine learning models range from inherently interpretable (eg. decision stumps, linear regressions, rule-sets, or single nearest neighbour classifiers) to uninterpretable black boxes (eg. random forest ensembles, deep neural networks, or markov random fields). In order to explain black-box models, methods have been developed to produce post-hoc model explanations (eg. saliency maps, counterfactual explanations, anchors, prototypes, etc). These can be either global, explaining model behaviour in general (eg. PDPs, Functional Decomposition, Mechanistic Interpretability etc), or local, explaining individual predictions made by a model (eg. ICE, LIME, anchors, counterfactuals). Some model explanation methods are specific to certain kinds of models (eg. TCAV, Saliency Maps, gini-importance), whereas others are considered to be model agnostic (counterfactual explanations, LIME, anchors, etc). SHAP is a widely used model-agnostic, post-hoc, local explanation method, that produces scores indicating the relative importance of various input features that led to a model prediction \cite{SHAP:10.5555/3295222.3295230}. SHAP is often the first technique used by practitioners interested in explaining the behaviour of black box models.

\subsection{Explainability in Experimentation and Production}
\label{appendix:exp_pro}

One use of explainability in machine learning is to continuously explain the predictions being made by artificial intelligence systems that have been deployed in the real world. This can help debug errors \cite{taly2021monitoring}, comply with regulations, and engender trust \cite{xai_book:10.5555/3265834}. Often, explanations are used to gauge if protected attributes are significantly impacting model outputs and bring transparency to otherwise opaque decision making processes \cite{Deck_2024}. When models are being deployed in real world settings and affecting large numbers of users, it can be important to monitor their behaviour and have \textit{explainability in production}, which requires explanations to be computed in a scalable manner.

A different use of explainability is in the lab, where researchers may want to understand model behaviour in order to study underlying phenomena. When using explainability during model development or exploratory data analysis, explanations can be used to understand the data generating process itself and building machine learning models and then using explainability techniques such as SHAP has been proposed as a means of scientific discovery \cite{xai4sci:9007737, xai4chem:Esterhuizen2022, D0NJ02592E, xai4bio:poshap, Garbulowski2021}. This strategy has been employed in diverse fields such as policy-making \cite{CAMPEDELLI2022101898}, 
physics \cite{Cremades2024}, engineering \cite{Sun2023, HU2024118022}, epidemiology \cite{WANG2024133707, Cao2022}, chemistry \cite{Fu2024, Cao2024}, 
healthcare \cite{EJIYI2023100166}, ecology \cite{GOU2024130651}, and geology \cite{NIU2023166662}, research where model explanations are used to derive insights about underlying natural phenomena using a machine learning model, rather than explain the behaviour of a model used in decision making. This can be termed \textit{explainability in experimentation}, and requires explanations to be robust and consistent.

These two use cases for model explanations are different, and present themselves with different desiderata and technical challenges. \RoT provides a scalable method that is important in production settings, as well as a robust method that does not suffer from causal vulnerabilities or the Rashomon effect important in experimentation settings.

\subsection{Implicit Assumptions about Causality}
\label{appendix:causality}
Machine learning classification is concerned with learning the conditional probability of obtaining a response \( y \) from a given set of inputs \( x \). Borrowing from Judea Pearl’s causal hierarchy \cite{Neuberg_2003}, this involves “level 1” reasoning - the ability to observe associations. There is broad consensus today that machine learning alone is unable to produce systems with higher “level 2” or “level 3” reasoning abilities involving the ability to answer “interventional” and “counterfactual” questions, respectively. \cite{Neuberg_2003}

The AI explainability literature so far has been surprisingly forgiving of this deficit typical in all modern machine learning systems. Sensitivity based model explanation methods like SHAP implicitly assume that models produce predictions using level 3 reasoning. If a model only possesses “level 1” observational reasoning, but an explanation method probes it with inputs that do not exist in naturally observed training data, then model predictions in these regions cannot be considered indicative of model behaviours.

SHAP's reliance on perturbing model inputs causes explanations to be overly sensitive to out of distribution model behaviour, which can be arbitrary \cite{kumar2020problemsshapleyvaluebasedexplanationsfeature}. The solutions proposed to this problem such as manifold-SHAP \cite{taufiq2023manifoldrestrictedinterventionalshapley} or causal SHAP \cite{janzing2019featurerelevancequantificationexplainable} necessarily involve solving the harder problems of density estimation \cite{vapnik1998statistical} or causal modeling \cite{Neuberg_2003}, as opposed to learning a simpler classifier decision boundary. Adversarial attacks have been proposed to exploit this vulnerability, and it has been shown that explanations can be manipulated this way, leading to fair-washing. \cite{adv_shap:10.1145/3375627.3375830}

While these two approaches are possible to use in theory, the associated problems with each often make them impossible to implement in practical scenarios. These problems are especially pronounced for uses of model explainability in experimentation, where many models are built and explained iteratively, and the focus is more on the explanations themselves than on deploying the model to real world production settings. In such scenarios, both requiring the user to first build a causal model or estimate the entire density of the data is a non-starter.

By contrast, \RoT makes no causal assumptions about the model, and does not need to solve the ``harder'' problems of density estimation either. \RoT generates explanations entirely through ``level 1'' observation.

\subsection{Ablation Studies and Sensitivity Analysis}
\label{appendix:sens_ablate}
Feature importances can be motivated in at least two distinct ways, involving the effect of:
\begin{enumerate}
    \item removing or ablating input features on model outputs to determine feature importances through their predictiveness \textbf{(ablation studies)}; or
    \item perturbing input features and measuring the corresponding perturbations on model outputs to determine feature importances through their impact on the model output \textbf{(sensitivity analysis)}.
\end{enumerate}
 Both these notions have been used widely. Examples of senstivity analysis include applications in chemistry \cite{sens_chem1:doi:10.1021/cr040659d} \cite{sens_chem2:100183.002223}, geology \cite{MCCUEN197337} \cite{sens_geo:doi:10.1080/13658810802094995} and engineering \cite{sens_engg:10.1137/1031159}. Ablation studies have roots in medicine and are widely used in machine learning \cite{DBLP:journals/corr/abs-1901-08644} today, and sometimes in other fields like physics \cite{Pegourie_1993} \cite{abl_phys:4033064} and biology \cite{ROSENTHAL1999201} \cite{MURRAY199613}.

\begin{table}[!t]
	\centering
    \caption{ \textbf{Situating \RoT among existing local and global XAI.} Among those model explanation methods that use feature predictiveness to determine importance, \RoT is the only local model explanation method. Amongst all local model explanation methods, \RoT is the only method that does not use feature sensitivity to determine feature importance. }
    \label{tab:sens-ablate}
    \begin{tabular}{r|lc|cl}
            & \multicolumn{1}{c}{\makecell{\textbf{Global Explanations for} \\ \textbf{Overall Model Behaviour}}} & & & \multicolumn{1}{c}{\makecell{\textbf{Local Explanations for} \\ \textbf{Individual Predictions}}} \\
\hline
\makecell[c]{\textbf{Feature Sensitivity} \\ \textbf{Determines} \\ \textbf{Feature Importance} \\ \textbf{(eg. Sensitivity Analysis)}}
 &   
\begin{tabular}{cl}
\\
1 & Partial Dependence Plots \\
2 & Accumulated Local Effects \\
3 & \vdots \\
\end{tabular}
& & &
\begin{tabular}{cl}
\\
1 & LIME \& SHAP \\
2 & Counterfactuals \\
3 & \vdots \\
\end{tabular}
\\ 
& & & & \\
\hline
& & & & \\
\makecell[c]{\textbf{Feature Predictiveness} \\ \textbf{Determines} \\ \textbf{Feature Importance} \\ \textbf{(eg. Ablation Studies)}}
& 
\begin{tabular}{cl}
\\
1 & Permutation Feature Importance \\
2 & MDI (for tree ensembles) \\
3 & \vdots \\
\end{tabular}
& & &
\multicolumn{1}{c}{ {\color{red} \huge \RoT} }

\end{tabular}
\end{table}

In existing machine learning explainability literature, there exist global methods motivated by sensitivity analysis, local methods motivated by sensitivty analysis, global methods motivated by ablation studies, but no local methods motivated by ablation studies. \RoT is the first model agnostic post-hoc local explanation method that defines feature importance using the predictiveness of features rather than the sensitivity of the output to perturbations, akin to ablation studies rather than sensitivity analysis. Table \ref{tab:sens-ablate} provides a non-exhaustive list of model explanation methods, classified along different axes: whether the explanations are global or local, and whether feature importance is determined by sensitivity or predictiveness. \RoT is a unique explanation method, as can be seen from the table.

\subsection{Problems with Model Explanations in Practice}
\label{appendix:rot_in_practice}
Returning to our previous examples of trying to debug LLM behaviour by explaining it's classification outputs, and of trying to monitor a large-scale proprietary e-commerce recommendation system, we can illustrate the common practical problems explaining machine learning predictions.

When trying to explain the behaviour of an LLM, it is not possible to arbitrarily modify the input features to the model. For an explanation method like SHAP, it then becomes unclear how to perturb the inputs features. Additionally, explanations can only be used to ascribe importances to input features, and nuanced analyses studying the relationship between the output and external concepts or latent features is difficult. \RoT is useful because not only does it not rely on input perturbations, but also because it allows the inclusion of external concepts when generating model explanations. TCAV is a global explanation method that explicitly uses latent concepts to provide model explanations \cite{pmlr-v80-kim18d}, but \RoT is able to do this on a local, per-prediction level.

When trying to audit the behaviour of proprietary systems, often model artifacts cannot be accessed to obtain output predictions for new input points . This makes explanation methods like SHAP impossible to use without first creating a model to mimic the original. In real world settings, it is often easier to explain such a mimic model, rather than explaining the actual model \cite{fiddler_surrogate_models} \cite{surrogate:10.1145/3366424.3383110}, and it is hard to assert that the mimic explanations generated also apply to the original model due the predictive multiplicity of classifiers \cite{pmlr-v119-marx20a}. This ``Rashomon effect'', where two models have similar predictions but different internal mechanisms \cite{rashomon:10206657}, makes model explanations obtained using a mimic model inherently untrustworthy when explaining the original model behaviour \cite{xai_rashomon:10.1007/978-3-031-43418-1_28}. \RoT sidesteps this issue entirely and does not need access to a model at all, it generates model explanations instead by looking at its predictions.

\section{RoT Implementation Details}
\label{M_rot}
Recall our formulation from Equation \ref{eqn:loss_formulation}. In all experiments, we treat $\ell$ as the log loss, and $F$ as the sigmoid function, arriving at an objective similar to logistic regression. We consider three variants of the importance function $f_{\theta_{j}}(x_j)$.

The linear form
\begin{equation}
f_{\theta_{j}}(x_j)=a_j(x_j+b_j)
\end{equation}
A more general additive form consisting of a linear form plus a mixture of Gaussians
\begin{equation}
f_{\theta_{j}}(x_j)=a_j(x_j+b_j)+\sum_{i<I} S_{i,j} e^{-\frac{(x_j-\mu_{i,j})^2}{\sigma_{i,j}}}
\end{equation}
While we explored the use of other additive models including polynomial kernels and the double exponentiated form proposed in \cite{agarwal2021neural}, the mixture of Gaussians combined with a linear term, showed better stability, and robustness to extreme values.

And finally, for explaining the behaviour of text-based classifiers, such as ChatGPT for zero-shot classification, we use a linear form shared over all tokens
\begin{equation}
f_{\theta_{j}}(x_j)=a\cdot x_j +b
\end{equation}
Here $x_j$ is now a vector representing the text embedding taken from a model such as BERT, and a and b are common parameters shared over all token locations $j\in\mathbb{J}$. Based on this, we provide two canonical implementations of \rot. One uses dropout on the features and a simple linear model to learn classifier logits or regression outputs using either cross entropy or mean squared error loss, and the other uses text embeddings, performing dropout on both embedding features and tokens, to explain zero shot classification tasks.

Our experiments were performed on a MacBook with 24 GB memory and an Apple M4 Pro Processor. All the code used for our experiments will be made public. We also plan to make \RoT available for the community as an open-source software package. We make use of existing implementations of explainers such as SHAP \cite{SHAP:10.5555/3295222.3295230} and LIME \cite{lime}, other open sourced data and models \cite{PredEx,veldanda2023investigating,markup,pima}, and evaluation benchmarks and code \cite{agarwal2022openxai,adv_shap:10.1145/3375627.3375830} previously released by the community.

\subsection{Derivation}
\label{appendix:form}
We wish to explain the behavior of a model \( C(\cdot) \), over an empiric data distribution \( X \). Given a data-point, \( x \in  X \) , we use \( x_j \) to refer to the \( j^\text{th} \) component of the feature vector of \( x \). We define our explanations as a set of functions, similarly indexed by \( j \), and refer to \( f_j(x_j) \) as the importance of \( x_j \). Given an arbitrary subset of features \( J \subseteq \mathbb{J} \) over a data-point \( x \), we wish to be able make predictions about the model response \( C(x) \) using only the sum of importance weights \( \sum_{j \in J} f_j (x_j) \)

In this way, we wish to be able to make a prediction about the model response \( C(x) \) using incomplete information about the feature vector of \( x \), where only \( j \in J \) of the feature values are known. The \( f_j(x_j) \) individually serve as feature importances, and when summed are used to predict the model output.

\subsubsection{Formulating Explanations from Incomplete Information}
We give a general formulation suitable both for modelling discrete decisions using a logistic regression approximation, and for modelling continuous regression tasks using a linear regression approximation. In practice, these can be replaced with any additive model.

The functions \( f_j \) are parameterized by weights \( \theta_j \) and the model is approximated by a function \( F \) of their sum and a global bias term \( G \). Recall that we want to find the optimal set of additive functions \( f \) such that for any data point \( x \in X \) and arbitrary subset of features \( J \) we can approximate \( C(x) \). 

\begin{align}
    C(x) &\approx F \left( \sum_{j \in J} f_{\theta_{j}}(x_j) + G \right) \;\; \forall J \in \mathcal{P}(\mathbb{J})
\end{align}

As mentioned above, we use the linear and logistic approximations of \( C(\cdot) \), in which \( f_{\theta_j}(x_j) = a_jx_j+b_j \). In the case of linear regression \( F \) is the identity function, whereas for logistic regression \( F \) is the logistic function \( \sigma \).

For regression models, this yields:

\begin{align} \label{reg_rot}
    C(x) &\approx \sum_{j \in J} \left( a_jx_j+b_j \right) + G  \;\; \forall J \in \mathcal{P}(\mathbb{J})
\end{align}

and for classification this yields:
\begin{align} \label{class_rot}
    C(x) &\approx \sigma \left( \sum_{j \in J} \left( a_jx_j+b_j \right) + G \right) \; \; \forall J \in \mathcal{P}(\mathbb{J})
\end{align}

For regression models, we minimise the mean squared error, further simplifying equation \ref{reg_rot}. The per feature bias term \( b_j \) can disregarded if the mean of every feature \( j \) is \( 0 \), and the global bias term becomes equal to the mean model response \( \overline{C}(x) \).

For classification models, we maximize the likelihood by marginalising over a distribution of randomly sampled features. We seek the values of \( \theta_j \) that minimize the expected loss \( \mathbb{L} [\cdot,\cdot] \) (typically cross-entropy error) over all subsets of features  \( J \subseteq \mathbb{J} \) and data-points \( x \in X \).

\begin{equation}
    \min \; \; \sum_{ J \subseteq \mathbb{J}} \left( p^{ |J| } \; (1-p)^{(|\mathbb{J}|-|J|)} \; \cdot \; \sum_{x \in X} \mathbb{L} \left[ C(x), \;\; \sigma \left( \sum_{j\in J}f_{\theta_{j}}(x_j) + G \right) \right] \right) \;
\end{equation}

Like the formulation of SHAP, this objective is exponentially large with respect to the feature space. However, a key difference that allows us solve it effectively is that the number of function calls to \( C \) is equal to the number of data points \( |X| \), and completely independent of the feature space \( J \). By commuting this is equivalent to

\begin{equation} \label{final_rot_app}
    \min \; \; \sum_{x \in X} \sum_{ J \subseteq \mathbb{J}} \left( p^{ |J| } \; (1-p)^{(|\mathbb{J}|-|J|)} \; \cdot \; \mathbb{L} \left[ C(x), \;\; \sigma \left( \sum_{j \in J}f_{\theta_{j}}(x_j) + G \right) \right] \; \right)
\end{equation}

Upon careful inspection, this is equivalent to performing dropout with probability \( p \) \cite{JMLR:v15:srivastava14a} on the initial approximation model \( \sigma \left( \sum_{j \in J}f_{\theta_{j}}(x_j) + G \right) \). This allows us to directly find the weights \( \theta_j \) (recall we defined \( f_{\theta_j} = a_jx_j + b_j \) ) and \( G \) using stochastic optimization by treating this as a regression problem regulated by dropout. Implementation of the method is straightforward, with the only subtlety to be aware of is that the weights and biases should be zero-initialized to avoid problems with vanishing gradients in the logistic regression loss.

\subsubsection{Uniqueness and Stability}
Each component of the objective defined in equation \eqref{final_rot_app} is convex. As such, if for every estimated variable there exists a strictly convex component containing that variable, the problem as a whole is strictly convex. Therefore, the overall objective has a unique minimum provided the individual functions \( f_{\theta_j}(\cdot) \) are well-posed and strictly convex with each function having a single minimum associated with it.

\subsubsection{From Approximation Model Weights to Interpretable Feature Attributions}
The optimization procedure returns a set of functions \( f_{\theta_{j}} \) that then need to be applied to each feature. The response \( f_{\theta_{j}}(x_{j}) \) can be directly interpreted as a signed measure of the importance of feature \( j \) on data-point \( x \). For logistic regression, the sum \( \sum_{j \in J}f_{\theta_{j}}(x_{j}) \) can be directly interpreted as probability in logit space of the classifier response \(C(x)\); while for linear regression the same sum is an estimate of the model prediction \(C(x)\).

\section{XAI for Regulation and Auditing}
\label{M_reg}

Enforcement of existing AI regulation often falters due to a lack of evidence, stemming from weak or missing requirements for transparency and model access for regulators, third-party auditors, and researchers \cite{wachter2017right, mitchell2023ai, hartmann2024addressing}. Gathering sufficient evidence of potentially illegal model behaviour is highly difficult without full API access that would allow regulators to feed the model data to evaluate its responses or oversee model outputs holistically \cite{casper2024black, wachter2021fairness, wachter2022theory}. By working without model call access and requiring only observations of prior model outputs or predictions, \RoT can bypass significant barriers to regulatory enforceability created by trade secrecy and confidentiality. \RoT thus democratizes access to explaining and auditing AI systems. This is crucial because strong provisions enabling access or auditing by third parties (e.g., civil society organizations, vetted researchers) are uncommon in AI regulation, with Article 40 of the EU Digital Services Act being a rare exception \cite{wachter2024limitations, EU_DSA}. 

While Article 74(13) grants market surveillance authorities access to the source code of a high-risk AI system under a limited set of circumstances \footnote{One such circumstance being that the testing, auditing and verification procedures and the provider’s documentation have been ``proved insufficient''}, Article 74(12) grants market surveillance authorities access to the training, validation, and testing datasets to carry out investigations into these systems. Importantly, this is limited to dataset access, and normally does not include full model access or enable regulators to feed new data to a model to check its responses \footnote{It is worth noting that full API access may be granted under the AI Act to general-purpose AI models (e.g., LLMs) if specific conditions are met and a request made by the European Commission. Specifically, Article 92 (1), (3) – (4) grants the EU AI Office the power to conduct evaluations of the general-purpose model if the EU Commission requests API access (or source code). However, even in such cases \RoT has clear advantages because it is one of the only computationally viable methods shown to reliably explain zero-shot classification tasks in LLMs at a meaningful scale.}. \RoT offers a clear advantage over other methods in this regard because it can compute accurate explanations using only these datasets, while others require full API access or training a mimic model.

\RoT can also be used to identify anti-competitive practices and prevent self-preferencing. By explaining model predictions and recommendations, \RoT can be used to infer ``how'' a digital service makes certain recommendations or ``why'' it behaves in a certain way. Competition law, however, does not always require establishing causality. A recent case against Google in the United States has shown, for example, that proving the mere existence of anti-competitive conditions is sufficient to show the law has been breached \cite{USDC2024Google}. In this context, even if specific instances of this behaviour in a production environment (e.g., actual transactions with customers) cannot be shown due to a lack of access to model outputs, \RoT can demonstrate that a model displays anti-competitive behaviour. \RoT can help demonstrate that model behaviour is likely to contravene the competition law prohibitions, and thus assist in the enforcement of competition law. Unlike measuring the sensitivity of mimic models, \RoT makes no causal assumptions and thus its ability to predict behaviour of models directly makes it a stronger tool to demonstrate potentially discriminatory behaviour.

\section{Experimental Details}

\subsection{Explaining Zero-Shot Image Classification from the GPT-4-mini API}
\label{M_pet_images}
We use the GPT-4o-mini API from OpenAI to label a balanced subset of 5000 images from a Kaggle dataset of cats and dogs \cite{pet_images_kaggle}. To do this we pass each image to the language model via the API, along with the prompt `Is this a dog or a cat?'. We then extract features from the images using mobilenetv3\cite{mobilenetv3} as a backbone. The image embeddings extracted this way are used as input features for a \RoT classifier, trained using 75\% of the labelled images. Explanations, comprising of positive or negative importances for the mobilenetv3 features, are computed for the 25\% test split, and visualised after per-image-normalisation as overlaid saliency maps. The GPT-4o-mini API has perfect classification accuracy, excluding ambiguous or incorrectly labelled images from the dataset, and the \RoT explainer has a training accuracy of 98.5\% and test accuracy of 97.6\%.

\subsection{Explaining Judicial Case Predictions from a Fine-Tuned RoBERTa Model}
\label{M_indian_cases}

From the published PredEx \cite{PredEx} data, we start by removing all annotations that contain hallucinations or text that we could not reliably match in the raw case text. We drop datapoints with missing case text or missing annotations, and datapoints where no text segments were found to be important or where the entire case text was found to be important. The data was annotated by law students who tried to identify text responsible for the judgement. However, the finetuned model making predictions does not always match the ground-truth predictions. Since the explanation annotations are produced from the ground-truth, but our explainers try to explain the model, there is a discrepancy. To resolve this, we drop all cases where the model predictions were incorrect, since we are interested in explaining the model but comparing our outputs with annotations which were produced independent of the model. From the remaining subset, we further drop those cases where highlighted segments were too ``short'' to be reliable: we enforce the condition that the minimum character length of a highlighted or non-highlighted segment in the text should be at least 75 characters. This threshold is large enough to ensure that common phrases do not get highlighted unnecessarily, and small enough to ensure we still have enough datapoints pass the filter and be included in our experiment.

This finally leaves us with a set of 502 cases, from which we compute SHAP and \RoT explanations on a random 20\% of cases (101 cases total). The \RoT explainer we build uses the remaining 401 cases as background, but results using all 502 cases or only the 101 explanation cases as background are qualitatively similar. For the 401 case corpus, \RoT is only fit on the final RoBERTa prediction per case, and not on the per-segment annotations within the case text.

We report the AUROC of the explanations with respect to the ground truth human annotations (Table \ref{tab:case_aucs}). Computing this metric involves comparing explanation importances (which can be negative) with human annotations (which are a binary 0 or 1). Since the annotations indicate whether text is important without direction (appeal rejection or acceptance), we post-process the feature importance by aligning them with the case prediction manually. That is, if the case appeal was rejected, we consider negative values to denote higher importance by flipping the sign of the importance for all segments, and if the case appeal was accepted, we consider the importance values as is. This way, positive values always indicate higher importance as defined by the annotators. This post-processing is not specific to \RoT and is common to all explainers because it is a peculiarity arising from the PredEx data.

For \RoT explanations, we use a straightforward implementation where dropout is used both on input features and on input tokens. For SHAP, we use the canonical partition explainer implementation (which defaults to 500 background samples) available for use with language models\footnote{this has been made part of the official SHAP package: \tturl{https://shap.readthedocs.io/en/latest/text_examples.html}, but not documented through any publications}. For LIME we use the text explainer with its default bag-of-words tokenizer, with its default 5000 background samples. In addition to abridged example in Figure \ref{fig:good_case_example_377}, program runtimes are documented in Figure \ref{fig:llm_speedups}. The \RoT and SHAP explanations for the unabridged case can be seen in Figures \ref{fig:good_case_example_377_RoT} and \ref{fig:good_case_example_377_SHAP}. Additionally, LIME and Integrated Gradients explanations for this particular case can be seen in Figures \ref{fig:good_case_example_377_LIME500}, \ref{fig:good_case_example_377_LIME5000}, and \ref{fig:good_case_example_377_IG} respectively. Results from other cases can also be analysed via \tturl{https://kairawal.github.io/Rule-of-Thumb-Explaining-Artificial-Intelligence-Systems-using-Partial-Information/JudicialCaseOutcomePrediction/Code/results.html}.

\subsection{Explaining Movie Review Classifications from an LLM API}
\label{M_movie_reviews}

The ERASER\cite{eraser} benchmark provides multiple text datasets with annotations indicating importance of constituent tokens. Most of the tasks however consist of retrieval (from an LLM context window), which can be evaluated using these annotations. The movie review dataset\cite{movies}, however, presents an exception. The task here is simple zero-shot classification and does not involve any explicit retrieval.

We use the GPT-4.1-nano API to obtain predictions for movie reviews (positive or negative), which we then seek to explain using \rot. Our dataset consists of of 1600 movie reviews from the ERASER benchmark dataset \cite{eraser}. To explain these LLM review sentiment classifications we repeat our process from explaining fine-tuned RoBERTa model predictions for judicial cases exactly, producing \RoT explanations for each text segment from the movie reviews and measuring how well they align with human annotations. The notable difference between the two setups is that when explaining judicial cases, we reused the embeddings from the finetuned RoBERTa model itself, whereas now we use ModernBERT \cite{modernbert}.

\begin{table}[!t]
	\centering
    \caption{ \textbf{Area under the precision-recall curve when comparing \RoT explanations with ground-truth annotations.} The underlying datapoints for this table are identical to Table \ref{tab:case_aucs} where we initially presented results from our experiment explaining the juducial case prediction AI system.}
    \label{tab:roberta_prauc}
    \begin{tabular}{l!{\vrule width 0.8pt}ccccc}
	\toprule
	 & \textbf{\rot} & \textbf{SHAP}& \textbf{Integrated Gradients} & \textbf{LIME} (5000 samples) & \textbf{Random} \\
	\midrule
	\textbf{Average PR-AUC} & \textit{\textbf{0.77}} & 0.76 & 0.69 & 0.68 & 0.53 \\
	\bottomrule
	\end{tabular}
\end{table}

\begin{table}[!b]
	\centering
    \caption{\textbf{Area under the Precision-Recall curve: Evaluating explanations by comparing them with ground-truth annotations.} We compare movie review sentiment ground truth labels with \RoT importances to measure the performance in terms of identifying important text segments. While \RoT is better than random, PR-AUC is sensitive to annotation density, and flipping the labels affects the metric significantly. For this reason we choose to use the AUROC metric instead.}
    \label{tab:movie_prauc}
    \begin{tabular}{r!{\vrule width 0.8pt}cc}
	\toprule
	 & \textbf{\rot} & \textbf{Random} \\
	\midrule
	\textbf{Average PR-AUC} & 0.58 & 0.26 \\
	(flipped) \textbf{Average PR-AUC} & 0.95 & 0.92 \\
	\bottomrule
	\end{tabular}
\end{table}

\paragraph{Choice of Evaluation Metric} For both LLM experiments discussed so far, we evaluate explanation quality using the area under the receiver-operating-characteristic (a plot of the true positive rate against the false positive rate). We choose this metric over the area under a precision-recall curve because the precision-recall curve is sensitive to annotation density \cite{prauc}. That is, the precision recall metric is highly sensitive to what proportion of tokens are deemed important for any given text. Since the annotations are booleans, this is dependent entirely on annotator idiosyncracies -- we simply do not have ground-truth importance scores per segment). To illustrate the impact of this, consider the areas under the precision-recall curve reported in the top row of Table \ref{tab:movie_prauc}. If we simply flip all the labels in the ground truth annotation and simultaneously flip the signs of the \RoT importances (the equivalent of finding / deleting unimportant segments instead of finding / selecting important segments from a given review), the metrics change as is visible in the bottom row of the table.

\subsection{Explaining Zero Shot Resume Filtering using Proprietary LLM APIs}
\label{M_resumes}

We replicate a study that found GPT-3 to be largely unbiased in resume filtering tasks \cite{veldanda2023investigating}. Starting with a synthetic corpus of resumes consisting of information-technology workers, accountants, aviation workers, construction workers, chefs, advocates, teachers, and salespeople; we use GPT-4.1-nano to select the IT workers from this corpus. In such a scenario where zero shot classifications are obtained from an LLM API which cannot be queried at will, \RoT is still able to explain predictions.

Just like explaining case outcome prediction and movies, we are able to produce token level importance scores using \rot. This time, we use BERT\cite{bert} embeddings instead of ModernBERT\cite{modernbert}, showing how our method is embedding agnostic and can be used with various text embeddings. We additionally augment explanations with non-input features so predictions are explained not just in terms of the resume, but also in terms of the demographic attributes of the candidates. The original study used race, gender, and political orientation, and we explicitly add this information to our explanation text (even though the predictions were made using only the resume summary). This is an additional ability unique to \rot, and is particularly useful for black-box algorithmic auditing. It allows us to generate feature importance explanations using a superset of the model inputs -- we include race, gender, and political orientation. Our results match the original study, and find that these features have negligible impact on model predictions.

\begin{figure}
	\centering
	\begin{tabular}{cc}
		\subfloat[\RoT importances for ``\textbf{infrastructure}'' across our resume corpus]{
			\includegraphics[width=0.48\textwidth]{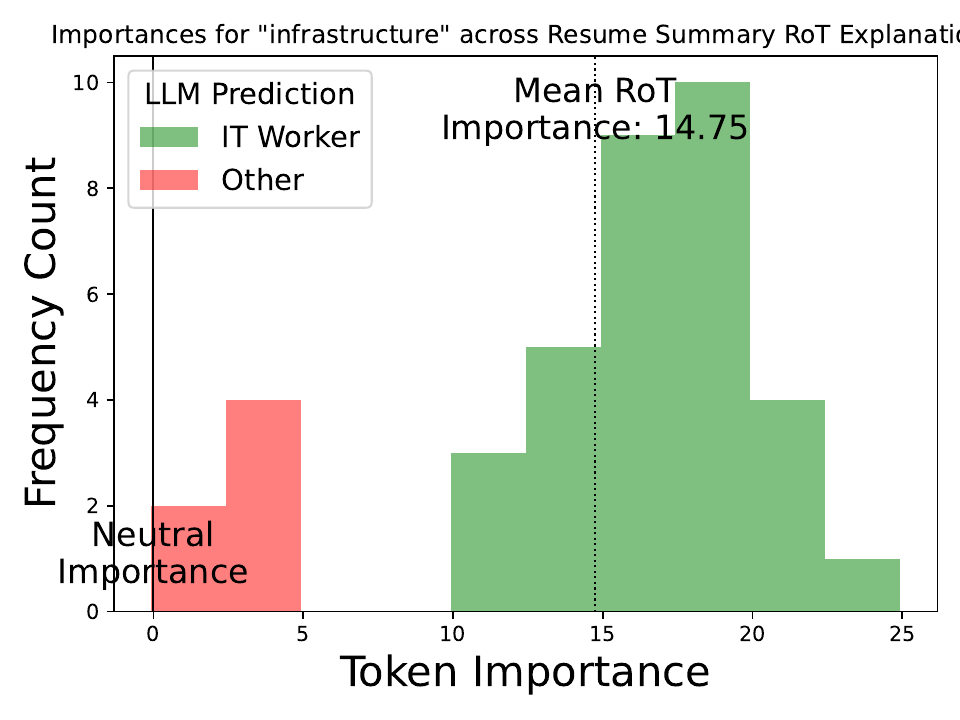}
		} & \subfloat[\RoT importances for ``\textbf{network}'' across our resume corpus]{
			\includegraphics[width=0.48\textwidth]{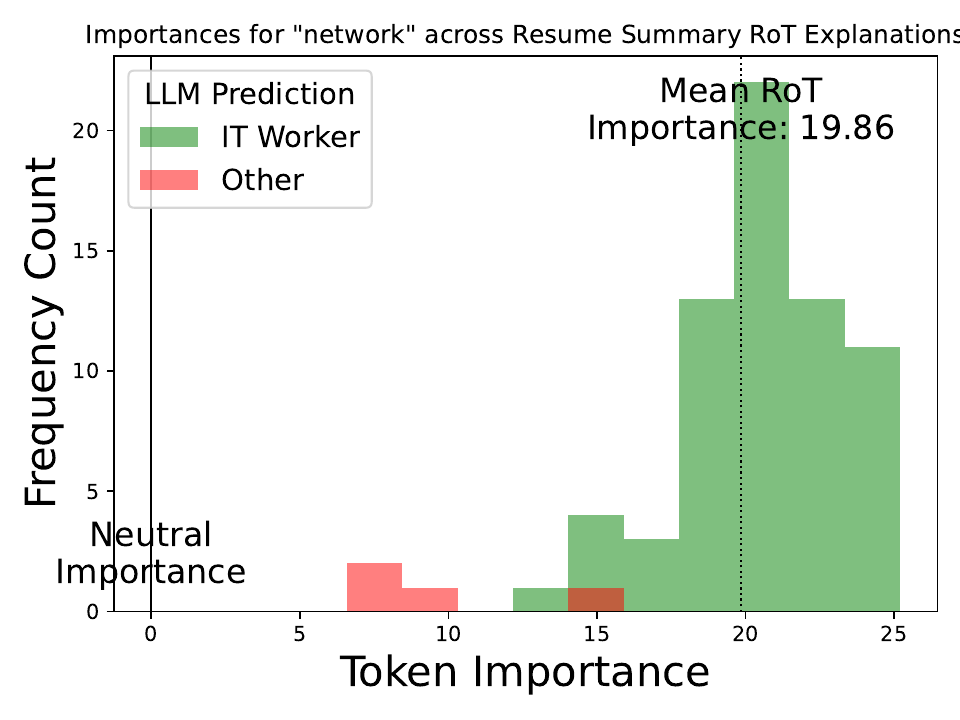}
		} \\
		\subfloat[\RoT importances for ``\textbf{teaching}'' across our resume corpus]{
			\includegraphics[width=0.48\textwidth]{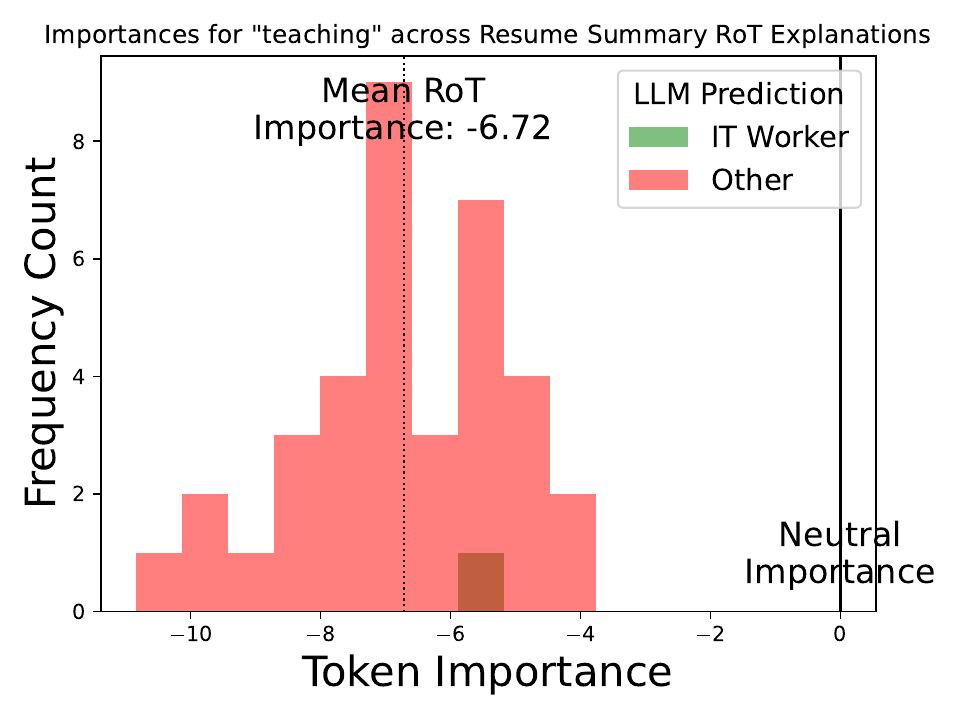}
		} & \subfloat[\RoT importances for ``\textbf{financial}'' across our resume corpus]{
			\includegraphics[width=0.48\textwidth]{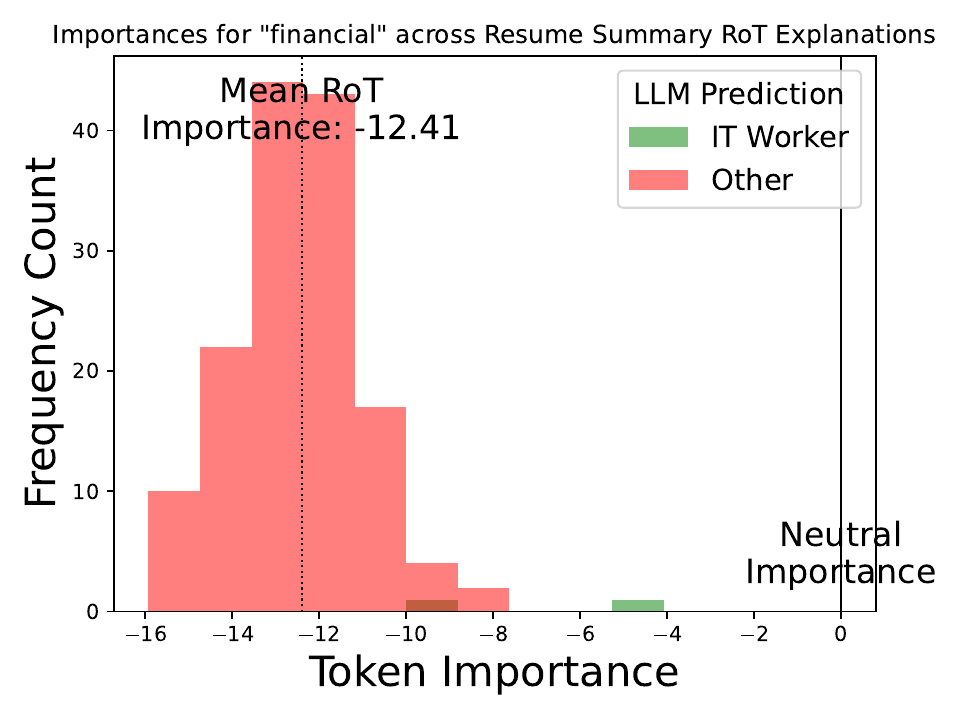}
		} \\
		\subfloat[\RoT importances for ``\textbf{construction}'' across our resume corpus]{
			\includegraphics[width=0.48\textwidth]{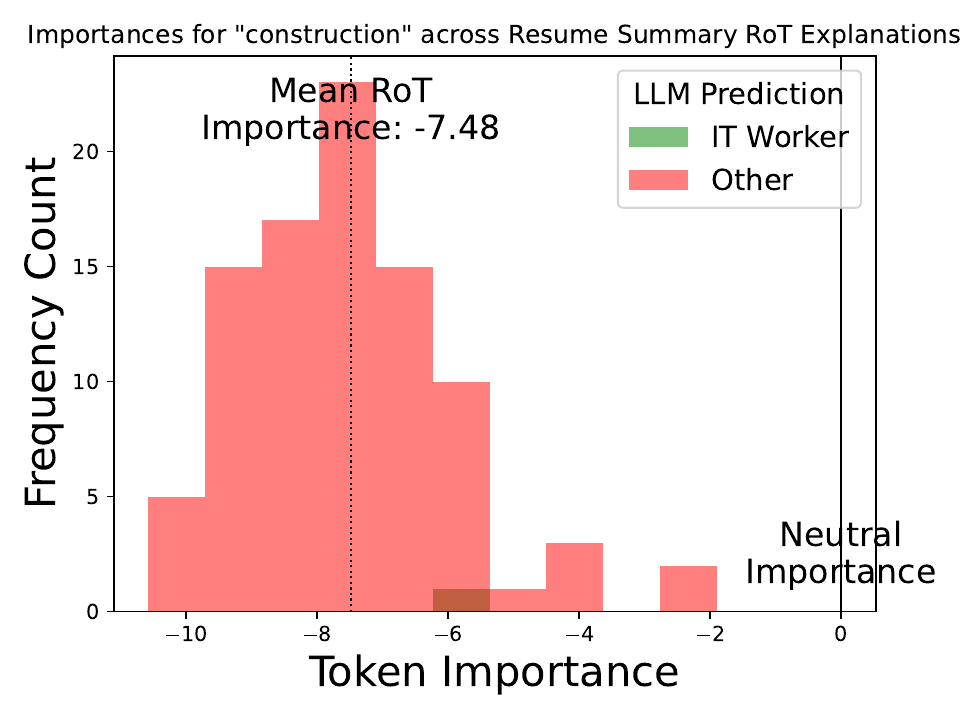}
		} & \subfloat[\RoT importances for ``\textbf{professional}'' across our resume corpus]{
			\includegraphics[width=0.48\textwidth]{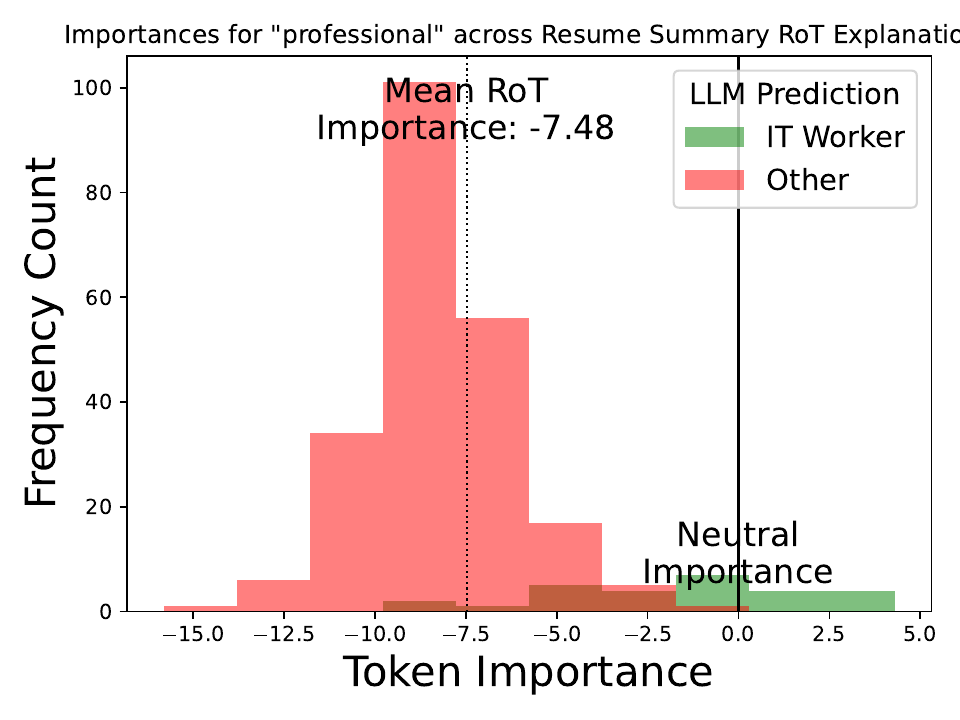}
		} \\
	\end{tabular}
    \caption{ \textbf{Select token importances from zero shot resume classification.} Histograms showing the importances obtained using \RoT for ``\textbf{infrastructure}'', ``\textbf{network}'', ``\textbf{teaching}'', ``\textbf{financial}'', ``\textbf{construction}'', and ``\textbf{professional}'' across resumes. The stacked bar plots are coloured green for token occurrences where the resume belonged to an IT worker, and red for others. Positive numbers on the X axis indicate \RoT importances towards hiring as IT workers, and negative numbers indicate non IT workers.}
    \label{fig:token_hists}
\end{figure}

In addition to the wordcloud in Figure \ref{fig:word_clouds}, we visualise the importance ascribed to a particular token across resume summaries. A cursory examination of these shows that IT related words are positively correlated with IT workers, whereas words belonging to other professions are negatively correlated. Some of these can be seen visualised in Figure \ref{fig:token_hists}.

\subsection{Auditing Proprietary Black-Box AI Systems without Mimic Models}
\label{M_markup}

Journalists at The Markup audited Amazon's product recommendation system. Since the model used is proprietary, it cannot be queried at will, which renders all previous explainers inapplicable. To circumvent this, product recommendation data was scraped from the Amazon website, consisting of product pairs -- the first and second product recommended for any particular search. A model would be built as a ``proxy'' to mimic Amazon product rankings, formulated as a binary classifier selecting one of the products from the scraped pair. Feature inputs to this model were also scraped, consisting of attributes like product stars (the difference in number of stars between the pair of models), and whether the picked model brand was Amazon, was sold by Amazon, shipped by Amazon, etc.

The Markup analysed this data by building a Random Forest model and then computing SHAP explanations of this model\footnote{open-source code from The Markup's reporting: \tturl{https://github.com/the-markup/investigation-amazon-brands}, with both the scraped data and hyperparameter tuning steps}. Following this methodology, they discovered Amazon to systematically promote products from its own brand over others, as SHAP considered ``brand is Amazon'' to be the most important feature, and ``product reviews'' to be second. The model trained and explained was not arbitrary, but chosen after a typical hyperparameter search. When we replicated this audit, we found several models that were (marginally) more accurate, which we list in Table \ref{table:markup_mimic_accuracies}. We found the scikit-learn default Random Forest had better performance than the Random Forest with hyperparameters chosen from The Markup's hyperparameter search. We also found the default Logistic Regression in scikit learn to have better performance, as did both regularised Logistic Regressions (C=0.01).

All our models had similar accuracies and formed a Rashomon set. Their internal mechanisms were distinct, as can be verified from the global explanations summarised in Table \ref{table:markup_mimic_importances}. This variation is also evident from the SHAP explanations for these models, which led to different conclusions about input feature importances depending on the mimic used. All models found ``brand is amazon'' to be the most important input feature, but following this, subsequent features had different importances depending on the model selected. For brevity, we present SHAP explanations from just Random Forest models in Figure \ref{fig:markup_bars}, but include results from all models in Figure \ref{fig:markup_bars_app} and Figure \ref{fig:markup_violins_app}.

\RoT explanations, by contrast, do not require mimic models. Results obtained from \RoT are therefore stronger in an auditing context, because they cannot be attributed to peculiarities in any mimic model, which can otherwise be used to absolve the original inaccessible, proprietary model.

\subsection{Literature Review to Determine the Prevalence of XAI for Scientific Discovery}
\label{M_lit_review}

We measured how often explainability is used as a means of scientific discovery to understand the natural phenomena that underlie models by performing a literature review.

To do this, we analysed papers fetched from the Scopus database related to various scientific fields that used some form of interpretable machine learning. We queried the Scopus API using the \texttt{TITLE-ABS-KEY} field with a search string constructed as follows:
\begin{verbatim}
(
    "explainable artificial intelligence" OR 
    "SHAP" OR 
    "interpretable machine learning"
) AND (
    "social science" OR
    "physics" OR 
    "chemistry" OR 
    "biology" OR 
    "geophysics" OR 
    "astrophysics" OR 
    "biochemistry" OR 
    "medicine" OR 
    "neuroscience" OR 
    "scientific discovery"
)
\end{verbatim}
This yielded 1151 articles, which we filtered using LLMs to analyse the abstracts for mentions of XAI for forming new scientific hypotheses. Using the LLM filter, we reduced our corpus to 111 articles, which we checked manually to verify the use of XAI in each paper. We found that 56 of these papers used XAI, and 39 of the 56 did so using the SHAP explainer. These article lists can be found at:
\begin{itemize}
    % \item \url{***.github.io/RoT/LitReview/scopus_papers.csv},
    % \item \url{***.github.io/RoT/LitReview/llm_filtered_papers.csv}, 
    % \item \url{***.github.io/RoT/LitReview/verified_xai_papers.csv}, and 
    % \item \url{***.github.io/RoT/LitReview/verified_shap_papers.csv}.

    \item \tturl{https://github.com/KaiRawal/Rule-of-Thumb-Explaining-Artificial-Intelligence-Systems-using-Partial-Information/blob/main/LitReview/a.Corpus_1151.csv}
    \item \tturl{https://github.com/KaiRawal/Rule-of-Thumb-Explaining-Artificial-Intelligence-Systems-using-Partial-Information/blob/main/LitReview/b.LLM_filtered_111.csv}
    \item \tturl{https://github.com/KaiRawal/Rule-of-Thumb-Explaining-Artificial-Intelligence-Systems-using-Partial-Information/blob/main/LitReview/c.Manual_XAI_56.csv}
    \item \tturl{https://github.com/KaiRawal/Rule-of-Thumb-Explaining-Artificial-Intelligence-Systems-using-Partial-Information/blob/main/LitReview/d.Manual_SHAP_39.csv}
\end{itemize}

respectively. Our reported statistic of 4.9\% papers using XAI for scientific discovery is thus a lower bound -- there might be papers using XAI that were missed by our LLM filter and never included in set of 111 papers we manually inspected. If the LLM false negative rate is assumed to be the same as the false positive ($55 / 111$) rate, our final result could be as high as 49\%. Our objective however is not to perform a thorough literature review or identify what proportion of papers use SHAP to generate new scientific discoveries. Instead, we merely wish to demonstrate that a significant amount of modern scientific discovery relies on sensitivity based explainers to generate hypotheses that seek to causally explain real-world phenomena.

\subsection{Identifying Uninformative Features as Unimportant for Scientific Discovery}
\label{M_fair_pca}

When using XAI for scientific discovery, we want explanations not to consider uninformative input features as important. We deliberately transform\cite{fair_pca} the ``age'' feature of the ``Pima Indians''\cite{pima} diabetes prediction dataset to be uncorrelated with diabetes, the target feature, and then compute explanations with various explainers to test which ones still find ``age'' to be important.

To make a feature statistically uncorrelated with the target, we adopt the FairPCA\cite{fair_pca} method to transform our dataset. For our dataset with 8 input features, this transformation works by learning a low dimensional space of 7 features, such that age information is deliberately lost. By transforming our data into this space, any model trained cannot utilise age information. Finally, we learn a Logistic Regression classifier on the data, and use standard SHAP, LIME, and \RoT explainers to obtain feature importances. As seen in Figure \ref{fig:swarmplots}, we find sensitivity based explainers unable to identify the unimportance of age, but \RoT does this perfectly.

\subsection{Invulnerability to Adversarial Attacks that Mask Important Features}
\label{M_adv_attack}

We consider a scenario where an adversary tries to mislead explainers into selecting spurious ``foil'' features as important, instead of the ``sensitive'' features the model actually uses to make decisions.

We use the same datasets as those used in the paper proposing the adversarial attack \cite{adv_shap:10.1145/3375627.3375830}, and follow its setup exactly: for the COMPAS \cite{compas} and Communities \& Crime \cite{communities_and_crime} datasets, the ``sensitive'' feature is ``race'' whereas for the German Credit \cite{german_credit} dataset it is ``gender''. All models are binary classifiers formulated as simple if statements depending on the sensitive feature, predicting recidivism risk for the former two datasets, and loan repayment probabilities for the latter.

The ``foil'' attributes used are synthetically added features for COMPAS and Communities and Crime, whereas for the German Credit \cite{german_credit} dataset, the foil used is not synthetic but a new feature that is a function of other input features. Further, instead of 3 experiments (one for each dataset), for COMPAS and Communities and Crime we conduct experiments twice -- once with a single synthetic foil, and once with 2 synthetic foils. The adversary modifies models to mislead explainers. These modifications work can be tailored to the particular perturbations used by a sensitivity based explainer, and we consider models designed to mislead both SHAP ($m_S$) and LIME ($m_L$), comparing both with \rot.

Like in the original paper, this setup leads to 5 experiments with each explainer: 1 German Credit, 2 COMPAS, and 2 Communities and Crime. For each datapoint in each experiment, we consider the absolute importance of the sensitive feature, average absolute importance of the foil features, and average absolute importance of all other features, forming a sample from a ternary distribution. We visualise this distribution of importances in Figure \ref{fig:advattack} using publicly available software \cite{mpltern, matplotlib}, combined from across each of the 5 experiments so each experiment has the same weight. We also present exact numerical results for detailed perusal in Table \ref{tab:advattack}.

\subsection{Benchmark Evaluations of \RoT Explanations with OpenXAI}

We analysed \RoT in traditional tabular modalities using an existing XAI benchmark, since it can be used as a drop-in replacement for other feature-importance explainers. Evaluating explanation quality is a nuanced problem, and we find standard benchmarks unsuitable to measure quality, but are able to observe that \RoT produces similar explanations as SHAP in a faster and more computationally efficient manner.
Sensitivity analysis underlies XAI evaluation benchmarks just like it underlies explainers. The OpenXAI benchmark \cite{agarwal2022openxai} consists of explanation evaluation metrics that either perform sensitivity analysis, or comparisons with ``ground-truth'' explanations.

OpenXAI considers Logistic Regression coefficients as the ``ground-truth'' explanation for all datapoints. Regardless of the specific feature values for a given datapoint, such ground-truth comparison metrics consider a single explanation as ideal across all input datapoints, promoting a single unchanging explanation regardless of specific input values. Using Logistic Regression coefficients OpenXAI deliberately promotes gradient methods or LIME, which are Taylor-series like approximations of AI system behaviour and can recover the Logistic Regression coefficients exactly for any input datapoint. While these explainers score better than SHAP in the published OpenXAI benchmark scores, they are undesirable in practice because they cannot explain individual predictions and always produce identical feature importance outputs. Due to these aforementioned issues, instead of reporting the official metrics from OpenXAI, we adopt the benchmark to verify that \RoT behaviour mirrors SHAP, and every input datapoint does not yield the same explanation output.

\label{M_openxai}

We compute explanations of the pretrained OpenXAI models by running publicly available code from the benchmark\cite{agarwal2022openxai} as provided\footnote{Adopted from \tturl{https://github.com/AI4LIFE-GROUP/OpenXAI/tree/main}}. We present results from the COMPAS dataset because it has the fewest features from among OpenXAI datasets, allowing us to present results succinctly. To select an initial point and a final point for plot (a), we find pairs of datapoints with different model predictions such that $l_0$ distance between them is 2 or less, and $l_1$ distance between them is as large as possible. For plot (b), we consider all pairs where $l_0$ distance is exactly 1 while maximising $l_1$ distances as before, and ensuring the feature changed is always ``length of stay''. While we present select results, we are able to repeat this process for all provided pretrained models and datasets: COMPAS, Home Equity Line of Credit (HELOC), Adult Income, and German Credit. These provide the runtime statistics seen in Tables \ref{tab:speedups_shap} and \ref{tab:speedups_lime}. % and Figure \ref{fig:all_speedups}.

Figure \ref{fig:openxai_bar_scatter} shows how changing input feature values do not always lead to changes in the corresponding feature importances. Only the SHAP and \RoT explainers show desirable behaviour, where the local explanation is truly a function of the datapoint being explained. LIME and gradients show the same feature importance regardless of the feature values, undermining the usefulness of OpenXAI metrics which consider these explanations to be the best. All gradient based methods in OpenXAI generated similar explanations, and here we only show Integrated Gradients for clarity.

\section{Extended Results}

\subsection{Benchmark Evaluations and Computational Efficiency}
\begin{figure}[!bh]
	\centering
    \captionsetup[subfloat]{justification=centering}
	\subfloat[Feature importances for an initial datapoint and for a final datapoint with\\a changed prediction. We expect input feature changes to change the\\corresponding importance, but this happens only for \RoT and SHAP. LIME\\and Gradients are constant despite changes in ``age'' and ``length of stay''.]{
		\includegraphics[width=0.61\linewidth]{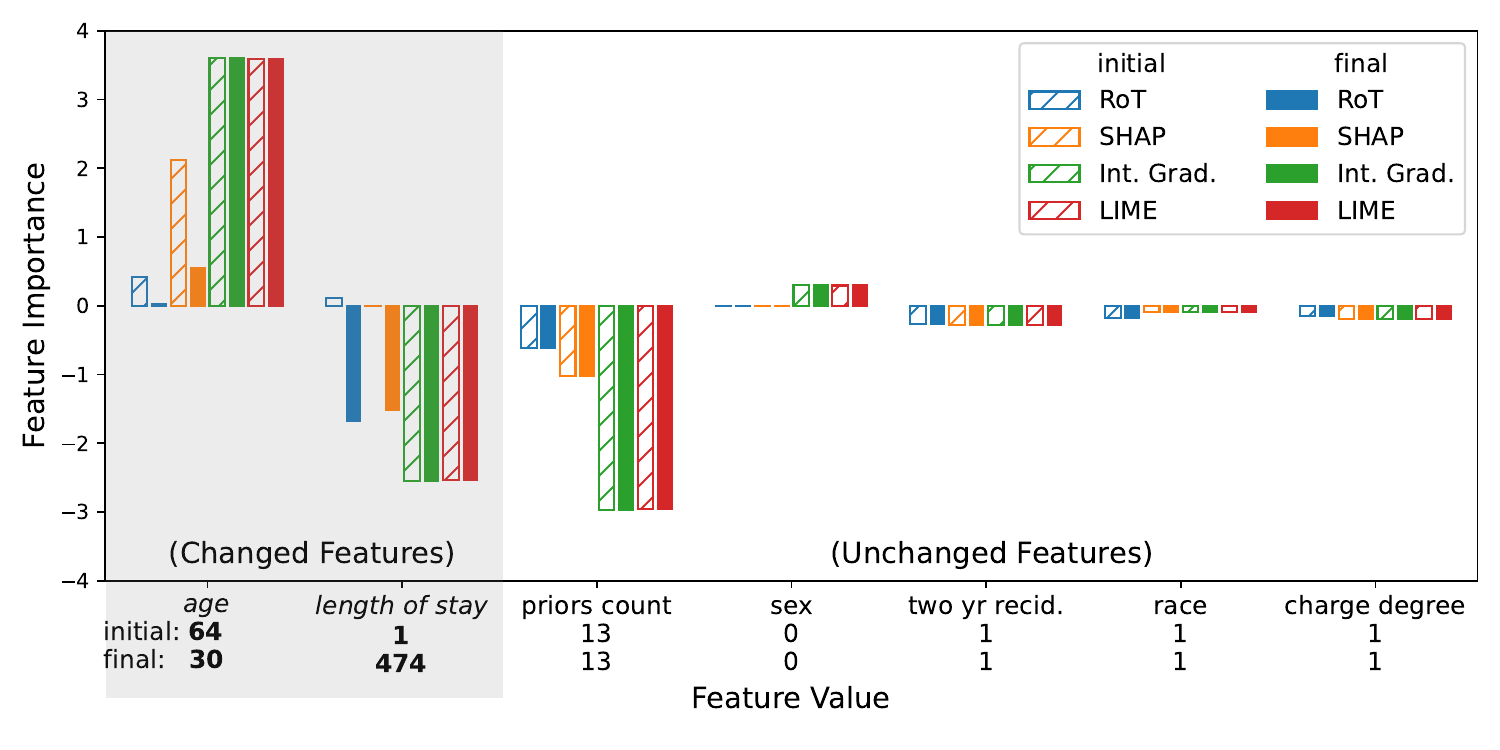}
	}
	\hfill
	\subfloat[Changes in ``length of stay'' feature\\importance for changes in input feature values.\\Like in \textbf{(a)}, \RoT and SHAP change but\\LIME and Gradients are constant]{
		\includegraphics[width=0.36\linewidth]{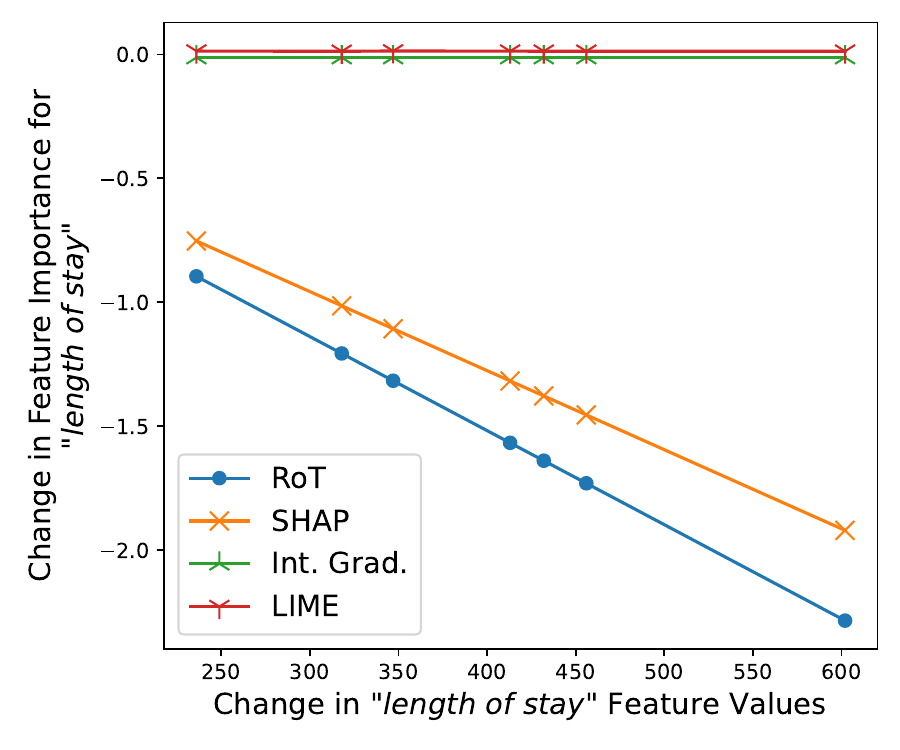}
	}
	\caption{ \textbf{Benchmark comparisons of \RoT with other feature importance explainers.} Analysing explainers using a pretrained binary classifier from the OpenXAI benchmark (COMPAS data). For any initial datapoint, if changing a feature value changes the class predicted by the AI system, that feature should be important. We show all feature importances for one such datapoint \textbf{(a)}, and the changes in one feature's importance across all such datapoints \textbf{(b)}.}
	\label{fig:openxai_bar_scatter}
\end{figure}

We found fundamental problems with the OpenXAI benchmark that prevented its adoption as-is. In Figure \ref{fig:openxai_bar_scatter} we show how \RoT and SHAP explanations for example datapoints in the benchmark are the only ones that actually vary when the prediction being explained varies. The ``ground-truth'' metrics in the OpenXAI leaderboard are known not to score SHAP explanations highly. However, we were able to use the benchmark to measure runtimes across explainers. Tables \ref{tab:speedups_shap} and \ref{tab:speedups_lime} document \RoT runtimes against SHAP and LIME across the OpenXAI benchmark and all our experiments with tabular data.

\begin{table}[!ht]
	\centering
    \small
    \caption{\textbf{Comparison of SHAP and \RoT run-times across experiments.} For each experiment we denote the dataset name and model type (Logistic Regression or Artificial Neural Network). The per-explanation run-time of \RoT is negligible, making it much faster than SHAP despite slower initialization times. This means that \RoT provides a greater speedup for larger datasets with more explanations to compute. }
    \label{tab:speedups_shap}
    \begin{tabular}{clr!{\vrule width 0.8pt}rrrr!{\vrule width 0.8pt}r}
	\toprule
	\thead{} & \multicolumn{2}{c!{\vrule width 0.8pt}}{\thead{\textbf{Data \& Model Details} \\ \textbf{(+ Num. Explanations)}}} & \thead{\textbf{\RoT Time} \\ \textbf{init (s)}} & \thead{\textbf{SHAP Time} \\ \textbf{init (s)}} & \thead{\textbf{\RoT Time} \\ \textbf{total (s)}} &  \thead{\textbf{SHAP Time} \\ \textbf{total (s)}} & \thead{\textbf{Speedup} \\ \textbf{Factor}} \\
	\midrule
	{\thead{ \textbf{AI Auditing} }} & Markup RF & 283 & 0.6358 & 0.0690 & 0.6359 & 295.41 & \textbf{464.54} \\
	\hline
	{\thead{ \textbf{Uninformative} \\ \textbf{Features} }} & \makecell[l]{ Pima Diabetes \\ FairPCA LR } & 154 & 1.0191 & 0.0007 & 1.0191 & 9.63 & \textbf{9.45} \\
	\hline
	\multirow{5}{*}{\thead{\textbf{Adversarial} \\ \textbf{Attack}}}
	& German Credit & 100 & 0.4273 & 0.0017 & 0.4274 & 32.12 & \textbf{75.17} \\
	& CC (1 foil) & 200 & 0.7037 & 0.0017 & 0.7038 & 138.99 & \textbf{197.47} \\
	& CC (2 foils) & 200 & 0.6782 & 0.0022 & 0.6783 & 136.19 & \textbf{200.78} \\
	& COMPAS (1 foil) & 618 & 0.6126 & 0.0021 & 0.6127 & 69.86 & \textbf{114.03} \\
	& COMPAS (2 foils) & 618 & 0.6091 & 0.0057 & 0.6092 & 68.43 & \textbf{112.32} \\
	\hline
	\multirow{8}{*}{\thead{\textbf{OpenXAI} \\ \textbf{Benchmark}}}
	& German Credit LR & 800 & 0.2621 & 0.0013 & 0.2768 & 233.55 & \textbf{843.88} \\
	& German credit ANN & 800 & 0.2596 & 0.0009 & 0.2741 & 169.50 & \textbf{618.49} \\
	& COMPAS LR & 4937 & 0.3344 & 0.0007 & 0.4127 & 189.93 & \textbf{460.23} \\
	& COMPAS ANN & 4937 & 0.3243 & 0.0011 & 0.4049 & 217.48 & \textbf{537.14} \\
	& HELOC LR & 7896 & 0.6619 & 0.0019 & 0.7843 & 522.34 & \textbf{666.00} \\
	& HELOC ANN & 7896 & 0.6473 & 0.0010 & 0.7733 & 413.89 & \textbf{535.20} \\
	& Adult Income LR & 36177 & 2.1747 & 0.0015 & 2.7339 & 1437.98 & \textbf{525.98} \\
	& Adult Income ANN & 36177 & 2.0596 & 0.0006 & 2.6237 & 1668.77 & \textbf{636.04} \\
	\midrule
	\thead{\textbf{Final Average}} & - & \textbf{6786.2} & \textbf{0.7606} & \textbf{0.0061} & \textbf{0.8647} & \textbf{373.60} & \textbf{432.08} \\
	\bottomrule
	\end{tabular}
\end{table}
\clearpage

\begin{table}[!h]
	\centering
    \small
    \caption{\textbf{Comparison of LIME and \RoT run-times across experiments.} For each experiment we denote the dataset name and model type (Logistic Regression or Artificial Neural Network). The OpenXAI LIME implementation differs from the canonical by reducing background samples used per explanation from 5000 to 1000. In order to make up for this inconsistency we report extrapolated OpenXAI LIME runtimes here by multiplying with 5. Like with SHAP, \RoT provides a greater speedup for larger datasets with more explanations to compute.}
    \label{tab:speedups_lime}
    \begin{tabular}{clr!{\vrule width 0.8pt}rrrr!{\vrule width 0.8pt}r}
	\toprule
	\thead{} & \multicolumn{2}{c!{\vrule width 0.8pt}}{\thead{\textbf{Data \& Model Details} \\ \textbf{(+ Num. Explanations)}}} & \thead{\textbf{\RoT Time} \\ \textbf{init (s)}} & \thead{\textbf{LIME Time} \\ \textbf{init (s)}} & \thead{\textbf{\RoT Time} \\ \textbf{total (s)}} &  \thead{\textbf{LIME Time} \\ \textbf{total (s)}} & \thead{\textbf{Speedup} \\ \textbf{Factor}} \\
	\midrule
	{\thead{ \textbf{AI Auditing} }} & Markup & 283 & 0.6358 & 0.0025 & 0.6359 & 21.95 & \textbf{34.52} \\
	\hline
	{\thead{ \textbf{Uninformative} \\ \textbf{Features} }} & \makecell[l]{ Pima Diabetes \\ FairPCA LR } & 154 & 1.0191 & 0.0020 & 1.0191 & 1.40 & \textbf{1.37} \\
	\hline
	\multirow{5}{*}{\thead{\textbf{Adversarial} \\ \textbf{Attack}}}
	& German Credit & 100 & 0.6934 & 0.0012 & 0.6935 & 2.39 & \textbf{3.45} \\
	& CC (1 foil) & 200 & 1.0922 & 0.0008 & 1.0923 & 86.02 & \textbf{78.75} \\
	& CC (2 foils) & 200 & 0.9712 & 0.0008 & 0.9713 & 84.87 & \textbf{87.37} \\
	& COMPAS (1 foil) & 618 & 0.9836 & 0.0025 & 0.9838 & 19.95 & \textbf{20.28} \\
	& COMPAS (2 foils) & 618 & 0.6865 & 0.0025 & 0.6866 & 20.87 & \textbf{30.40} \\
	\hline
	\multirow{8}{*}{\thead{\textbf{OpenXAI} \\ \textbf{Benchmark}}}
	& German Credit LR & 800 & 0.2621 & 0.0003 & 0.2768 & 1071.57 & \textbf{3871.97} \\
	& German Credit ANN & 800 & 0.2596 & 0.0004 & 0.2741 & 1073.84 & \textbf{3918.41} \\
	& COMPAS LR & 4937 & 0.3344 & 0.0004 & 0.4127 & 246.86 & \textbf{598.18} \\
	& COMPAS ANN & 4937 & 0.3243 & 0.0017 & 0.4049 & 420.53 & \textbf{1038.64} \\
	& HELOC LR & 7896 & 0.6619 & 0.0006 & 0.7843 & 2309.85 & \textbf{2945.13} \\
	& HELOC ANN & 7896 & 0.6473 & 0.0063 & 0.7733 & 3991.78 & \textbf{5161.80} \\
	& Adult Income LR & 36177 & 2.1747 & 0.0012 & 2.7339 & 2057.12 & \textbf{752.44} \\
	& Adult Income ANN & 36177 & 2.0596 & 0.0015 & 2.6237 & 7396.13 & \textbf{2818.98} \\
	\midrule
	\thead{\textbf{Final Average}} & - & \textbf{6786.2} & \textbf{0.8537} & \textbf{0.0016} & \textbf{0.9577} & \textbf{1253.67} & \textbf{1308.99} \\
	\bottomrule
	\end{tabular}
\end{table}
% \clearpage

\subsection{Evaluating Zero-Shot LLM Classifier Models with XAI}
\label{subsubsec:llm}
\begin{figure}[!bh]
	\centering
	\includegraphics[width=0.9\linewidth]{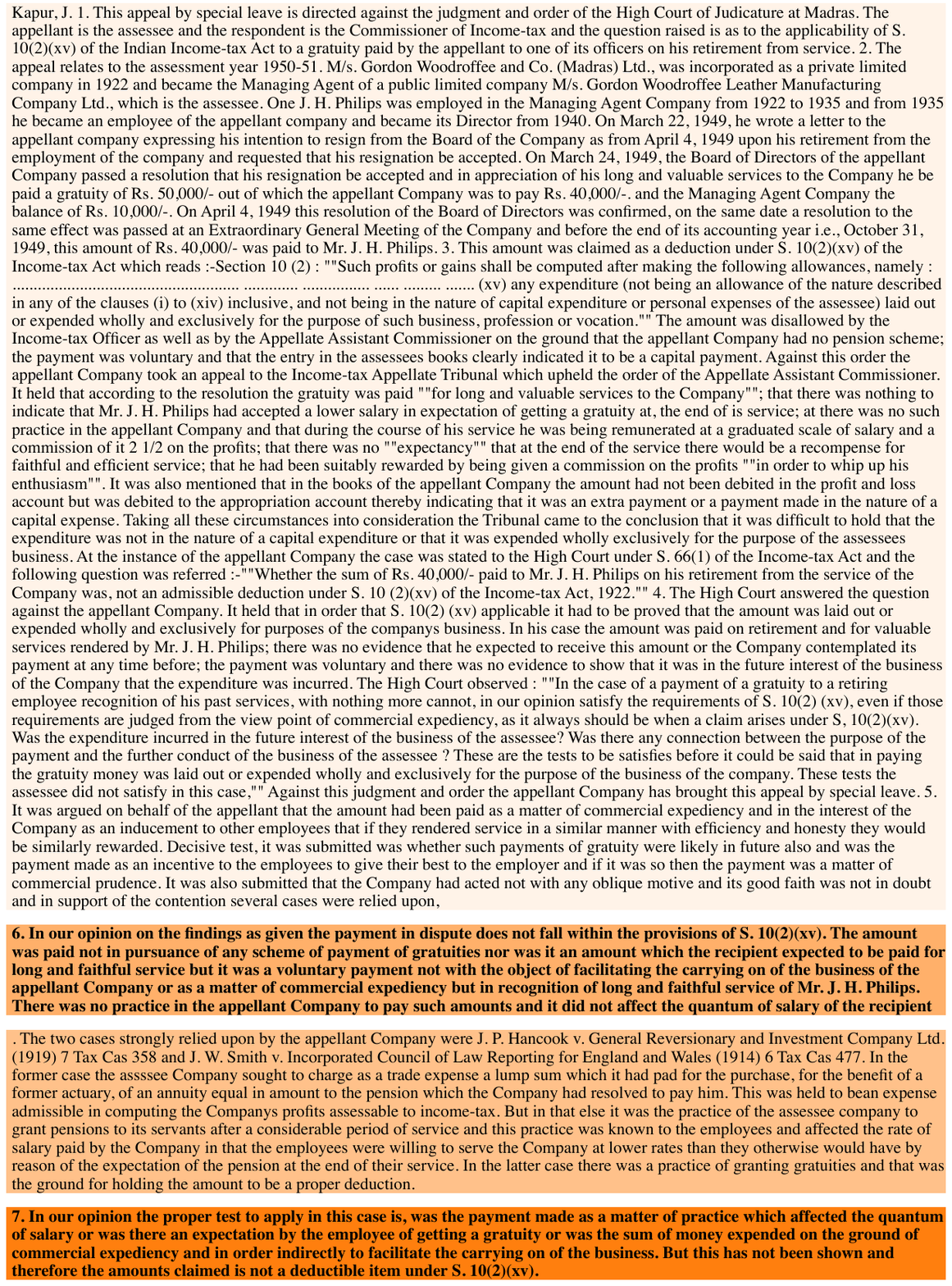}
    \caption{\textbf{Unabridged \RoT explanation for the case from Figure \ref{fig:good_case_example_377}, where the appeal was rejected.} The ``ground-truth'' importance annotation is indicated using a bold font, and the highlights are produced using \rot. \RoT aligns well with the annotation.}
    \label{fig:good_case_example_377_RoT}
\end{figure}

\begin{figure}
	\centering
	\includegraphics[width=0.9\linewidth]{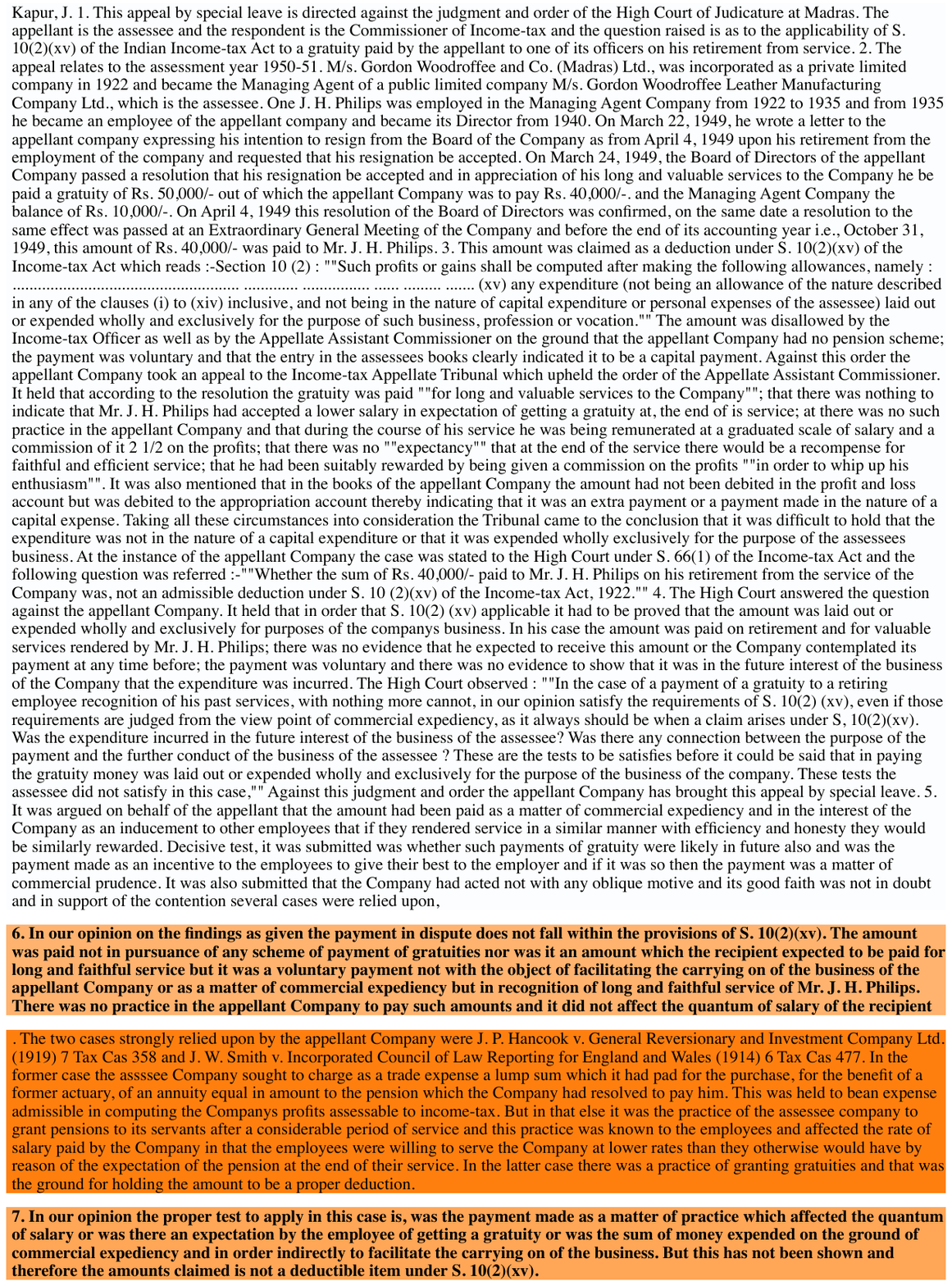}
    \caption{\textbf{Unabridged SHAP explanation for the case from Figure \ref{fig:good_case_example_377}, where the appeal was rejected.} The ``ground-truth'' importance annotation is indicated using a bold font, and the highlights are produced using SHAP. SHAP does not align with human annotation as well as \rot.}
    \label{fig:good_case_example_377_SHAP}
\end{figure}

\begin{figure}
	\centering
	\includegraphics[width=0.9\linewidth]{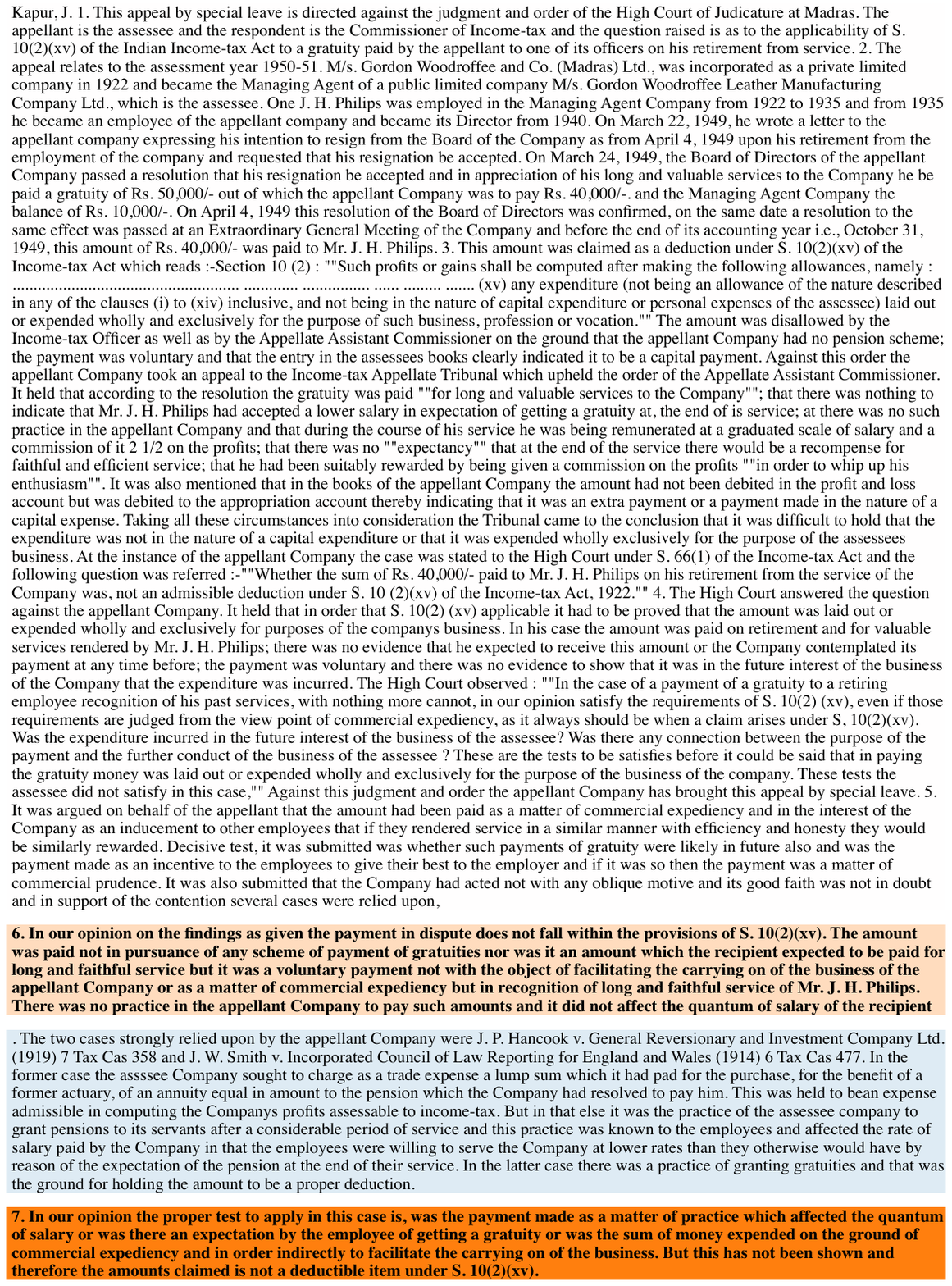}
    \caption{\textbf{Unabridged Integrated Gradients explanation for the case from Figure 2, where the appeal
was rejected.} The “ground-truth” importance annotation is indicated using a bold font, and the
highlights are produced using Integrated Gradients.
    }
    \label{fig:good_case_example_377_IG}
\end{figure}

\begin{figure}
	\centering
	\includegraphics[width=0.9\linewidth]{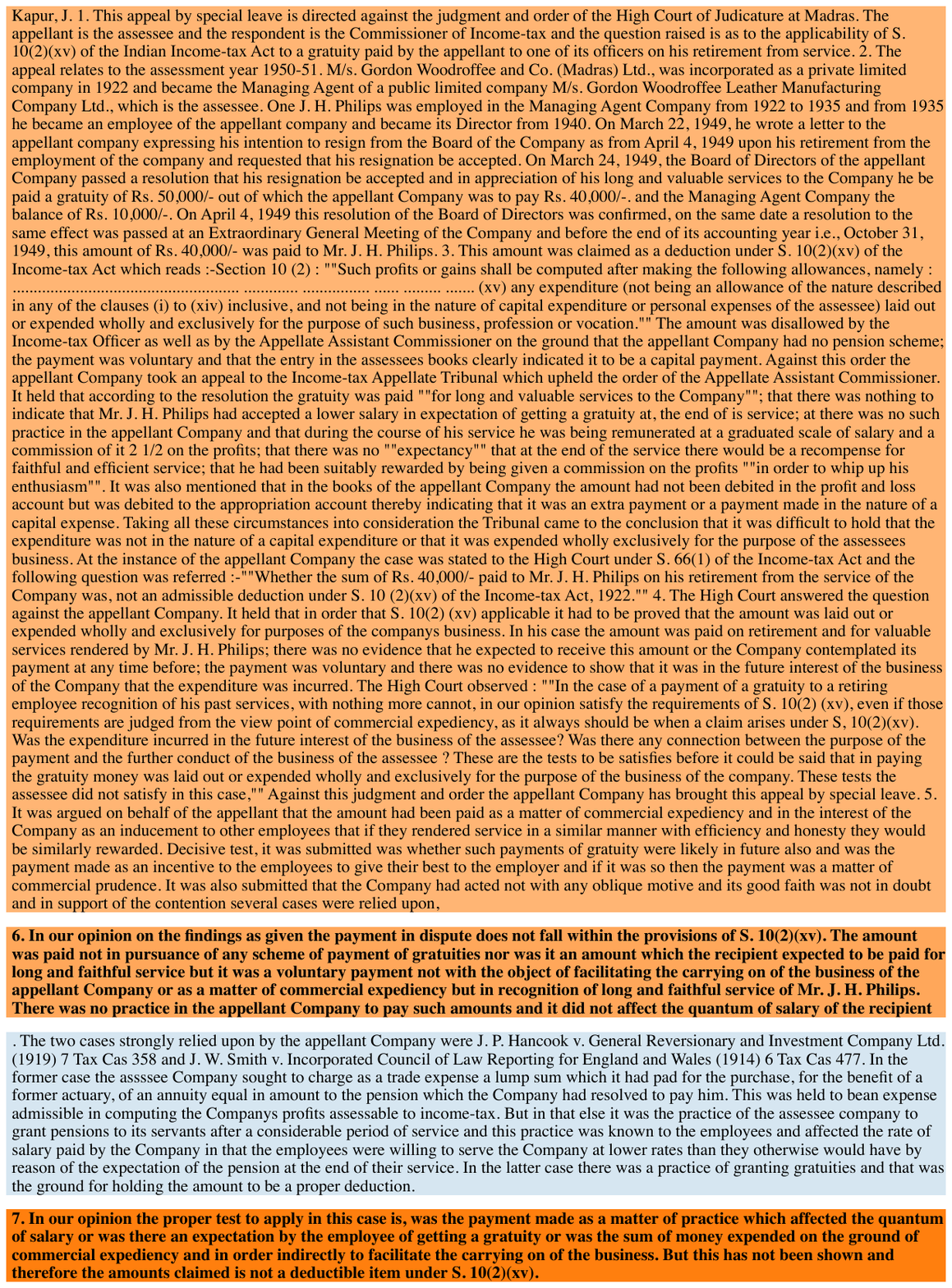}
    \caption{\textbf{Unabridged LIME explanation for the case from Figure 2, where the appeal
was rejected.} The “ground-truth” importance annotation is indicated using a bold font, and the highlights are produced using LIME with 500 background samples (reduced from 5000).}
    \label{fig:good_case_example_377_LIME500}
\end{figure}

\begin{figure}
	\centering
	\includegraphics[width=0.9\linewidth]{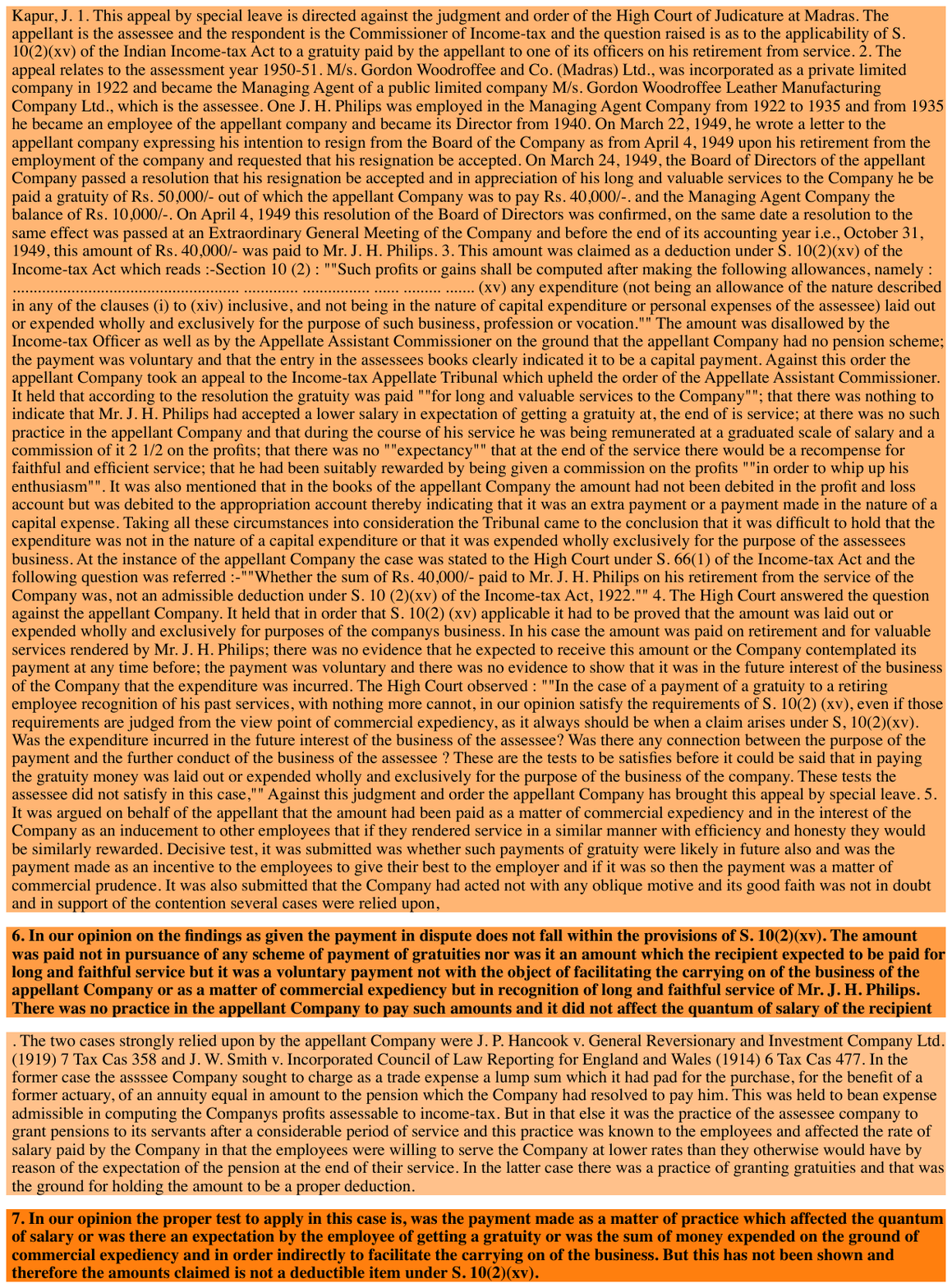}
    \caption{\textbf{Unabridged LIME explanation for the case from Figure 2, where the appeal
was rejected.} The “ground-truth” importance annotation is indicated using a bold font, and the highlights are produced using LIME with 5000 background samples (default configuration).}
    \label{fig:good_case_example_377_LIME5000}
\end{figure}

Unabridged examples of the case prediction explanation from Figure \ref{fig:good_case_example_377} using \RoT (Figure \ref{fig:good_case_example_377_RoT}), SHAP (Figure \ref{fig:good_case_example_377_SHAP}), Integrated Gradients (Figure \ref{fig:good_case_example_377_IG}), LIME reduced to 500 background samples to reduce program runtime (Figure \ref{fig:good_case_example_377_LIME500}), and LIME with 5000 default background samples (Figure \ref{fig:good_case_example_377_LIME5000}).

\clearpage

\clearpage

\clearpage

\clearpage

\clearpage

\subsection{Auditing Proprietary Black-Box AI Systems without Mimic Models}
\label{subsubsec:markup}
\paragraph{Replicating The Markup's audit of Amazon's Product Recommendation Model}

\begin{figure}
	\centering
	\begin{tabular}{cc}
		\subfloat[The Markup model finds the most important feature to usually be either ``brand is amazon'' or ``product reviews'']{
			\includegraphics[width=0.42\textwidth]{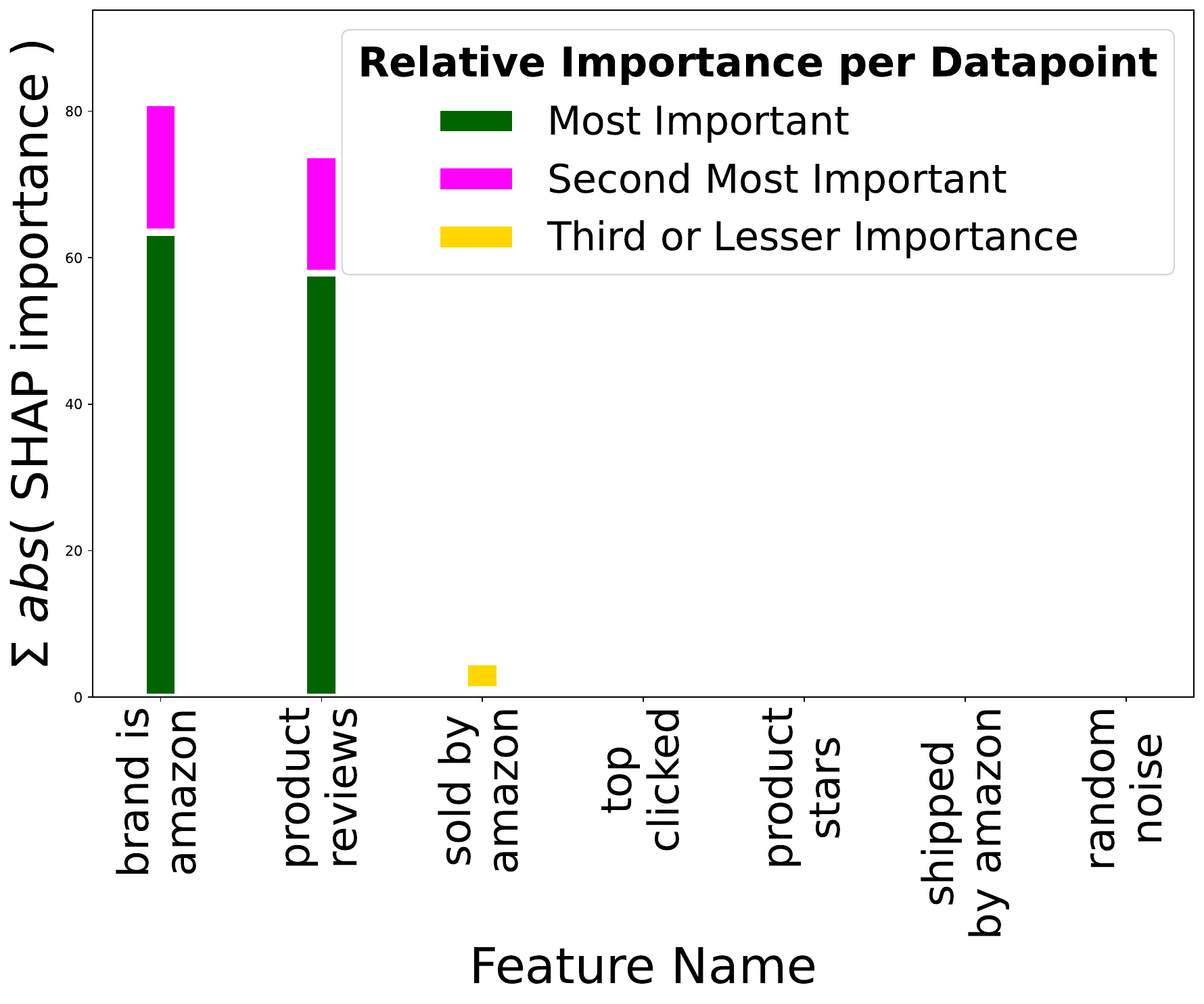}
		} & \subfloat[The default random forest model picks a more diverse set of features as most important]{
			\includegraphics[width=0.42\textwidth]{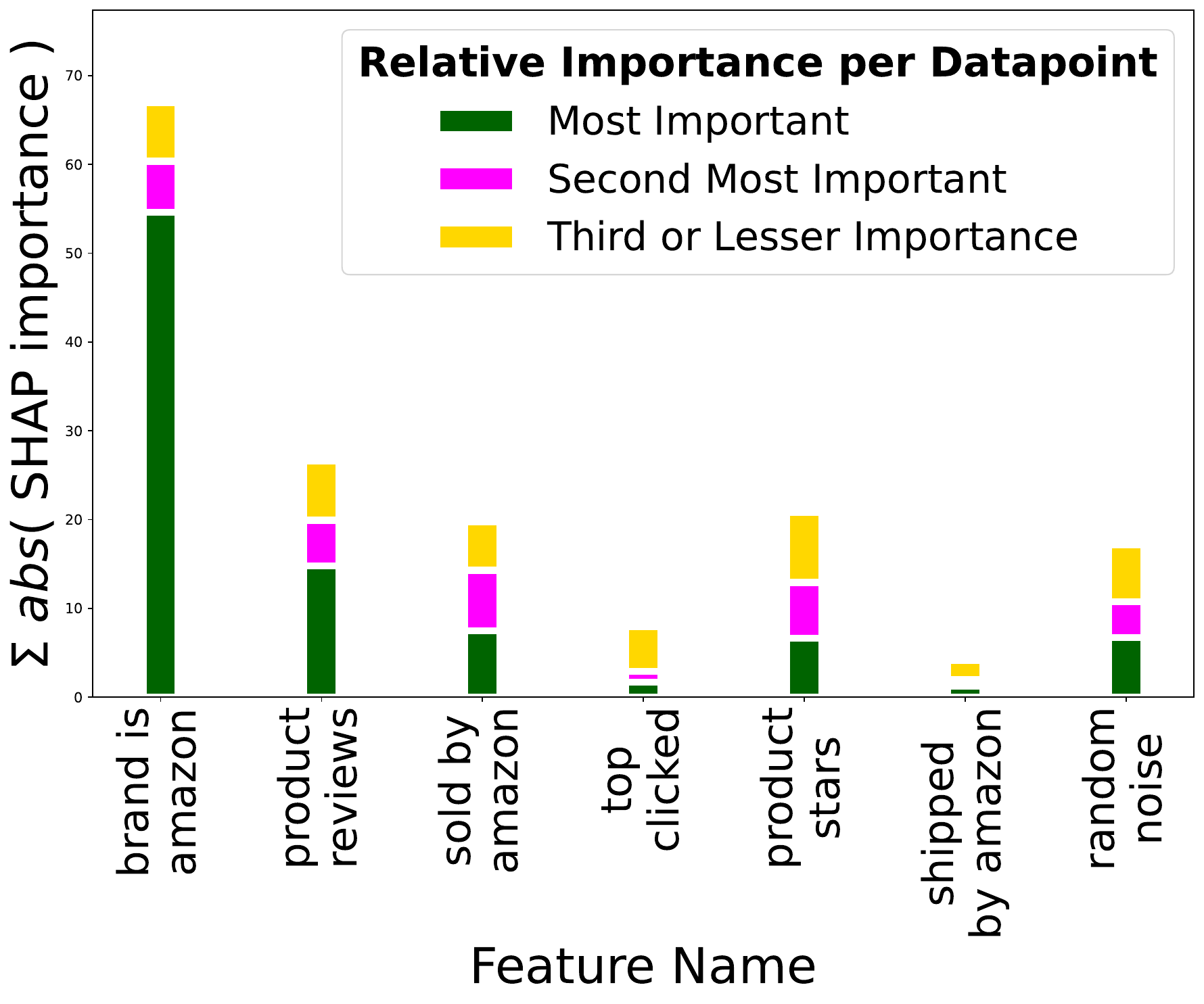}
		} \\
		\subfloat[The logistic regression model picks a different set of features as most important]{
			\includegraphics[width=0.42\textwidth]{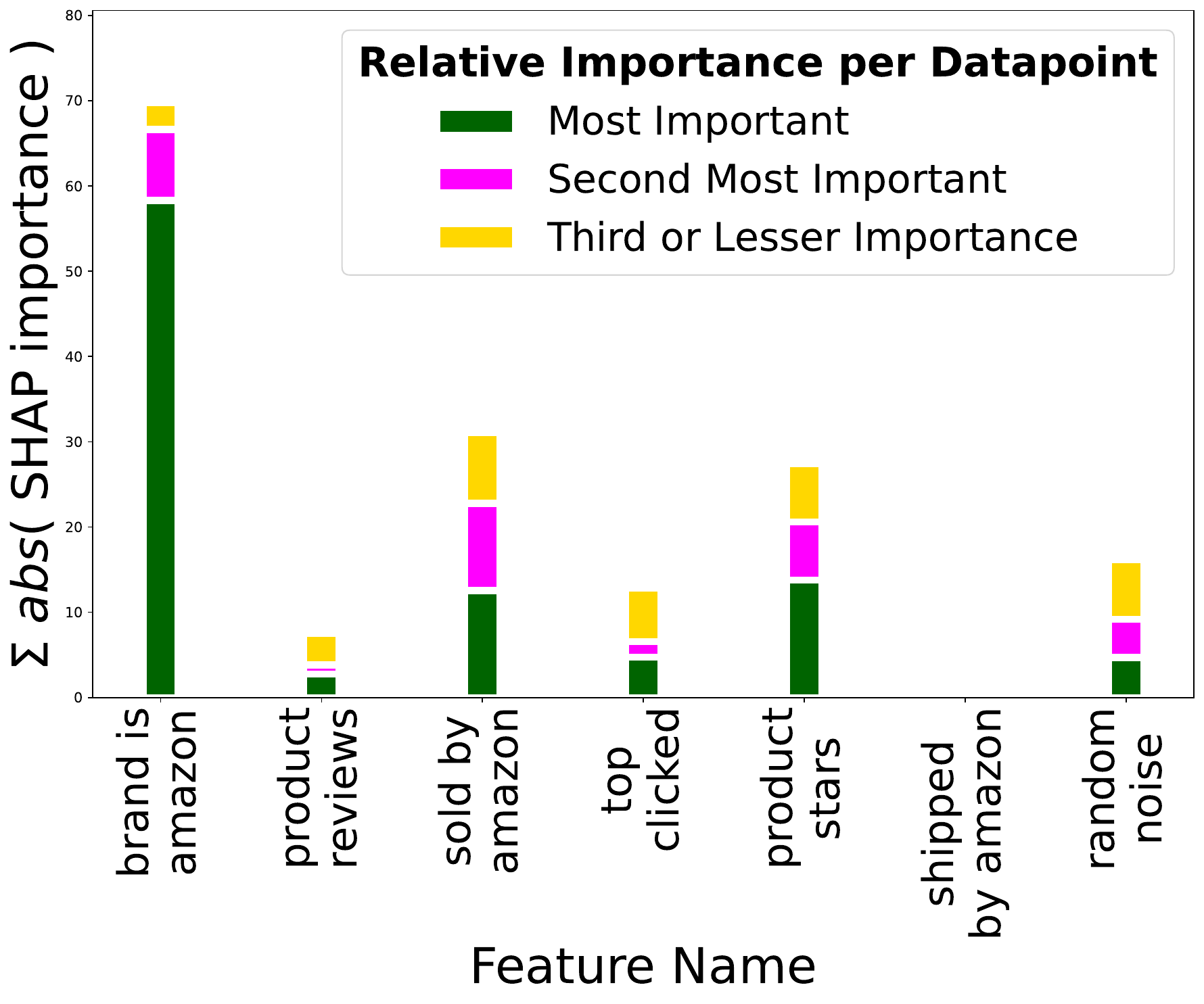}
		} & \subfloat[The L1 regularized logistic regression picks only ``brand is amazon'' as important]{
			\includegraphics[width=0.42\textwidth]{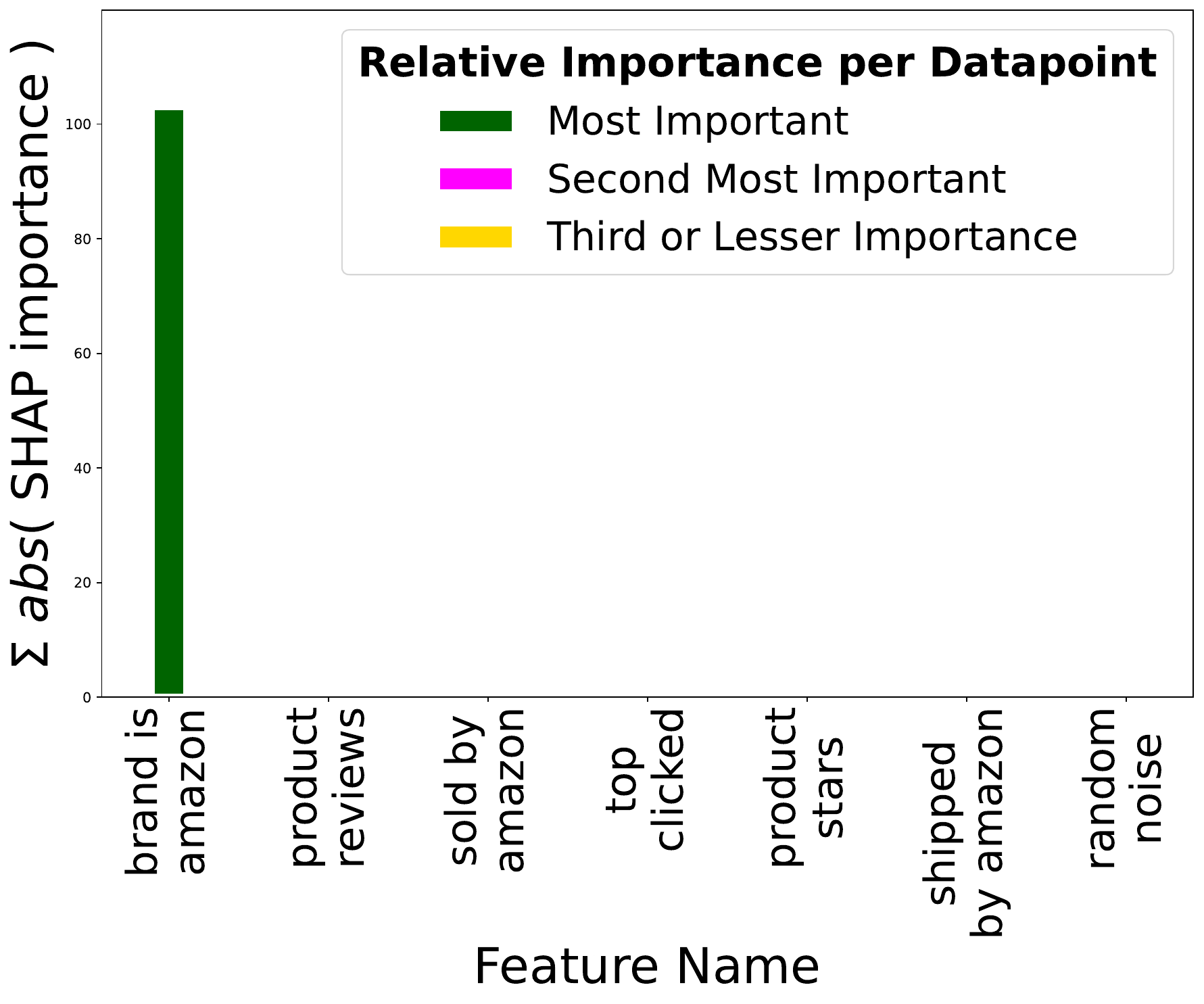}
		} \\
		\subfloat[The L2 regularized logistic regression model picks a different set of features as important]{
			\includegraphics[width=0.42\textwidth]{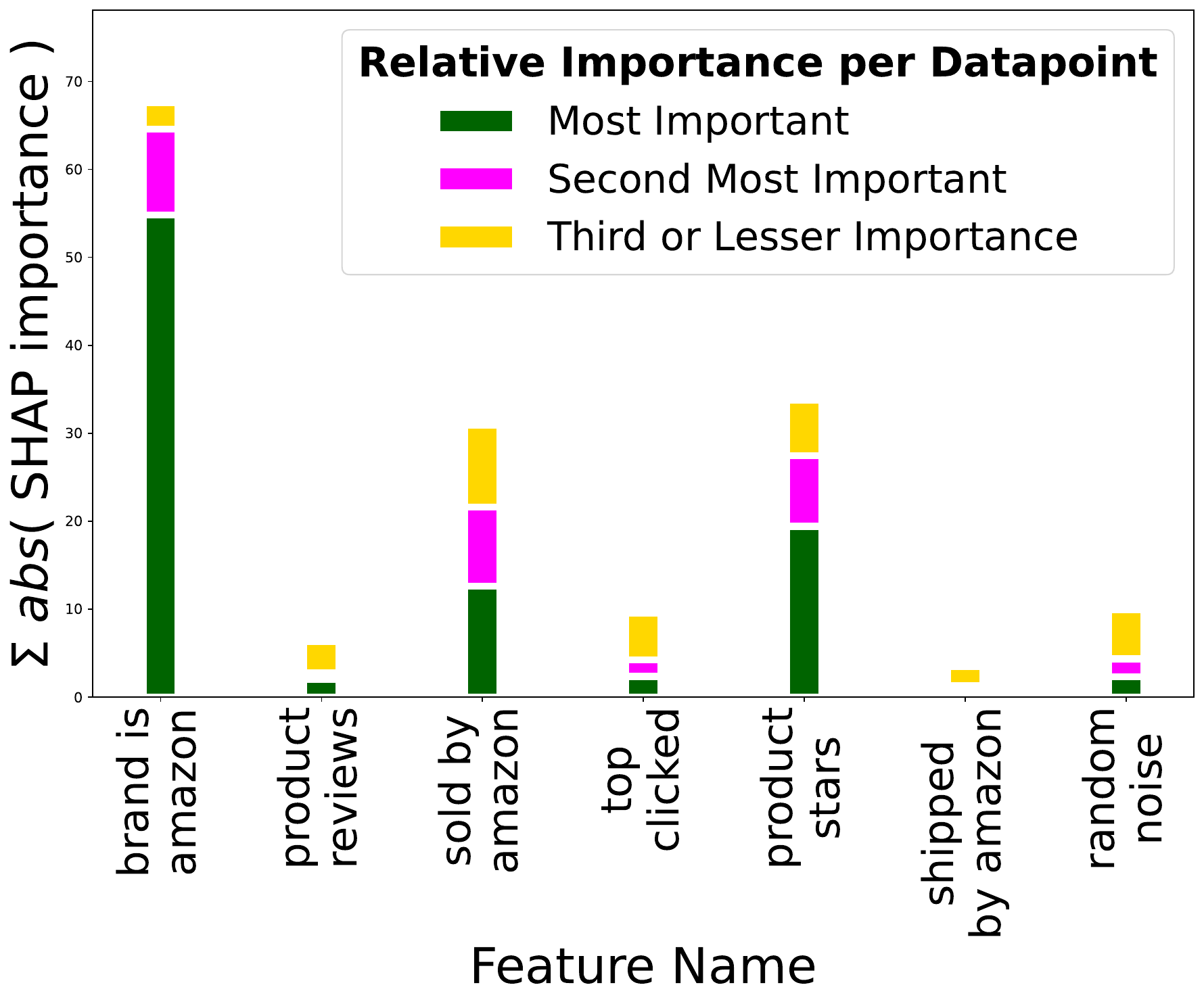}
		} & \subfloat[\RoT usually finds ``brand is amazon'' to be the most important, and often finds ``sold by amazon'' to be the second most important]{
			\includegraphics[width=0.42\textwidth]{Figures/markup/rot_bar.pdf}
		} \\
	\end{tabular}
    \caption{ \textbf{Auditing a proprietary recommendation system using XAI.} SHAP finds a different ordering of importances for the features, depending on the mimic model used, whereas \RoT provides a consistent result and uncovers an additional insight highlighting the importance of ``sold by amazon''}
    \label{fig:markup_bars_app}
\end{figure}

\begin{figure}
	\centering
	\begin{tabular}{cc}
		\subfloat[The Markup model finds ``brand is amazon'' and ``product reviews'' to have importances of high magnitude]{
			\includegraphics[width=0.47\textwidth]{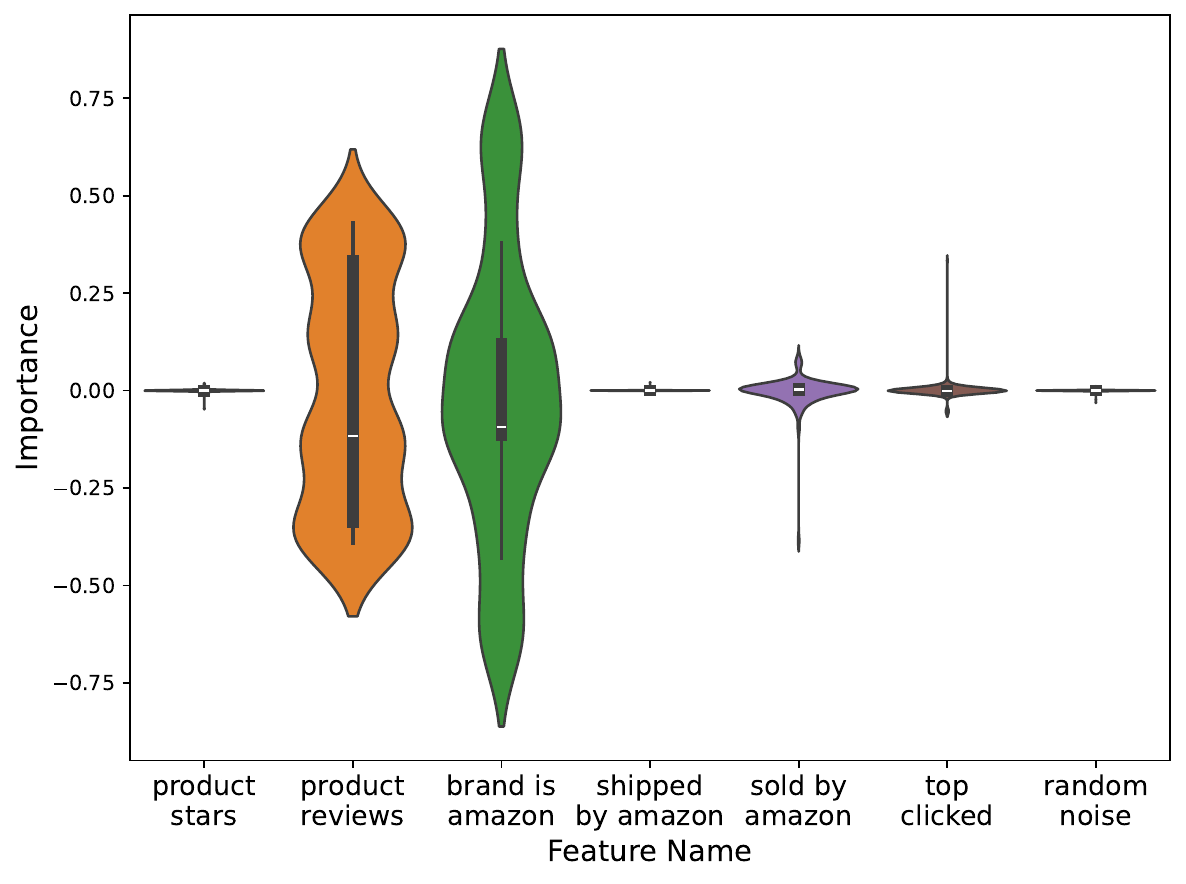}
		} & \subfloat[The default random forest model generates a different profile of importance distributions]{
			\includegraphics[width=0.47\textwidth]{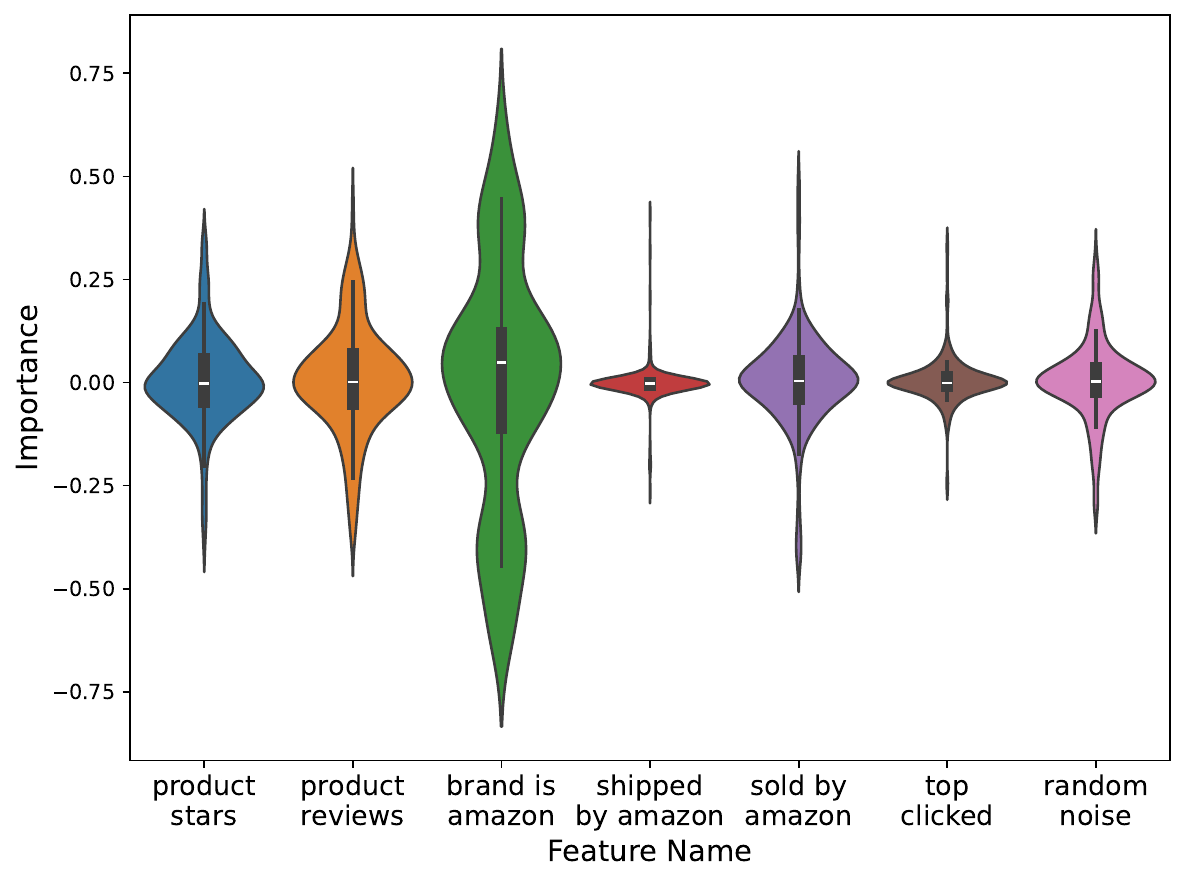}
		} \\
		\subfloat[The logistic regression model finds a different set of importances than both random forest models]{
			\includegraphics[width=0.47\textwidth]{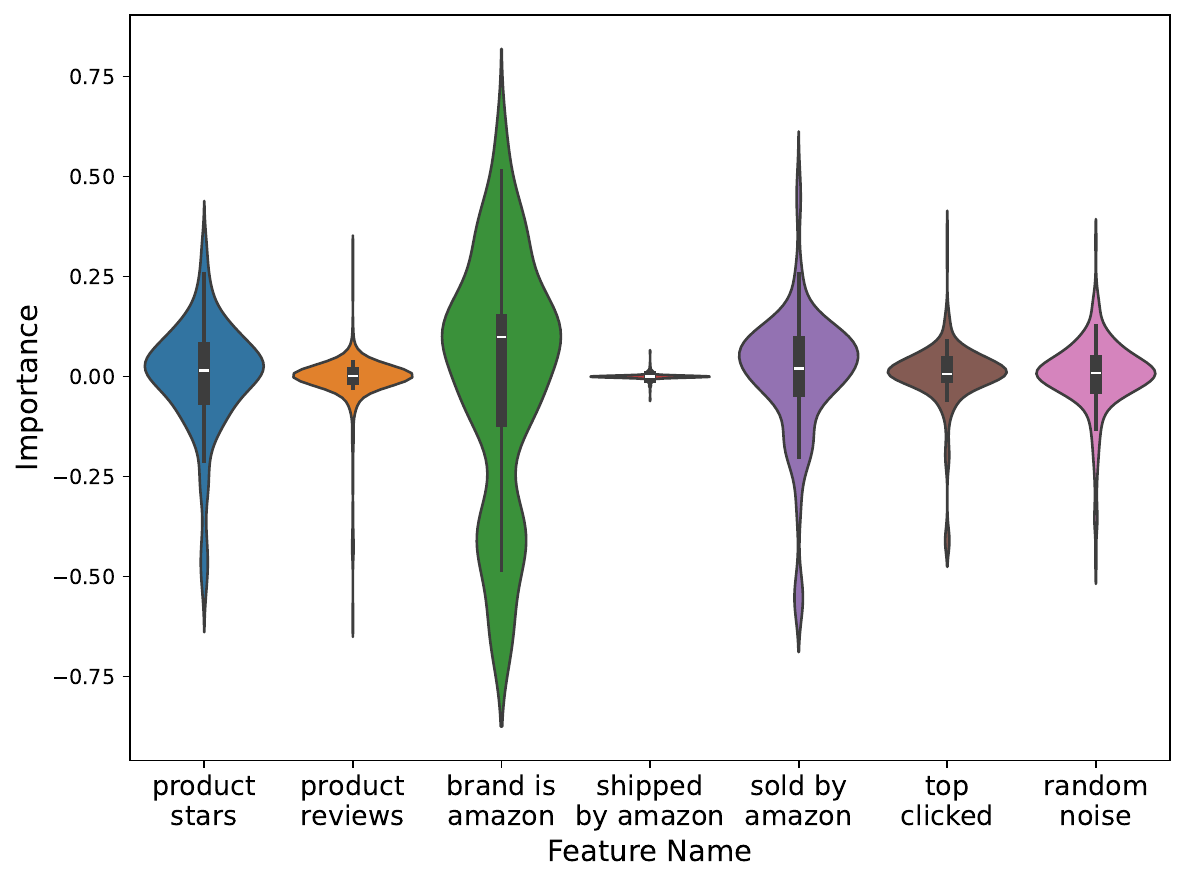}
		} & \subfloat[The L1 regularized regression model produces considers only ``brand is amazon'' to be important]{
			\includegraphics[width=0.47\textwidth]{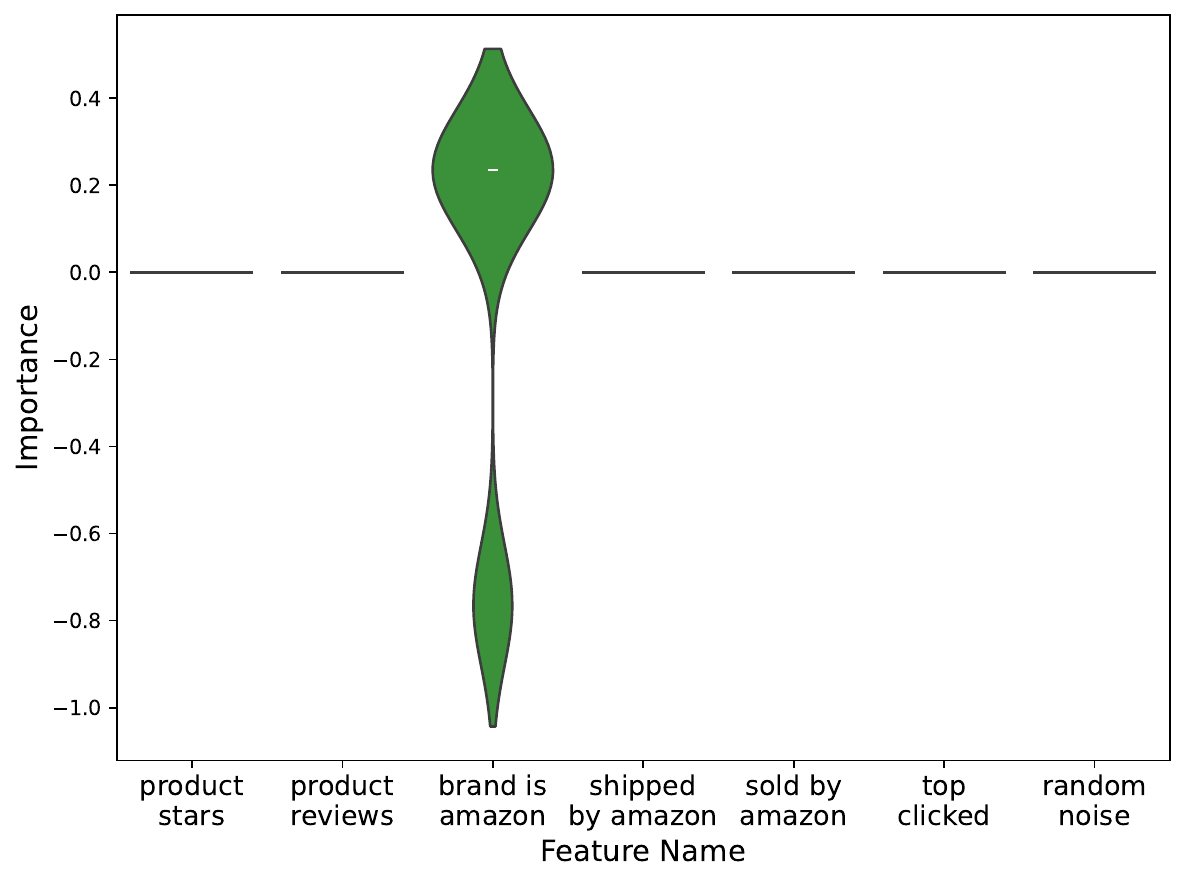}
		} \\
		\subfloat[The L2 regularized regression model produces a different importance distribution]{
			\includegraphics[width=0.47\textwidth]{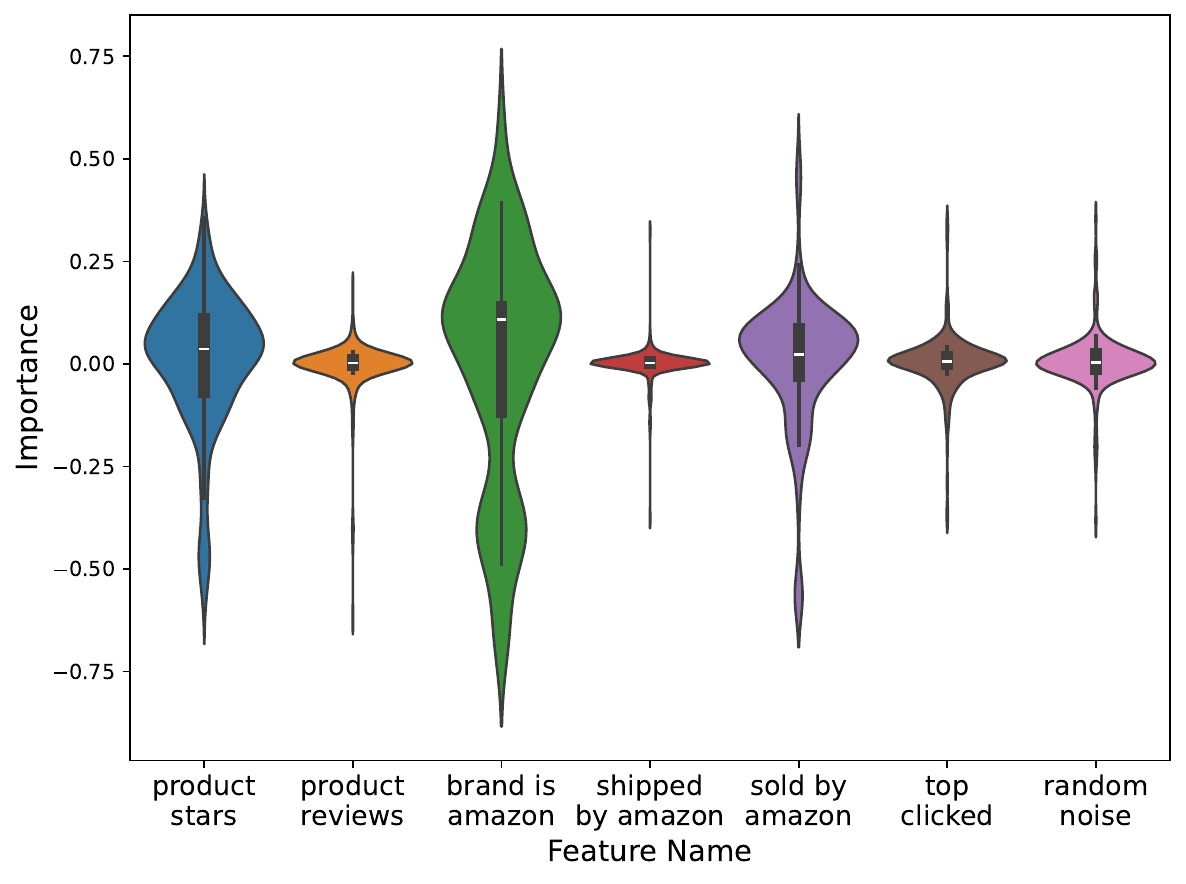}
		} & \subfloat[\RoT importances can be computed without any mimic models]{
			\includegraphics[width=0.47\textwidth]{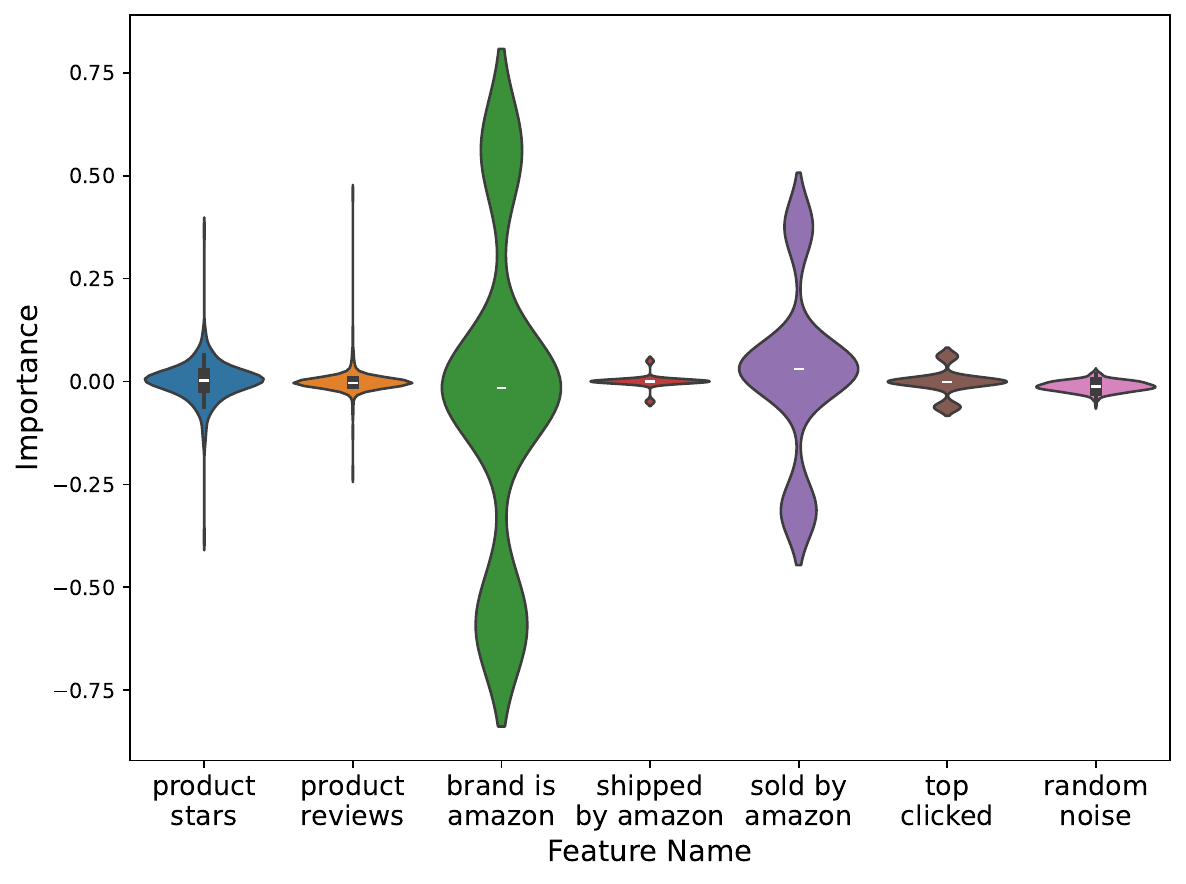}
		} \\
	\end{tabular}
    \caption{ \textbf{Auditing a proprietary recommendation system using XAI.} SHAP Importance Distributions for Input Features Can Vary by Choice of Mimic Model, whereas \RoT provides a Single and Robust Set of Importance Distributions for all Features, and Cannot be Manipulated by Choice of Mimic Model}
    \label{fig:markup_violins_app}
\end{figure}

Using SHAP on a mimic model is not guaranteed to recover explanations faithful to the original model being audited. We constructed different ``mimic'' models that emulated the Amazon product recommendation system, and showed that SHAP explanations contradicted each other for the Markup created Random Forest, and the default hyper-parameter Random Forest. We now extend this analysis to the other logistic regression mimic models we created, producing extended versions of Figure \ref{fig:markup_bars} showing SHAP explanation distributions and SHAP explanation feature rankings respectively in Figures \ref{fig:markup_violins_app} and \ref{fig:markup_bars_app} respectively. We also show global importances for all models, verifying that mimic models do indeed have different internal mechanisms despite producing predictions with similar accuracy rates.

\begin{table}[!h]
	\centering
    \caption{\textbf{Global feature importances (normalised) for the mimic models from Table \ref{table:markup_mimic_accuracies}.} The two most important features per model are marked in \textbf{bold}. These global explanations (Mean Decrease in Impurity for RF, model coefficients for LogReg) vary significantly, showing that models in the Rashomon set make the same prediction through different internal mechanisms.}
    \label{table:markup_mimic_importances}
    \begin{tabular}{l!{\vrule width 0.8pt}ccccc}
	\toprule
	& \textbf{RF (Markup)} & \textbf{RF (Default)} & \textbf{LogReg} & \textbf{LogReg (L1)} & \textbf{LogReg (L2)} \\
	\midrule
	\textbf{product stars} & 0.021 & 0.122 & 0.068 & 0.0 & 0.084 \\
	\rowcolor{gray!15} \textbf{product reviews} & \textbf{0.055} & \textbf{0.246} & 0.042 & 0.0 & 0.032 \\
	\textbf{brand is amazon} & \textbf{0.853} & \textbf{0.255} & \textbf{0.650} & \textbf{1.0} & \textbf{0.587} \\
	\rowcolor{gray!15} \textbf{shipped by amazon} &  0.006 & 0.027 & 0.003 & 0.0 & 0.012 \\
	\textbf{sold by amazon} & 0.052 & 0.082 & \textbf{0.177} & 0.0 & \textbf{0.245} \\
	\rowcolor{gray!15} \textbf{top clicked} &  0.002 & 0.034 & 0.032 & 0.0 & 0.022 \\
	\textbf{random noise} & 0.010 & 0.233 & 0.027 & 0.0 & 0.018 \\
	\bottomrule
	\end{tabular}
\end{table}
\clearpage

\clearpage

\clearpage

\begin{figure}
	\centering
	\begin{tabular}{cc}
		\subfloat[The Markup model finds the most important feature to be ``brand is amazon'' with ``sold by amazon'' second]{
			\includegraphics[width=0.42\textwidth]{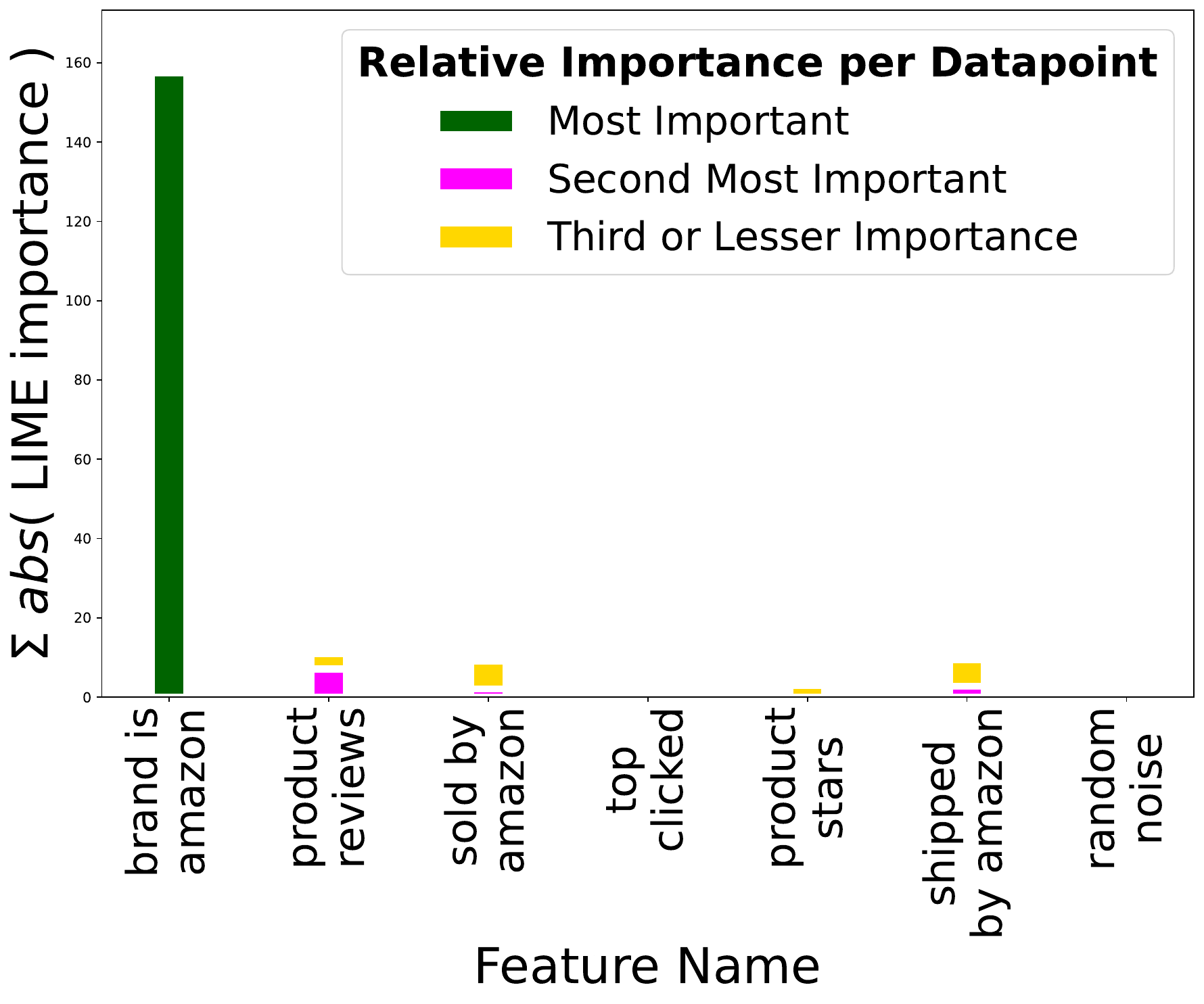}
		} & \subfloat[The default random forest model picks a more diverse set of features as most important]{
			\includegraphics[width=0.42\textwidth]{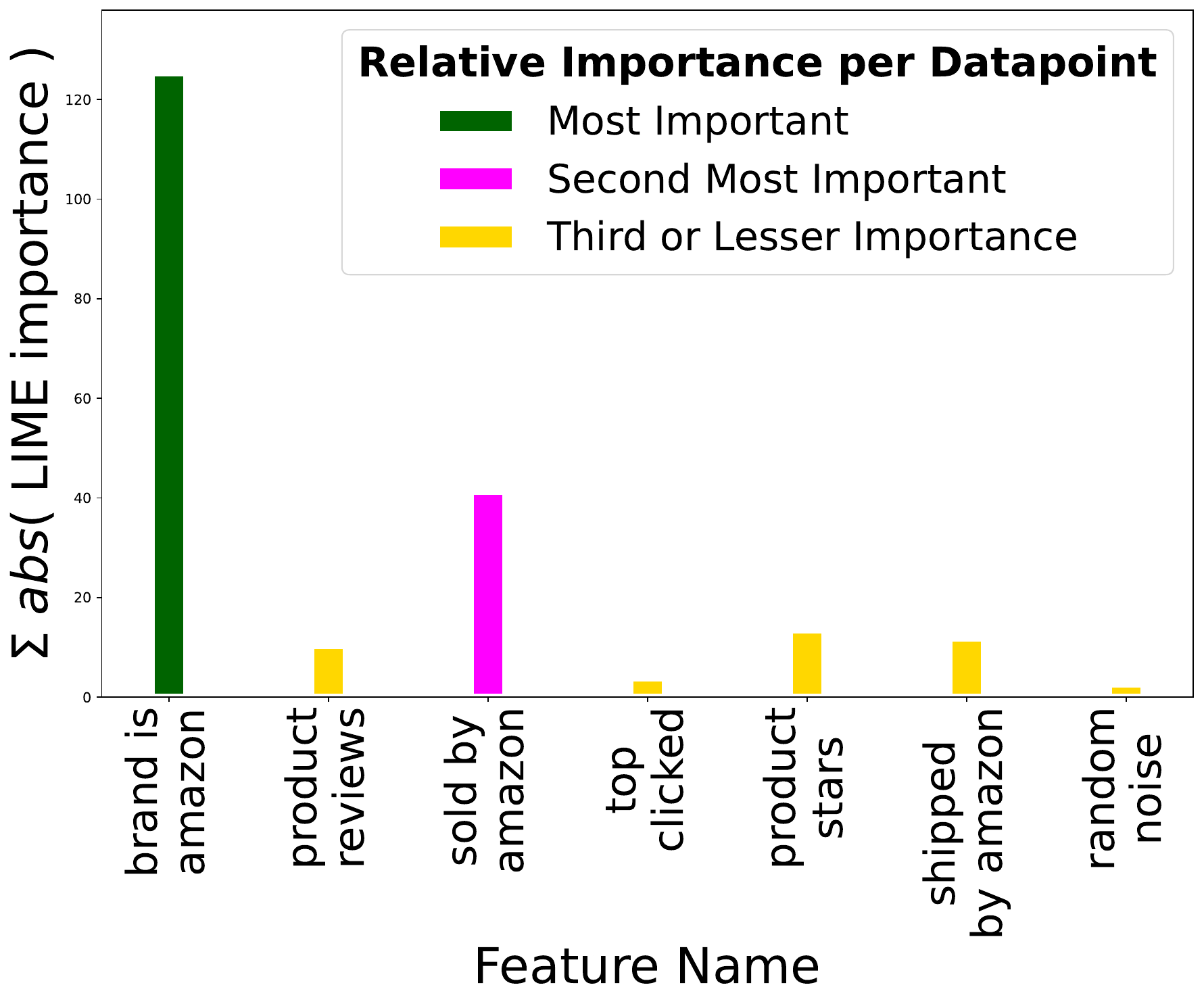}
		} \\
		\subfloat[The logistic regression model picks a different set of features as most important]{
			\includegraphics[width=0.42\textwidth]{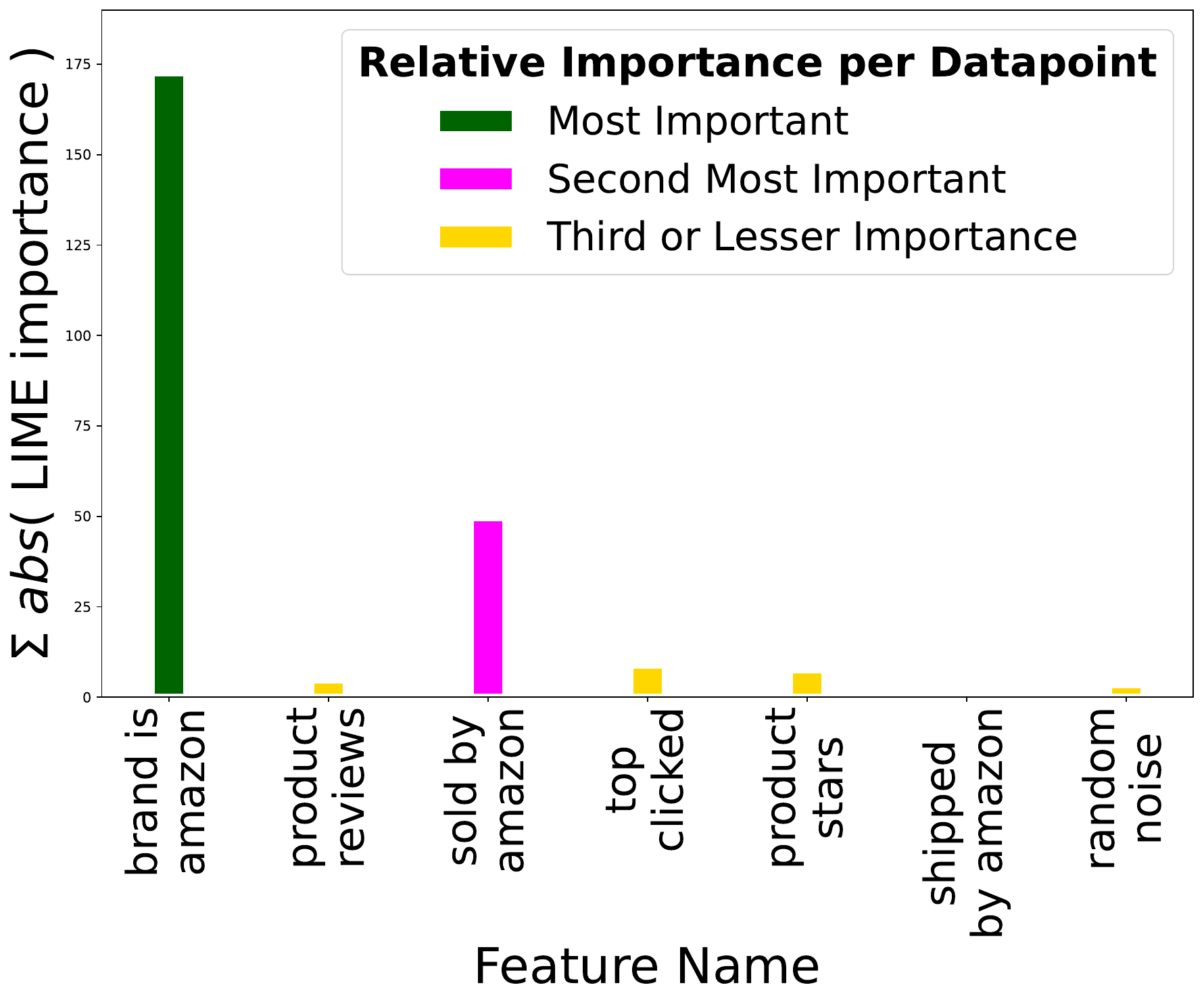}
		} & \subfloat[The L1 regularized logistic regression picks only ``brand is amazon'' as important]{
			\includegraphics[width=0.42\textwidth]{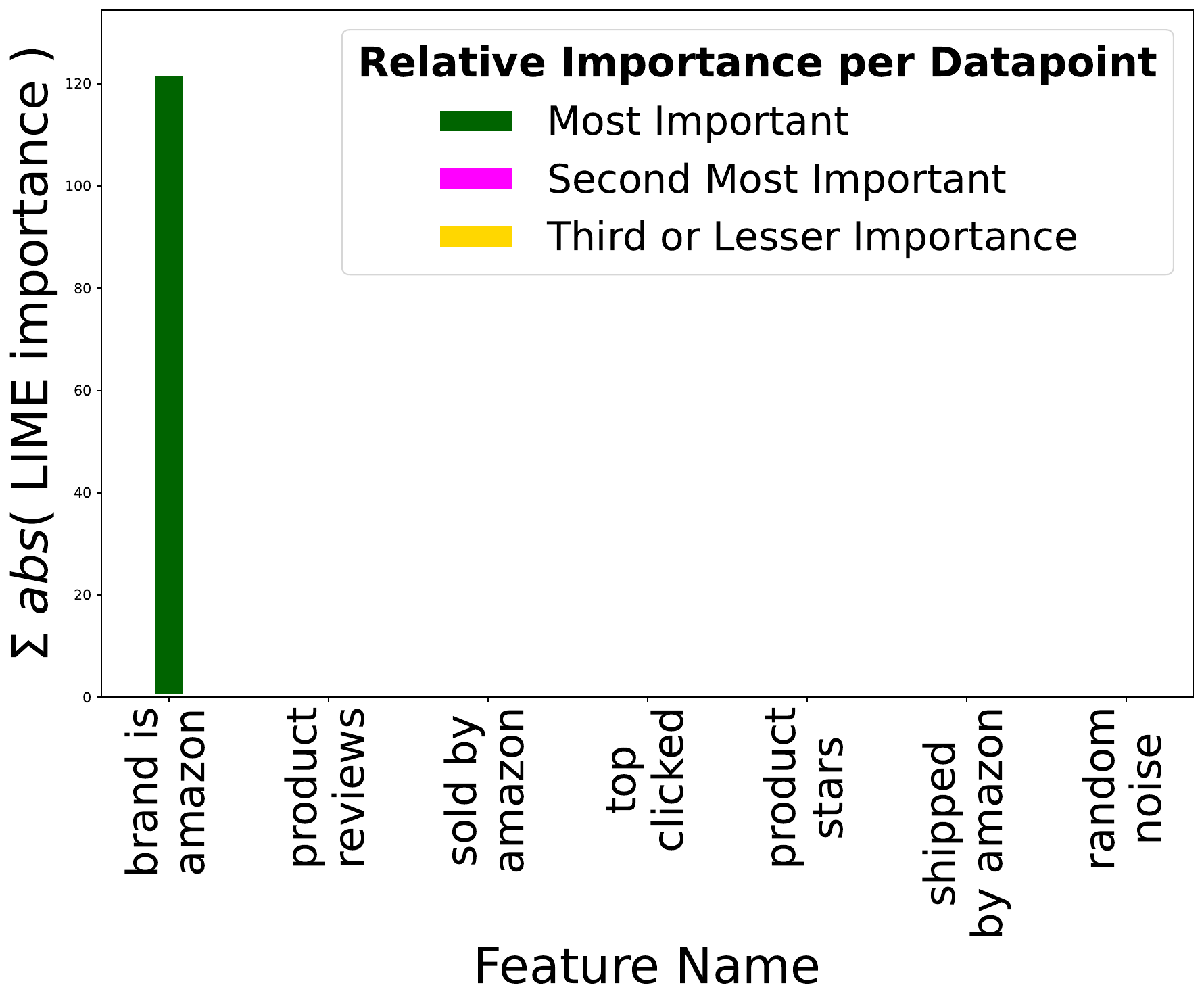}
		} \\
		\subfloat[The L2 regularized logistic regression model picks a different set of features as important]{
			\includegraphics[width=0.42\textwidth]{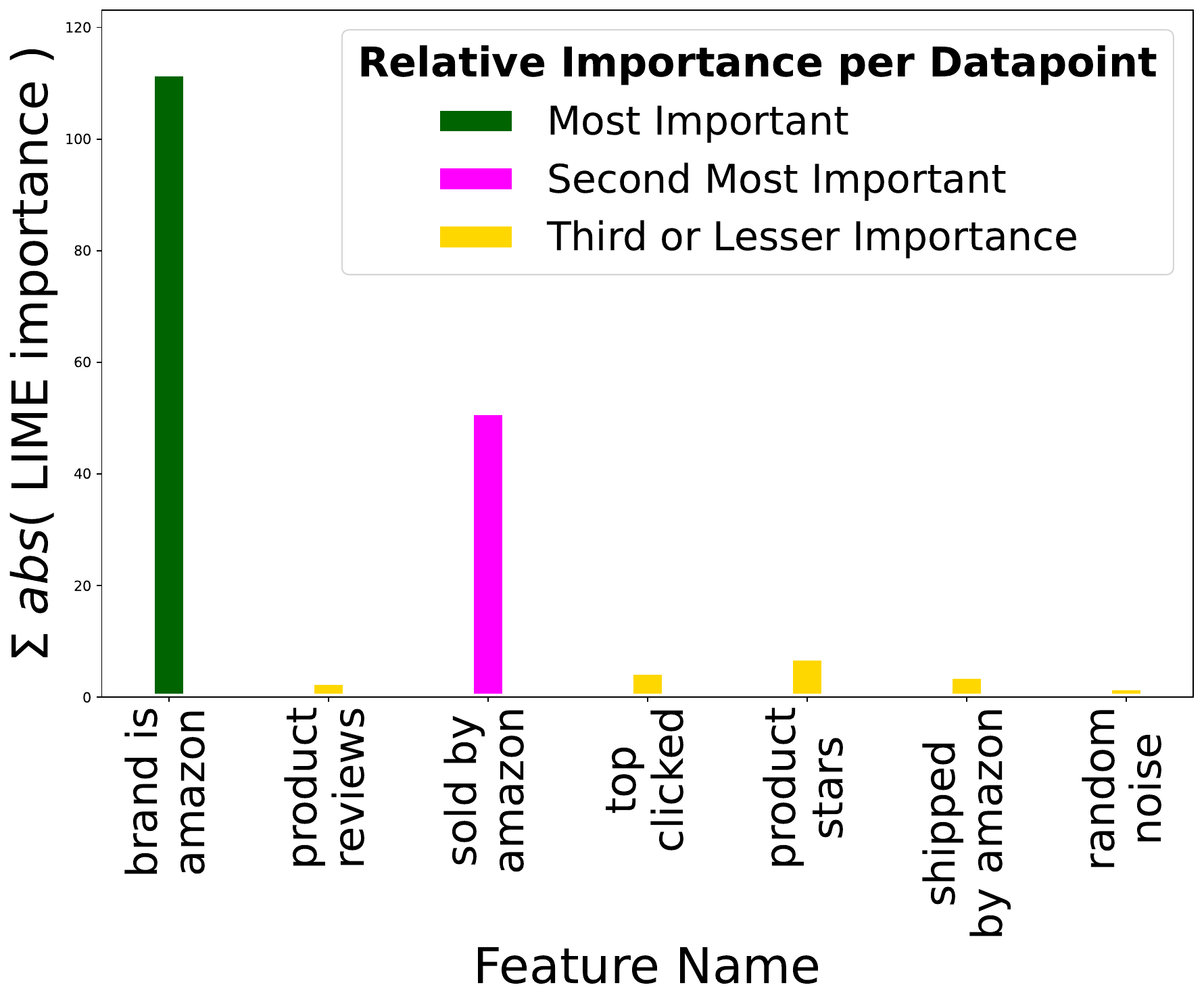}
		} & \subfloat[\RoT usually finds ``brand is amazon'' to be the most important, and often finds ``sold by amazon'' to be the second most important]{
			\includegraphics[width=0.42\textwidth]{Figures/markup/rot_bar.pdf}
		} \\
	\end{tabular}
    \caption{ \textbf{Auditing a proprietary recommendation system using XAI.} LIME finds a different ordering of importances for the features, depending on the mimic model used, whereas \RoT provides a consistent result and uncovers an additional insight highlighting the importance of ``sold by amazon''}
    \label{fig:markup_bars_app_lime}
\end{figure}
\clearpage

\clearpage

\begin{figure}
	\centering
	\begin{tabular}{cc}
		\subfloat[The Markup model finds ``brand is amazon'' and ``product reviews'' to have importances of high magnitude]{
			\includegraphics[width=0.47\textwidth]{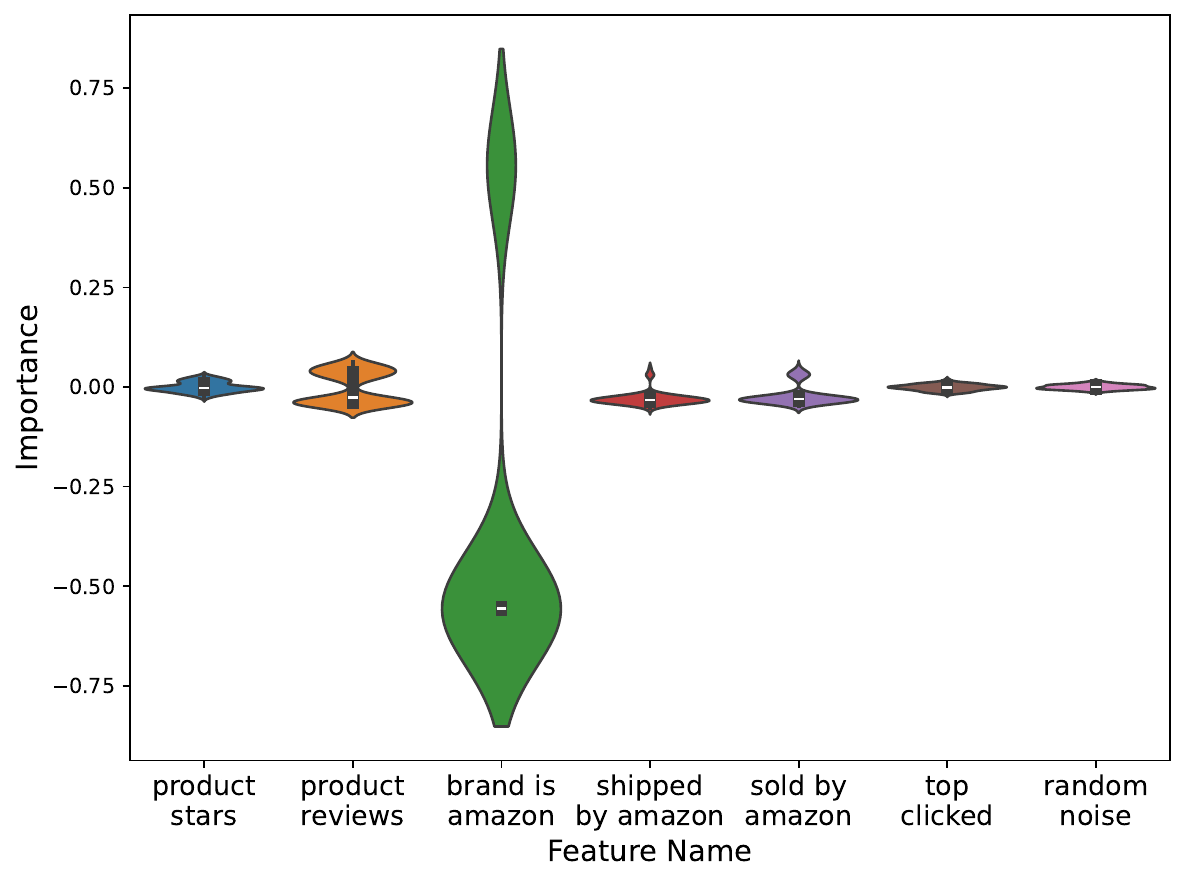}
		} & \subfloat[The default random forest model generates a different profile of importance distributions]{
			\includegraphics[width=0.47\textwidth]{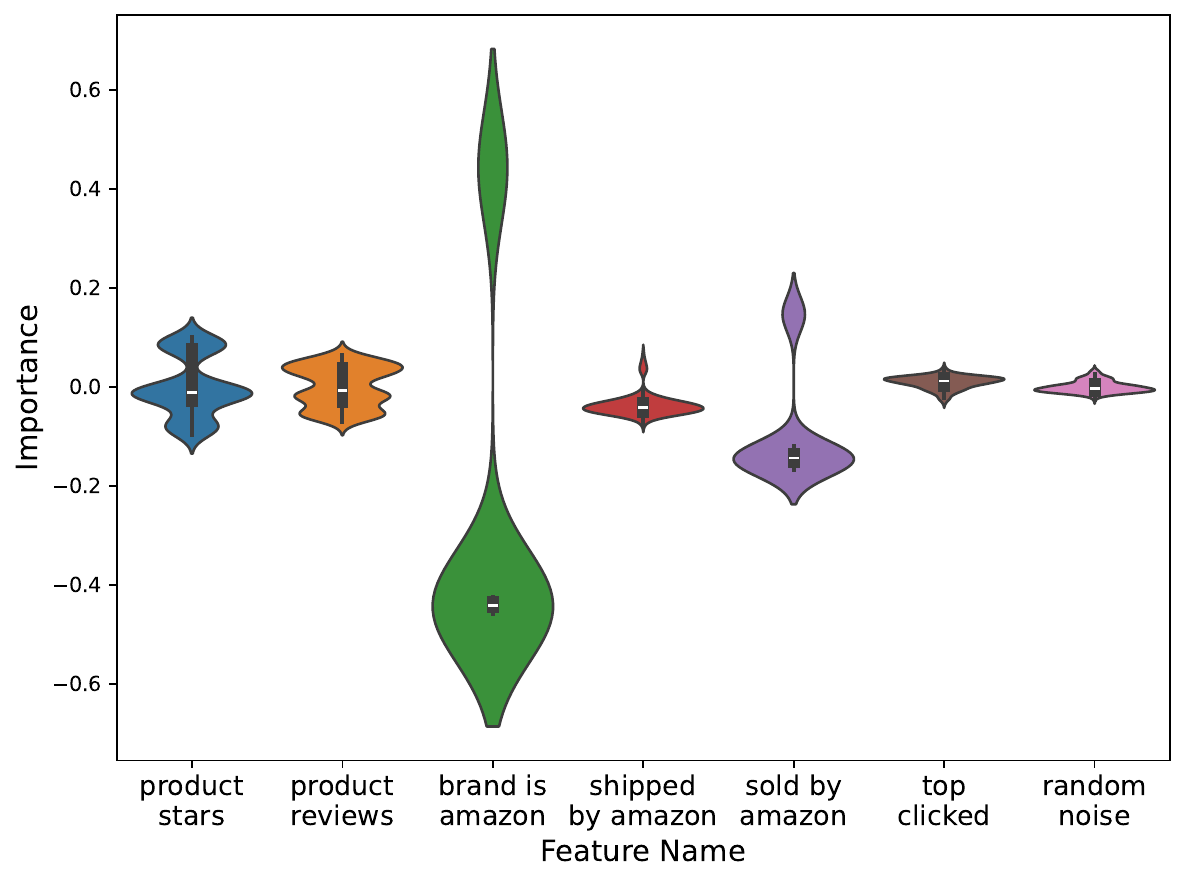}
		} \\
		\subfloat[The logistic regression model finds a different set of importances than both random forest models]{
			\includegraphics[width=0.47\textwidth]{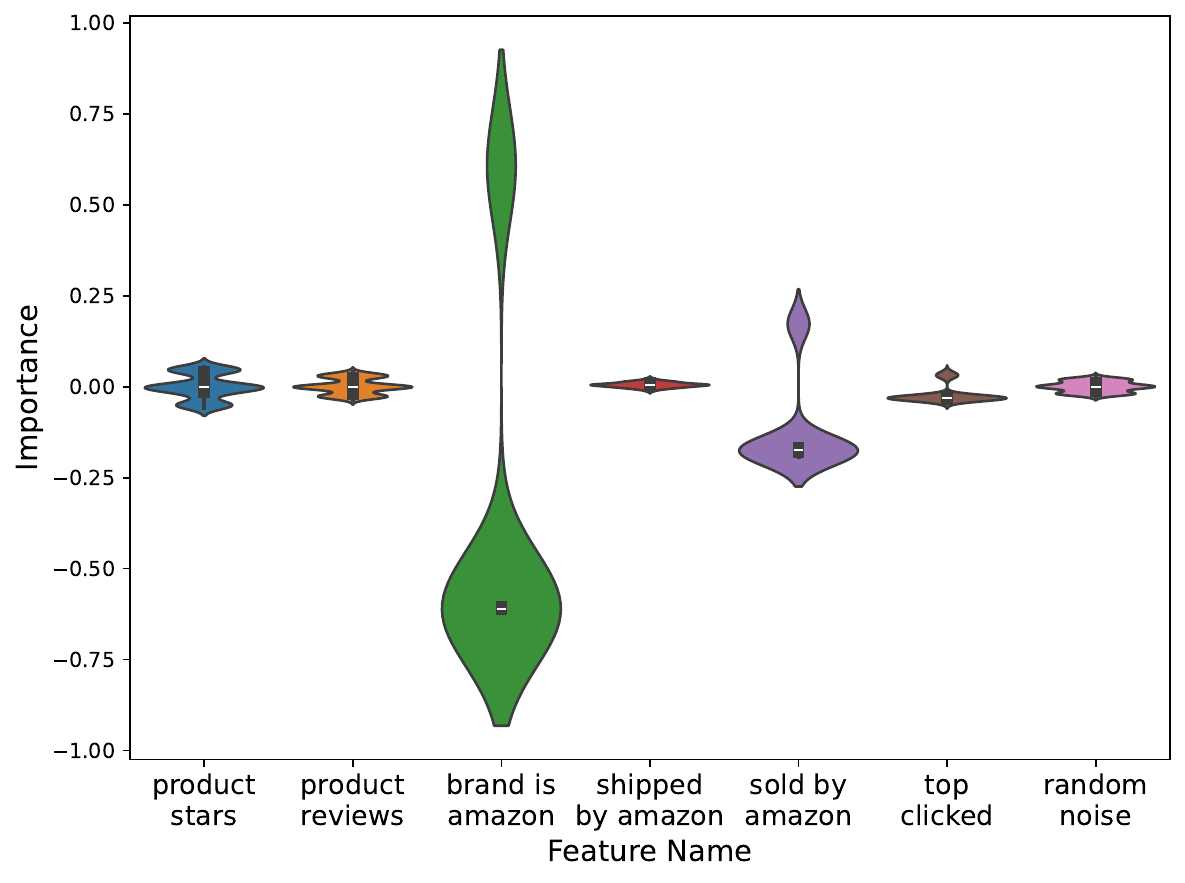}
		} & \subfloat[The L1 regularized regression model produces considers only ``brand is amazon'' to be important]{
			\includegraphics[width=0.47\textwidth]{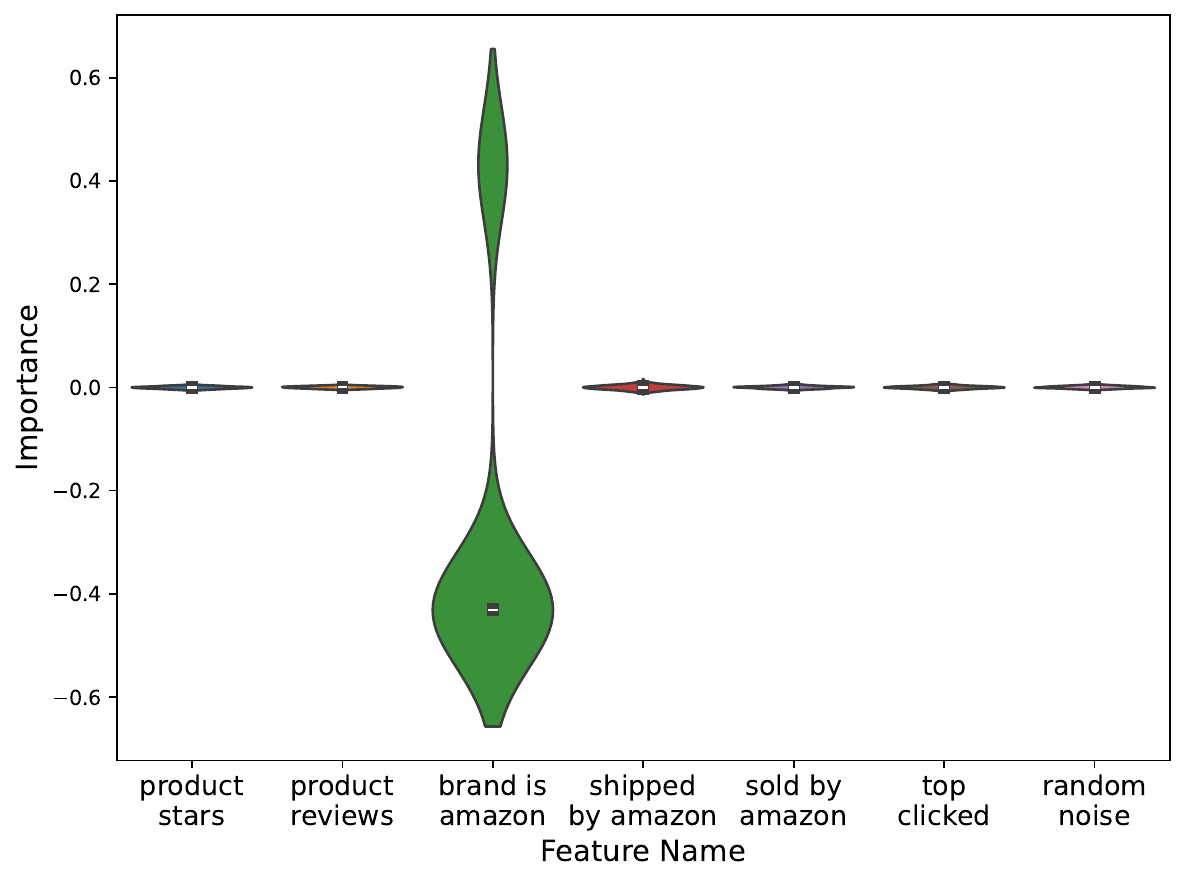}
		} \\
		\subfloat[The L2 regularized regression model produces a different importance distribution]{
			\includegraphics[width=0.47\textwidth]{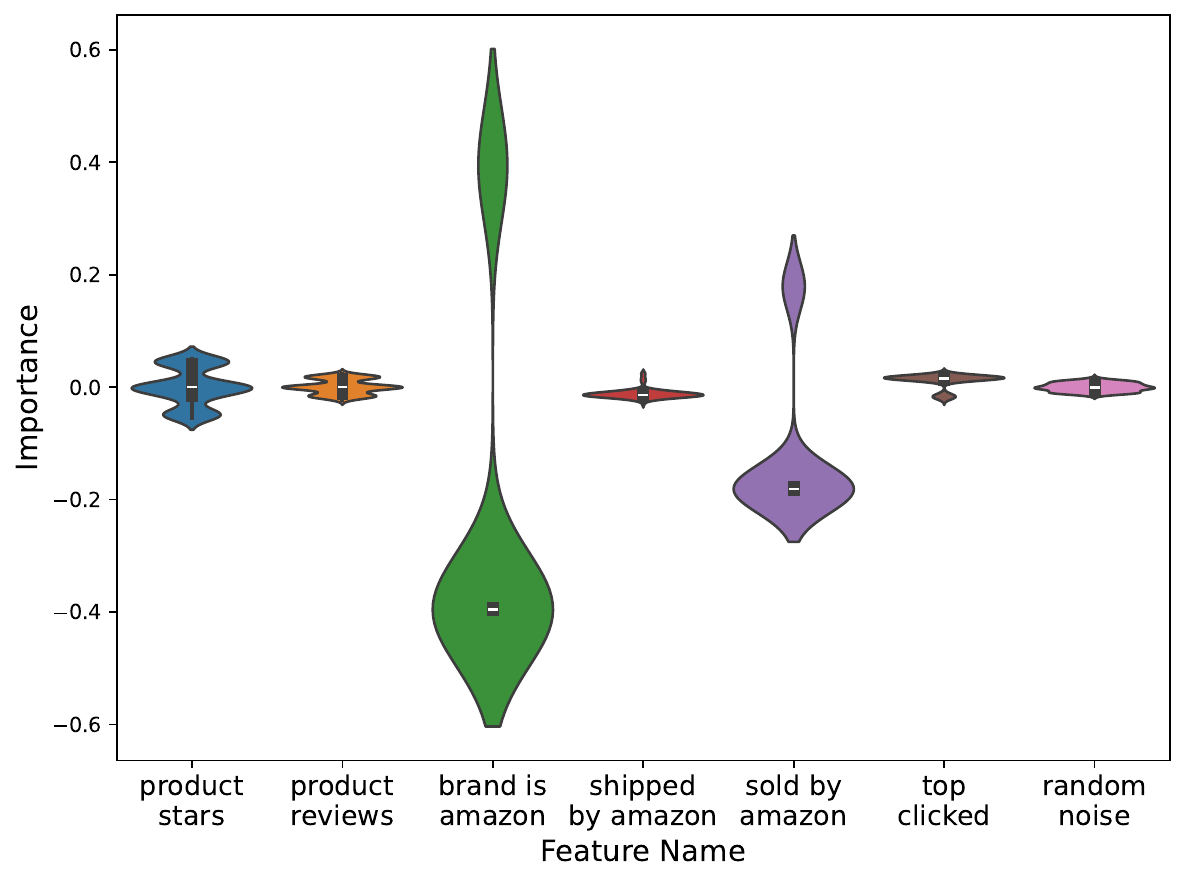}
		} & \subfloat[\RoT importances can be computed without a mimic model]{
			\includegraphics[width=0.47\textwidth]{Figures/markup/rot_violin.pdf}
		} \\
	\end{tabular}
    \caption{ \textbf{Auditing a proprietary recommendation system using XAI.} LIME Importance Distributions for Input Features Can Vary by Choice of Mimic Model, whereas \RoT provides a Single and Robust Set of Importance Distributions for all Features, and Cannot be Manipulated by Choice of Mimic Model}
    \label{fig:markup_violins_app_lime}
\end{figure}
\clearpage

\subsection{Scientific Discovery using XAI}
\label{subsubsec:science}
\paragraph{Identifying Important Features in the presence of Misleading Features}
Table \ref{tab:advattack} effectively summarises the most important feature across datapoints, as picked by various explainers. This is the underlying explanation data that was plotted in the ternary distribution plots in Figure \ref{fig:advattack}. To provide a more granular analysis, we also provide additional confusion matrices comparing the first, second, and third most important features as selected by various explainers across various datasets. in Tables \ref{tab:conf_matrices1}, \ref{tab:conf_matrices2}, \ref{tab:conf_matrices3}, \ref{tab:conf_matrices4}, and \ref{tab:conf_matrices5}.

\begin{table}[!b]
	\centering
    \small
    \caption{ \textbf{\RoT Successfully Determines the Sensitive Feature as Most Important, Despite the Presence of Misleading Foil Features}: adversarially crafted models ${m_L}$ and ${m_S}$ are designed to mislead LIME and SHAP respectively into selecting the Foil features as most important. For both adversarial models, we verify that \RoT successfully identifies the Sensitive feature as most important. In each case, we present the percentage of predictions that yield a particular input feature as most important. {\color{teal} \textbf{Successes are in bold green}}, {\color{purple} \textbf{adversarial failures in bold red}}, and {\color{blue} \textbf{non-adversarial failures in bold blue}.} }
    \label{tab:advattack}
    \begin{tabular}{lcc!{\vrule width 0.8pt} r @{\hspace{1.2cm}} r @{\hspace{1.2cm}} r @{\hspace{1.2cm}} r}
	\toprule
	\multirow{2}{*}{\thead{ \textbf{Dataset} }} &
	\multirow{2}{*}{\thead{ \textbf{Adversarial} \\ \textbf{Model} }} & 
	\multirow{2}{*}{\thead{ \textbf{Explainer} }} &
	\multicolumn{4}{c}{\thead{\textit{Percentage of Dataset where the Most Important Input Feature is:}}} \\ 
	 & & &
	\thead{a) \textbf{Sensitive}} &
	\thead{b) \textbf{Foil \#1}} &
	\thead{c) \textbf{Foil \#2}} &
	\thead{d) \textbf{Other}} \\
	\midrule
	\multirow{4}{*}{\makecell[l]{ \textbf{German} \\ \textbf{Credit} } }
	 & \multirow{2}{*}{ \makecell[c]{ \(\bm{m_L}\) \\ \textbf{\textit{(1 foil)}} } } 
	 & LIME & 0.0 & {\color{purple} \textbf{84.0}} & na & {\color{blue} \textbf{16.0}} \\
	 & & \rot & {\color{teal} \textbf{100.0}} & 0.0 & na & 0.0 \\ 
	\cdashline{2-7}[1pt/2pt]
	 & \multirow{2}{*}{ \makecell[c]{ \(\bm{m_S}\) \\ \textbf{\textit{(1 foil)}} } } 
	 & SHAP & 0.0 & {\color{purple} \textbf{100.0}} & na & 0.0 \\
	 & & \rot & {\color{teal} \textbf{100.0}} & 0.0 & na & 0.0 \\
	\hdashline
	\multirow{8}{*}{ \makecell[l]{ \textbf{COMPAS} } } 
	 & \multirow{2}{*}{ \makecell[c]{ \(\bm{m_L}\) \\ \textbf{\textit{(1 foil)}} } } 
	 & LIME & 0.0 & {\color{purple} \textbf{100.0}} & na & 0.0 \\
	 & & \rot & {\color{teal} \textbf{100.0}} & 0.0 & na & 0.0 \\ 
	\cdashline{2-7}[1pt/2pt]
	 & \multirow{2}{*}{ \makecell[c]{ \(\bm{m_S}\) \\ \textbf{\textit{(1 foil)}} } } 
	 & SHAP & 1.9 & {\color{purple} \textbf{90.9}} & na & {\color{blue} \textbf{7.1}} \\
	 & & \rot & {\color{teal} \textbf{100.0}} & 0.0 & na & 0.0 \\ 
	\cdashline{2-7}[1pt/2pt]
	 & \multirow{2}{*}{ \makecell[c]{ \(\bm{m_L}\) \\ \textbf{\textit{(2 foils)}} } } 
	 & LIME & 0.3 & {\color{purple} \textbf{43.9}} & {\color{purple} \textbf{47.9}} & {\color{blue} \textbf{7.9}} \\
	 & & \rot & {\color{teal} \textbf{100.0}} & 0.0 & 0.0 & 0.0 \\ 
	\cdashline{2-7}[1pt/2pt]
	 & \multirow{2}{*}{ \makecell[c]{ \(\bm{m_S}\) \\ \textbf{\textit{(2 foils)}} } } 
	 & SHAP & {\color{teal} \textbf{35.0}} & {\color{purple} \textbf{18.9}} & {\color{purple} \textbf{22.3}} & {\color{blue} \textbf{23.8}} \\
	 & & \rot & {\color{teal} \textbf{100.0}} & 0.0 & 0.0 & 0.0 \\ 
	\hdashline
	\multirow{8}{*}{ \makecell[l]{ \textbf{Communities} \\ \textbf{and Crime} } } 
	 & \multirow{2}{*}{ \makecell[c]{ \(\bm{m_L}\) \\ \textbf{\textit{(1 foil)}} } } 
	 & LIME & 0.0 & {\color{purple} \textbf{99.5}} & na & 0.5 \\
	 & & \rot & {\color{teal} \textbf{89.5}} & 0.0 & na & {\color{blue} \textbf{10.5}} \\ 
	\cdashline{2-7}[1pt/2pt]
	 & \multirow{2}{*}{ \makecell[c]{ \(\bm{m_S}\) \\ \textbf{\textit{(1 foil)}} } } 
	 & SHAP & 4.0 & {\color{purple} \textbf{87.0}} & na & {\color{blue} \textbf{9.0}} \\
	 & & \rot & {\color{teal} \textbf{90.5}} & 0.0 & na & {\color{blue} \textbf{9.5}} \\ 
	\cdashline{2-7}[1pt/2pt]
	 & \multirow{2}{*}{ \makecell[c]{ \(\bm{m_L}\) \\ \textbf{\textit{(2 foils)}} } } 
	 & LIME & 0.0 & {\color{purple} \textbf{14.5}} & {\color{purple} \textbf{74.5}} & {\color{blue} \textbf{11.0}} \\
	 & & \rot & {\color{teal} \textbf{91.0}} & 0.0 & 0.0 & {\color{blue} \textbf{9.0}} \\
	\cdashline{2-7}[1pt/2pt]
	 & \multirow{2}{*}{ \makecell[c]{ \(\bm{m_S}\) \\ \textbf{\textit{(2 foils)}} } } 
	 & SHAP & 1.0 & {\color{purple} \textbf{49.0}} & {\color{purple} \textbf{50.0}} & 0.0 \\
	 & & \rot & {\color{teal} \textbf{90.0}} & 0.0 & 0.0 & {\color{blue} \textbf{10.0}} \\ 
	\bottomrule
	\end{tabular}
\end{table}

% \begin{table}
% 	\centering
%     \caption{ \textbf{\RoT Successfully Determines the Sensitive Feature as Most Important, Despite the Presence of Misleading Foil Features}: adversarially crafted models ${m_L}$ and ${m_S}$ are designed to mislead LIME and SHAP respectively into selecting the Foil features as most important. For both adversarial models, we verify that \RoT successfully identifies the Sensitive feature as most important. In each case, we present the percentage of predictions that yield a particular input feature as most important. {\color{teal} \textbf{Successes are in bold green}}, {\color{purple} \textbf{adversarial failures in bold red}}, and {\color{blue} \textbf{non-adversarial failures in bold blue}.} }
%     \label{tab:advattack}
%     \includegraphics[page=25,trim=1 1 1 1,clip,width=1.0\linewidth]{figures/Combined.pdf}
% \end{table}

\begin{table}
	\centering
    \caption{\textbf{COMPAS dataset, with a synthetic foil feature}: Confusion matrices showing what percentage of explanations select particular features as the most, second most, and third most important features in explanations. Ideally, the most important feature recovered by an explainer should be the only one used (highlighted {\color{green} green}), whereas the adversary tries to have the explaners always pick a different feature (highlighted {\color{red} red}). The most important feature (first column) from each matrix is replicated in Table \ref{tab:advattack}  }
    \label{tab:conf_matrices1}
    \begin{tabular}{cc}
		\subfloat[SHAP explanations for adversarially crafted model]{
			\begin{tabular}{c|ccc}
			\makecell{\textbf{Importance} \\ \textbf{Rank}} & \textbf{1} & \textbf{2} & \textbf{3} \\
			\hline
			\rowcolor{red!25}
			\textbf{Synthetic Foil \#1} & \textbf{90.9} & \textbf{7.3} & 1.8 \\
			\rowcolor{green!25}
			\textbf{race} & 1.9 & \textbf{55} & \textbf{19.9} \\
			\rowcolor{gray!25}
			\textbf{other features} & \textbf{7.1} & \textbf{37.7} & \textbf{78.3} \\
			\end{tabular}
		} &
		\subfloat[\RoT explanations for adversarial model]{
			\begin{tabular}{c|ccc}
			\makecell{\textbf{Importance} \\ \textbf{Rank}} & \textbf{1} & \textbf{2} & \textbf{3} \\
			\hline
			\rowcolor{red!25}
			\textbf{Synthetic Foil \#1} & 0.0 & 0.0 & 0.0 \\
			\rowcolor{green!25}
			\textbf{race} & \textbf{100.0} & 0.0 & 0.0 \\
			\rowcolor{gray!25}
			\textbf{other features} & 0.0 & \textbf{100.0} & \textbf{100} \\
			\end{tabular}
		} \\
		\\
		\subfloat[LIME explanations for adversarial model]{
			\begin{tabular}{c|ccc}
			\makecell{\textbf{Importance} \\ \textbf{Rank}} & \textbf{1} & \textbf{2} & \textbf{3} \\
			\hline
			\rowcolor{red!25}
			\textbf{Synthetic Foil \#1} & \textbf{100.0} & 0.0 & 0.0 \\
			\rowcolor{green!25}
			\textbf{race} & 0.0 & \textbf{23.1} & \textbf{18.8} \\
			\rowcolor{gray!25}
			\textbf{other features} & 0.0 & \textbf{76.9} & \textbf{81.2} \\
			\end{tabular}
		} &
		\subfloat[\RoT explanations for adversarial model]{
			\begin{tabular}{c|ccc}
			\makecell{\textbf{Importance} \\ \textbf{Rank}} & \textbf{1} & \textbf{2} & \textbf{3} \\
			\hline
			\rowcolor{red!25}
			\textbf{Synthetic Foil \#1} & 0.0 & 0.0 & 0.0 \\
			\rowcolor{green!25}
			\textbf{race} & \textbf{100.0} & 0.0 & 0.0 \\
			\rowcolor{gray!25}
			\textbf{other features} & 0.0 & \textbf{100.0} & \textbf{100.0} \\
			\end{tabular}
		} \\
	\end{tabular}
\end{table}

% \begin{table}
% 	\centering
%     \caption{\textbf{COMPAS dataset, with a synthetic foil feature}: Confusion matrices showing what percentage of explanations select particular features as the most, second most, and third most important features in explanations. Ideally, the most important feature recovered by an explainer should be the only one used (highlighted {\color{green} green}), whereas the adversary tries to have the explaners always pick a different feature (highlighted {\color{red} red}). The most important feature (first column) from each matrix is replicated in Table \ref{tab:advattack}  }
%     \label{tab:conf_matrices1}
%     \includegraphics[page=26,trim=1 1 1 1,clip,width=1.0\linewidth]{figures/Combined.pdf}
% \end{table}

\begin{table}
	\centering
    \caption{\textbf{COMPAS dataset, with two synthetic foil features}: Confusion matrices showing what percentage of explanations select particular features as the most, second most, and third most important features in explanations. Ideally, the most important feature recovered by an explainer should be the only one used (highlighted {\color{green} green}), whereas the adversary tries to have the explaners always pick a different feature (highlighted {\color{red} red}). The most important feature (first column) from each matrix is replicated in Table \ref{tab:advattack}  }
    \label{tab:conf_matrices2}
    \begin{tabular}{cc}
		\subfloat[SHAP explanations for adversarially crafted model]{
			\begin{tabular}{c|ccc}
			\makecell{\textbf{Importance} \\ \textbf{Rank}} & \textbf{1} & \textbf{2} & \textbf{3} \\
			\hline
			\rowcolor{red!25}
			\textbf{Synthetic Foil \#1} & \textbf{18.9} & \textbf{23.6} & \textbf{26.4} \\
			\rowcolor{red!25}
			\textbf{Synthetic Foil \#2} & \textbf{22.3} & \textbf{27.0} & \textbf{18.4} \\
			\rowcolor{green!25}
			\textbf{race} & \textbf{\textit{35.0}} & \textbf{23.6} & \textbf{12.6} \\
			\rowcolor{gray!25}
			\textbf{other features} & \textbf{23.8} & \textbf{25.7} & \textbf{42.6} \\
			\end{tabular}
		} &
		\subfloat[\RoT explanations for adversarial model]{
			\begin{tabular}{c|ccc}
			\makecell{\textbf{Importance} \\ \textbf{Rank}} & \textbf{1} & \textbf{2} & \textbf{3} \\
			\hline
			\rowcolor{red!25}
			\textbf{Synthetic Foil \#1} & 0.0 & 1.6 & 8.1 \\
			\rowcolor{red!25}
			\textbf{Synthetic Foil \#2} & 0.0 & 0.0 & 0.0 \\
			\rowcolor{green!25}
			\textbf{race} & \textbf{100.0} & 0.0 & 0.0 \\
			\rowcolor{gray!25}
			\textbf{other features} & 0.0 & \textbf{98.4} & \textbf{91.9} \\
			\end{tabular}
		} \\
		\\
		\subfloat[LIME explanations for adversarial model]{
			\begin{tabular}{c|ccc}
			\makecell{\textbf{Importance} \\ \textbf{Rank}} & \textbf{1} & \textbf{2} & \textbf{3} \\
			\hline
			\rowcolor{red!25}
			\textbf{Synthetic Foil \#1} & \textbf{43.9} & \textbf{39.0} & \textbf{7.0} \\
			\rowcolor{red!25}
			\textbf{Synthetic Foil \#2} & \textbf{47.9} & \textbf{27.5} & \textbf{10.7} \\
			\rowcolor{green!25}
			\textbf{race} & 0.3 & \textbf{3.1} & \textbf{13.3} \\
			\rowcolor{gray!25}
			\textbf{other features} & \textbf{7.9} & \textbf{30.4} & \textbf{69.1} \\
			\end{tabular}
		} &
		\subfloat[\RoT explanations for adversarial model]{
			\begin{tabular}{c|ccc}
			\makecell{\textbf{Importance} \\ \textbf{Rank}} & \textbf{1} & \textbf{2} & \textbf{3} \\
			\hline
			\rowcolor{red!25}
			\textbf{Synthetic Foil \#1} & 0.0 & 0.0 & 0.0 \\
			\rowcolor{red!25}
			\textbf{Synthetic Foil \#2} & 0.0 & 0.0 & 0.0 \\
			\rowcolor{green!25}
			\textbf{race} & \textbf{100.0} & 0.0 & 0.0 \\
			\rowcolor{gray!25}
			\textbf{other features} & 0.0 & \textbf{100.0} & \textbf{100.0} \\
			\end{tabular}
		} \\
	\end{tabular}
\end{table}

% \begin{table}
% 	\centering
%     \caption{\textbf{COMPAS dataset, with two synthetic foil features}: Confusion matrices showing what percentage of explanations select particular features as the most, second most, and third most important features in explanations. Ideally, the most important feature recovered by an explainer should be the only one used (highlighted {\color{green} green}), whereas the adversary tries to have the explaners always pick a different feature (highlighted {\color{red} red}). The most important feature (first column) from each matrix is replicated in Table \ref{tab:advattack}  }
%     \label{tab:conf_matrices2}
%     \includegraphics[page=27,trim=1 1 1 1,clip,width=1.0\linewidth]{figures/Combined.pdf}
% \end{table}

\begin{table}
	\centering
    \caption{\textbf{Communities and Crime dataset, with a synthetic foil feature}: Confusion matrices showing what percentage of explanations select particular features as the most, second most, and third most important features in explanations. Ideally, the most important feature recovered by an explainer should be the only one used (highlighted {\color{green} green}), whereas the adversary tries to have the explaners always pick a different feature (highlighted {\color{red} red}). The most important feature (first column) from each matrix is replicated in Table \ref{tab:advattack}  }
    \label{tab:conf_matrices3}
    \begin{tabular}{cc}
		\subfloat[SHAP explanations for adversarially crafted model]{
			\begin{tabular}{c|ccc}
			\makecell{\textbf{Importance} \\ \textbf{Rank}} & \textbf{1} & \textbf{2} & \textbf{3} \\
			\hline
			\rowcolor{red!25}
			\textbf{Synthetic Foil \#1} & \textbf{87.0} & \textbf{12.5} & 0.5 \\
			\rowcolor{green!25}
			\textbf{race} & \textbf{4.0} & \textbf{40.5} & \textbf{9.5} \\
			\rowcolor{gray!25}
			\textbf{other features} & \textbf{9.0} & \textbf{42.5} & \textbf{90.0} \\
			\end{tabular}
		} &
		\subfloat[\RoT explanations for adversarial model]{
			\begin{tabular}{c|ccc}
			\makecell{\textbf{Importance} \\ \textbf{Rank}} & \textbf{1} & \textbf{2} & \textbf{3} \\
			\hline
			\rowcolor{red!25}
			\textbf{Synthetic Foil \#1} & 0.0 & 0.0 & 0.0 \\
			\rowcolor{green!25}
			\textbf{race} & \textbf{90.5} & \textbf{4.5} & 0.5 \\
			\rowcolor{gray!25}
			\textbf{other features} & \textbf{9.5} & \textbf{95.5} & \textbf{99.5} \\
			\end{tabular}
		} \\
		\\
		\subfloat[LIME explanations for adversarial model]{
			\begin{tabular}{c|ccc}
			\makecell{\textbf{Importance} \\ \textbf{Rank}} & \textbf{1} & \textbf{2} & \textbf{3} \\
			\hline
			\rowcolor{red!25}
			\textbf{Synthetic Foil \#1} & \textbf{99.5} & 0.0 & 0.0 \\
			\rowcolor{green!25}
			\textbf{race} & 0.0 & 0.5 & 0.0 \\
			\rowcolor{gray!25}
			\textbf{other features} & 0.5 & \textbf{99.5} & \textbf{100} \\
			\end{tabular}
		} &
		\subfloat[\RoT explanations for adversarial model]{
			\begin{tabular}{c|ccc}
			\makecell{\textbf{Importance} \\ \textbf{Rank}} & \textbf{1} & \textbf{2} & \textbf{3} \\
			\hline
			\rowcolor{red!25}
			\textbf{Synthetic Foil \#1} & 0.0 & 0.0 & 0.0 \\
			\rowcolor{green!25}
			\textbf{race} & \textbf{89.5} & \textbf{5.5} & 1.0 \\
			\rowcolor{gray!25}
			\textbf{other features} & \textbf{10.5} & \textbf{94.5} & \textbf{99.0} \\
			\end{tabular}
		} \\
	\end{tabular}
\end{table}

% \begin{table}
% 	\centering
%     \caption{\textbf{Communities and Crime dataset, with a synthetic foil feature}: Confusion matrices showing what percentage of explanations select particular features as the most, second most, and third most important features in explanations. Ideally, the most important feature recovered by an explainer should be the only one used (highlighted {\color{green} green}), whereas the adversary tries to have the explaners always pick a different feature (highlighted {\color{red} red}). The most important feature (first column) from each matrix is replicated in Table \ref{tab:advattack}  }
%     \label{tab:conf_matrices3}
%     \includegraphics[page=28,trim=1 1 1 1,clip,width=1.0\linewidth]{figures/Combined.pdf}
% \end{table}

\begin{table}
	\centering
    \caption{\textbf{Communities and Crime dataset, with two synthetic foil features}: Confusion matrices showing what percentage of explanations select particular features as the most, second most, and third most important features in explanations. Ideally, the most important feature recovered by an explainer should be the only one used (highlighted {\color{green} green}), whereas the adversary tries to have the explaners always pick a different feature (highlighted {\color{red} red}). The most important feature (first column) from each matrix is replicated in Table \ref{tab:advattack}  }
    \label{tab:conf_matrices4}
    \begin{tabular}{cc}
		\subfloat[SHAP explanations for adversarially crafted model]{
			\begin{tabular}{c|ccc}
			\makecell{\textbf{Importance} \\ \textbf{Rank}} & \textbf{1} & \textbf{2} & \textbf{3} \\
			\hline
			\rowcolor{red!25}
			\textbf{Synthetic Foil \#1} & \textbf{49.0} & \textbf{23.5} & \textbf{7.5} \\
			\rowcolor{red!25}
			\textbf{Synthetic Foil \#2} & \textbf{50.0} & \textbf{21.5} & \textbf{14.5} \\
			\rowcolor{green!25}
			\textbf{race} & 1.0 & \textbf{21.5} & \textbf{15.5} \\
			\rowcolor{gray!25}
			\textbf{other features} & 0.0 & \textbf{33.5} & \textbf{62.5} \\
			\end{tabular}
		} &
		\subfloat[\RoT explanations for adversarial model]{
			\begin{tabular}{c|ccc}
			\makecell{\textbf{Importance} \\ \textbf{Rank}} & \textbf{1} & \textbf{2} & \textbf{3} \\
			\hline
			\rowcolor{red!25}
			\textbf{Synthetic Foil \#1} & 0.0 & 0.0 & 0.0 \\
			\rowcolor{red!25}
			\textbf{Synthetic Foil \#2} & 0.0 & 0.0 & 0.0 \\
			\rowcolor{green!25}
			\textbf{race} & \textbf{90.0} & \textbf{4.5} & 1.5 \\
			\rowcolor{gray!25}
			\textbf{other features} & \textbf{10.0} & \textbf{95.5} & \textbf{98.5} \\
			\end{tabular}
		} \\
		\\
		\subfloat[LIME explanations for adversarial model]{
			\begin{tabular}{c|ccc}
			\makecell{\textbf{Importance} \\ \textbf{Rank}} & \textbf{1} & \textbf{2} & \textbf{3} \\
			\hline
			\rowcolor{red!25}
			\textbf{Synthetic Foil \#1} & \textbf{14.5} & \textbf{23.5} & \textbf{8.5} \\
			\rowcolor{red!25}
			\textbf{Synthetic Foil \#2} & \textbf{74.5} & \textbf{11.5} & \textbf{3.5} \\
			\rowcolor{green!25}
			\textbf{race} & 0.0 & 0.5 & 1.0 \\
			\rowcolor{gray!25}
			\textbf{other features} & \textbf{11.0} & \textbf{64.5} & \textbf{87.0} \\
			\end{tabular}
		} &
		\subfloat[\RoT explanations for adversarial model]{
			\begin{tabular}{c|ccc}
			\makecell{\textbf{Importance} \\ \textbf{Rank}} & \textbf{1} & \textbf{2} & \textbf{3} \\
			\hline
			\rowcolor{red!25}
			\textbf{Synthetic Foil \#1} & 0.0 & 0.0 & 0.0 \\
			\rowcolor{red!25}
			\textbf{Synthetic Foil \#2} & 0.0 & 0.0 & 0.0 \\
			\rowcolor{green!25}
			\textbf{race} & \textbf{91.0} & \textbf{4.0} & 1.0 \\
			\rowcolor{gray!25}
			\textbf{other features} & \textbf{9.0} & \textbf{96.0} & \textbf{99.0} \\
			\end{tabular}
		} \\
	\end{tabular}
\end{table}

% \begin{table}
% 	\centering
%     \caption{\textbf{Communities and Crime dataset, with two synthetic foil features}: Confusion matrices showing what percentage of explanations select particular features as the most, second most, and third most important features in explanations. Ideally, the most important feature recovered by an explainer should be the only one used (highlighted {\color{green} green}), whereas the adversary tries to have the explaners always pick a different feature (highlighted {\color{red} red}). The most important feature (first column) from each matrix is replicated in Table \ref{tab:advattack}  }
%     \label{tab:conf_matrices4}
%     \includegraphics[page=29,trim=1 1 1 1,clip,width=1.0\linewidth]{figures/Combined.pdf}
% \end{table}

\begin{table}
	\centering
    \caption{\textbf{German credit dataset, with an engineered foil feature (LoanRate as \% of Income)}: Confusion matrices showing what percentage of explanations select particular features as the most, second most, and third most important features in explanations. Ideally, the most important feature recovered by an explainer should be the only one used (highlighted {\color{green} green}), whereas the adversary tries to have the explaners always pick a different feature (highlighted {\color{red} red}). The most important feature (first column) from each matrix is replicated in Table \ref{tab:advattack}  }
    \label{tab:conf_matrices5}
    \begin{tabular}{cc}
		\subfloat[SHAP explanations for adversarially crafted model]{
			\begin{tabular}{c|ccc}
			\makecell{\textbf{Importance} \\ \textbf{Rank}} & \textbf{1} & \textbf{2} & \textbf{3} \\
			\hline
			\rowcolor{red!25}
			\textbf{LoanRate \% of Income} & \textbf{100.0} & 0.0 & 0.0 \\
			\rowcolor{green!25}
			\textbf{gender} & 0.0 & \textbf{40.0} & 5.0 \\
			\rowcolor{gray!25}
			\textbf{other features} & 0.0 & \textbf{60.0} & \textbf{95.0} \\
			\end{tabular}
		} &
		\subfloat[\RoT explanations for adversarial model]{
			\begin{tabular}{c|ccc}
			\makecell{\textbf{Importance} \\ \textbf{Rank}} & \textbf{1} & \textbf{2} & \textbf{3} \\
			\hline
			\rowcolor{red!25}
			\textbf{LoanRate \% of Income} & 0.0 & 0.0 & 0.0 \\
			\rowcolor{green!25}
			\textbf{gender} & \textbf{100.0} & 0.0 & 0.0 \\
			\rowcolor{gray!25}
			\textbf{other features} & 0.0 & \textbf{100.0} & \textbf{100.0} \\
			\end{tabular}
		} \\
		\\
		\subfloat[LIME explanations for adversarial model]{
			\begin{tabular}{c|ccc}
			\makecell{\textbf{Importance} \\ \textbf{Rank}} & \textbf{1} & \textbf{2} & \textbf{3} \\
			\hline
			\rowcolor{red!25}
			\textbf{LoanRate \% of Income} & \textbf{84.0} & 0.0 & 0.0 \\
			\rowcolor{green!25}
			\textbf{gender} & 0.0 & 1.0 & 1.0 \\
			\rowcolor{gray!25}
			\textbf{other features} & \textbf{16.0} & \textbf{99.0} & \textbf{99.0} \\
			\end{tabular}
		} &
		\subfloat[\RoT explanations for adversarial model]{
			\begin{tabular}{c|ccc}
			\makecell{\textbf{Importance} \\ \textbf{Rank}} & \textbf{1} & \textbf{2} & \textbf{3} \\
			\hline
			\rowcolor{red!25}
			\textbf{LoanRate \% of Income} & 0.0 & 0.0 & 2.0 \\
			\rowcolor{green!25}
			\textbf{gender} & \textbf{100.0} & 0.0 & 0.0 \\
			\rowcolor{gray!25} 
			\textbf{other features} & 0.0 & \textbf{100.0} & \textbf{98.0} \\
			\end{tabular}
		} \\
	\end{tabular}
\end{table}

\end{document}